\documentclass[final]{article}
\usepackage[utf8]{inputenc}
\usepackage[T1]{fontenc}

\usepackage{algorithm, algpseudocode}
\usepackage{amsmath, amsfonts, amssymb}
\usepackage{appendix}
\usepackage{bm}
\usepackage{extarrows}
\usepackage{graphicx}
\usepackage{hyperref}
\usepackage{multirow}
\usepackage{subcaption}

\usepackage[nonatbib, preprint]{style}
\usepackage[style=ieee]{biblatex}
\newcommand{\RR}{\mathbb{R}}
\renewcommand{\vec}{\boldsymbol}

\title{Probabilistic Pyramidal Optical Flow via Iterated Extended Kalman Filtering over Image Resolutions}
\author{
    \begin{tabular}{c}
        \textbf{Andy Wang} \\
        \texttt{awang124@jh.edu}
    \end{tabular}
    \hspace{1cm}
    \begin{tabular}{c}
        \textbf{Mario Micheli} \\
        \texttt{mmiche18@jhu.edu}
    \end{tabular} \\[4ex]
    \textnormal{Department of Applied Mathematics and Statistics} \\
    \textnormal{The Johns Hopkins University} \\
    \textnormal{Baltimore, MD 21218}
}
\begin{document}
\maketitle
\linespread{1.25}\selectfont

\begin{abstract}
Accurate motion perception and interpretation are central components of modern computer vision systems, with applications throughout science and engineering. Optical flow estimation is the mathematical computation of motion fields from sequential pairs of images in a video. Classical algorithms give deterministic point estimates, while deep learning models capable of uncertainty-aware estimation are data-intensive black boxes. The domain of structurally transparent probabilistic optical flow methods remains underdeveloped. We discuss two popular pyramidal optical flow algorithms---Lucas-Kanade and Farnebäck---and recognize that their recursion across image resolution levels parallels the recursion of the iterated extended Kalman filter (IEKF) traditionally across timesteps. We embed these algorithms within a modified IEKF to produce two novel probabilistic optical flow algorithms---Probabilistic Lucas-Kanade (PLK) and Probabilistic Farnebäck (PFB). For each of these, we develop three variants tailored to different assumptions; their respective measurement models are constructed using zeroth-order intensity constancy, first-order gradient constancy, and a combined formulation. We discuss an adaptive noise covariance estimation heuristic using measurement residuals for outlier and occlusion rejection. We evaluate our algorithms on the MPI-Sintel and KITTI-2015 benchmark datasets, and achieve substantial endpoint and F1 error reductions over classical baselines. Our methods also natively produce full flow covariance matrices, providing a geometrically interpretable quantification of estimation confidence and spatial conditioning.
\end{abstract}

\vspace{0.5em}

\newpage
\section{Introduction}
\label{sec:1}
The ability to correctly perceive and interpret motion is a fundamental pillar of modern computer vision, serving as a critical sensory link between static visual data and dynamic physical environments. As vision systems grow ever more integrated into increasingly complex, unstructured real-world settings, the demand for robust, trustworthy motion estimation has grown exponentially. Optical flow, the driving force behind such estimation, underpins a vast interdisciplinary array of high-stakes applications. In autonomous driving, it predicts trajectories of crossing pedestrians and oncoming vehicles, enabling proactive obstacle avoidance. In robotics, it forms the backbone of visual odometry, empowering terrain navigation with human-like spatial awareness. In medical imaging, it tracks tissue motion to inform pathological detection, while in fluid mechanics, it maps fluid motion to guide optimal engineering and design. In video processing, it drives compression and stabilization for smooth output with reduced overhead, while in surveillance systems, it pinpoints anomalous crowd flows for preemptive threat detection. Across these domains, optical flow fulfills a quintessential requirement: vision systems must not only recognize \underline{what} constitutes their environment, but also comprehend precisely \underline{how} that environment evolves.

At its core, \textit{optical flow} is the quantitative estimation of the apparent motion of brightness patterns between consecutive video frames. As physical objects, or a camera itself, move through a scene, the light projected onto the imaging sensor shifts: optical flow algorithms attempt to describe this shift by computing a planar two-dimensional vector field, assigning a displacement, or flow, vector to each pixel in the image array.

The historical pursuit of achieving accurate optical flow estimation is anchored by a rich literature of classical algorithms. Gibson in 1950 \cite{Gibson1950} proposed its seminal formulation, while its first foundational frameworks were proposed in 1981: Horn and Schunck \cite{Horn1981} introduced a variational framework, minimizing an energy combining a data term with a regularization term, to penalize deviation from a smooth flow field, while Lucas and Kanade \cite{Lucas1981} assumed nearby flows to be similar, allowing for construction of an overdetermined system solved by least-squares optimization. Bouguet in 2000 \cite{Bouguet2000} created a pyramidal construction of Lucas and Kanade's method, allowing large displacements to be estimated at coarser scales, then combined with precise corrections at finer resolutions. Farnebäck in 2003 \cite{Farneback2003} employed polynomial expansion, computing flows using quadratic approximations to image intensity neighborhoods. These classical paradigms, among others, possess a common characteristic: thorough interpretability. By having been built upon explicit mathematical principles, their behaviors, constraints, and failure modes can be rigorously analyzed.

In recent years, this landscape has been radically reshaped by the advent of deep learning: modern advances are characterized by rapid architectural evolution aimed at maximizing benchmark performance. FlowNet \cite{Dosovitskiy2015} demonstrated that an end-to-end convolutional neural network could directly regress flow fields from image pairs. PWC-Net \cite{Sun2018} reintroduced classical techniques, including spatial pyramids, warping, and cost volumes to significantly reduce model size. RAFT \cite{Teed2020} abandoned these pyramids in favor of iteratively updated flow fields via a recurrent GRU network, further improving estimate accuracy. More recently, transformer-based architectures, like FlowFormer \cite{Huang2022}, have leveraged attention mechanisms to capture long-range contextual flow dependencies. Some architectures, like U$^2$Flow \cite{Sun2026}, output uncertainty estimates alongside flow estimates, but all these methods share a common flaw: they are fundamentally ``black-boxes", consisting of millions of parameters and exhibiting patterns and behaviors difficult to analyze or theoretically justify. They also require massive amounts of annotated training data, and those computing uncertainties have largely opaque internal feature representations, prone to failure in out-of-distribution settings.

Thus despite these advancements, a critical gap remains at the intersection of uncertainty quantification and mathematical transparency. Classical methods are deeply interpretable but fundamentally deterministic, yielding at best post-hoc heuristic confidence measures, while modern methods can be made probabilistic but cannot be truly qualitatively understood. There are notable exceptions: ProbFlow \cite{Wannenwetsch2017} produced flow and uncertainty estimates via variational inference, proving that transparent uncertainty is possible, though it required computationally intensive graphical model inference. Continuing to bridge this gap is not a mere academic exercise, but rather a paramount imperative for the smooth operation of optical flow applications. In safety-critical systems, deterministic flow estimates are insufficient: if a self-driving vehicle encounters an unusual object or unprecedented weather condition, it must understand its confidence in its estimated motions, allowing in dubious scenarios for control to be returned to a human driver or an emergency braking protocol to be triggered. Simultaneously, confidently incorrect answers are catastrophic, barring such scenarios from proper handling: thus each uncertainty estimate must be transparent and traceable.

This work addresses this gap by fusing the transparency of classical optical flow algorithms with principled uncertainty quantification via recursive state estimation. Specifically, we produce:
\begin{itemize}
    \item \textbf{Probabilistic Lucas-Kanade}: a thorough construction of a novel optical flow estimation method, integrating Bouguet's ``pyramidal Lucas-Kanade" algorithm with iterated extended Kalman filtering over image resolution levels, with three variants, one utilizing an image intensity constancy constraint, another utilizing an analogous gradient constancy constraint, and a third combining both.
    \item \textbf{Probabilistic Farnebäck}: a similar construction of a second novel method, analogously integrating Farnebäck's algorithm with iterated extended Kalman filtering, with the same three variants.
    \item \textbf{Performance Improvement}: concrete estimation error reductions by our proposed methods over their classical baselines against established benchmarks.
    \item \textbf{Transparent Understanding}: discussions of how our algorithm designs mathematically dictate their flow and uncertainty estimates.
\end{itemize}

Our work is organized as follows:
\begin{enumerate}
    \item \textbf{Introduction}: this section.
    \item \textbf{Background}: we formally define the optical flow problem, then review the Lucas-Kanade and Farnebäck algorithms, alongside Kalman filtering.
    \item \textbf{Methodology}: we expand upon the motivation for probabilistic optical flow, then detail the core architectures of our proposed framework.
    \item \textbf{Evaluation}: we describe our benchmark datasets, their evaluation metrics, our algorithm parameter settings, and technical implementation details.
    \item \textbf{Results}: we compare our benchmark performance to those of classical algorithms and provide visualizations.
    \item \textbf{Discussion}: we analyze our results, including how our parameter settings affect them, discuss limitations of our algorithms, and propose avenues for further exploration.
\end{enumerate}

\section{Background}
\label{sec:2}
Before presenting our probabilistic motion framework, we must first establish the mathematical foundations of their core components. This section prescribes precise definitions for image intensities and optical flows, then details two distinct classical paradigms for optical flow estimation---the intensity gradient-based approach by Lucas and Kanade, and the polynomial expansion-based technique by Farnebäck---alongside recursive state estimation via the Kalman filter and its extended and iterated variants.

\subsection{Fundamentals}
\label{sec:2.1}
A \textit{digital image} is a map $\vec{I}$ from a discrete 2-dimensional domain $\mathcal{D}$ (i.e. a lattice of points) to an intensity---an $n$-tuple of values over a discrete domain $\mathcal{R}$ encoding color, luminance, etc.
\begin{equation}
    \vec{I}:\underbrace{\{0,\cdots,H-1\}\times\{0,\cdots,W-1\}}_{\mathcal{D}}\to{\underbrace{\{0,\cdots,C-1\}}_{\mathcal{R}}}^n
\end{equation}

We call this map the \textit{intensity}, and denote its input as $(x,y)$---its corresponding output is thus denoted $I(x,y)$. For example, an 8-bit 720p RGB image has $H=720,W=1280,C=2^8,n=3$. \\\\ We restrict our attention to \textit{grayscale} images, those which assign to each point solely a luminance (no color, etc.), namely those for which $n=1$, denoted $I$ (without vector boldface). \\\\ A \textit{video} may be viewed simultaneously as a sequence of images $(I_t)_{t\in\mathcal{T}}$, called \textit{frames}, indexed by discrete timesteps $\mathcal{T}$, or as a map from the Cartesian product of $\mathcal{D}$ with these timesteps:
\begin{equation}
    I:\mathcal{D}\times\underbrace{\{0,\cdots,T-1\}}_{\mathcal{T}}\to\mathcal{R}
\end{equation}

In this latter case we denote its input as $(x,y,t)$---its output is thus $I(x,y,t)$. We will adopt either view in certain situations for notational convenience. \\\\ Finally, we define a \textit{Gaussian image pyramid} \cite{Burt1981} as a sequence consisting of the same ``image" (in the real-world sense) at varying resolutions. Given an image $I$, to construct a pyramid of height $L$, we define its ``bottom" or ``finest level" (``finest" in terms of resolution) as our original image: $I^{(L)}\triangleq I$. We then recursively construct each next ``higher" or ``coarser level" $I^{(\ell)}$, where $\ell$ ranges from $L-1$ to $1$, by first convolving our current level $I^{(\ell)}$ with a $5\times5$ Gaussian weighting kernel $w$, then subsampling the result by a factor of 2. Namely, $\forall(x,y)\in\{0,\cdots,\lfloor H/2\rfloor\}\times\{0,\cdots,\lfloor W/2\rfloor\}$:
\begin{equation}
    I^{(\ell-1)}(x,y)=(w*I^{(\ell)})(2x,2y)
\end{equation}

More explicitly:
\begin{equation}
\label{eq:pyr}
    I^{(\ell-1)}(x,y)=\sum_{m=-2}^2\sum_{n=-2}^2w(m,n)I^{(\ell)}(2x-m,2y-n)
\end{equation}

We denote by $\mathcal{D}^{(\ell)}$ the reduced domain of $I^{(\ell)}$, for $\ell=1,\cdots,L$:
\begin{equation}
    \mathcal{D}^{(\ell)}\triangleq\{0,\cdots,\lfloor H/2^{L-\ell}\rfloor\}\times\{0,\cdots,\lfloor W/2^{L-\ell}\rfloor\}
\end{equation}

\begin{figure}[t]
    \centering
    \includegraphics[width=0.9\textwidth]{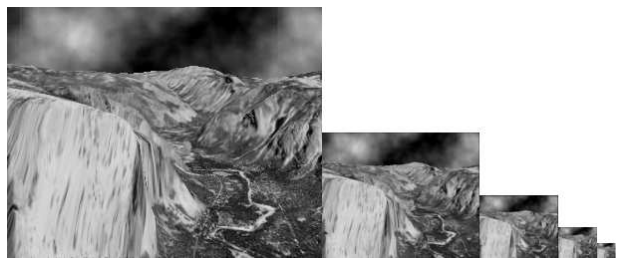}
    \caption{$5$-Level Gaussian Image Pyramid}
    \label{fig:pyramid}
\end{figure}

\subsection{Lucas-Kanade Optical Flow}
\label{sec:2.2}

\subsubsection{Optical Flow Constraint}
\label{sec:2.2.1}
The Lucas-Kanade method \cite{Lucas1981} is a landmark optical flow estimation approach, which exploits three fundamental assumptions. First, that the intensity of a moving object in an image sequence remains invariant over its trajectory, giving the \textit{Brightness Constancy Equation} (BCE). Namely, for a pixel moving from $(x,y)$ at time $t$ to $(x+dx,y+dy)$ at time $t+dt$, we assume:
\begin{equation}
\label{eq:bce}
    I(x,y,t)=I(x+dx,y+dy,t+dt)
\end{equation}

To derive a computable linear system, we assume second that displacement between frames is sufficiently small (i.e. sub-pixel), such that we may apply a first-order Taylor expansion:
\begin{equation}
\label{eq:bcete}
    I(x+dx,y+dy,t+dt)\approx I(x,y,t)+\underbrace{\frac{\partial I}{\partial x}(x,y,t)}_{I_x(x,y)}dx+\underbrace{\frac{\partial I}{\partial y}(x,y,t)}_{I_y(x,y)}dy+\underbrace{\frac{\partial I}{\partial t}(x,y,t)}_{I_t(x,y)}dt
\end{equation}

Substituting this into the BCE, subtracting $I(x,y,t)$, and dividing by $dt$ gives the \textit{Optical Flow Constraint Equation} (OFCE):
\begin{equation}
\label{eq:ofce}
    I_x(x,y)\underbrace{\frac{dx}{dt}}_{u(x,y)}+I_y(x,y)\underbrace{\frac{dy}{dt}}_{v(x,y)}+I_t(x,y)=0
\end{equation}

Here $I_x(x,y),I_y(x,y)$ denote the spatial gradients of $I$, evaluated at $(x,y)$, $I_t(x,y)$ denotes the temporal gradient likewise, and $\vec{f}(x,y)\triangleq(u(x,y),v(x,y))$ denotes the pixel's displacement, or \textit{optical flow}, between frames $t$ and $t+dt$.

\subsubsection{Intensity Gradients}
\label{sec:2.2.2}
As our intensities are discrete, we employ numerical approximations to ``true differentiation". For the spatial gradients, convolution with a central difference kernel ($[-1\ 0\ 1]$) provides a first-order approximation, but is sensitive to high-frequency noise. We thus use \textit{Sobel operators} \cite{Sobel1968}, which apply a weighted averaging kernel ($[-1\ 2\ -1]$) orthogonal to the axis of differentiation.
\begin{equation}
    S_x=\begin{bmatrix}-1&0&1\\-2&0&2\\-1&0&1\end{bmatrix}\qquad S_y=\begin{bmatrix}-1&-2&-1\\0&0&0\\1&2&1\end{bmatrix}
\end{equation}

Namely, $I_x$ is computed by convolution with $S_x$, which approximates the derivative along the rows while smoothing along the columns:
\begin{equation}
\label{eq:gradx}
    I_x(x,y)=(S_x*I)(x,y)=\sum_{m=-1}^1\sum_{n=-1}^1S_x(m,n)I(x-m,y-n)
\end{equation}

Conversely, $I_y$ is computed by convolution with $S_y$, which approximates the derivative along the columns while smoothing along the rows:
\begin{equation}
\label{eq:grady}
    I_y(x,y)=(S_y*I)(x,y)=\sum_{m=-1}^1\sum_{n=-1}^1S_y(m,n)I(x-m,y-n)
\end{equation}

As convolution will extend beyond the image domain at points near coordinate boundaries (i.e. where $x+m\notin\{0,\cdots,H-1\}$ or $y+n\notin\{0,\cdots,W-1\}$), we must employ a border handling strategy. Zero-padding extends a signal of intensities $[a,b,c,d]$ by applying zeros (i.e. $[\cdots,0,0\mid a,b,c,d\mid0,0,\cdots]$), causing sharp discontinuities at boundaries, which introduce spurious high-frequency artifacts into computed gradients. Clamp-padding applies the value at the boundary (i.e. $[\cdots,a,a\mid a,b,c,d\mid d,d,\cdots]$), which, while avoiding discontinuities, introduces flat regions at boundaries, which suppress gradient magnitudes. We thus use \textit{mirror-padding}, which reflects the signal about the boundary (i.e. $[\cdots,b,a\mid a,b,c,d\mid d,c,\cdots]$), mitigating both issues by ensuring intensities and their local texture extend smoothly beyond their original domain, yielding accurate gradient estimates at near-boundary pixels. Formally, for a 1-dimensional signal $f$ of length $N$:
\begin{equation}
\label{eq:pad}
    \forall k\in\{1,2,\cdots\}\qquad f(-k)=f(k)\qquad f(N-1+k)=f(N-1-k)
\end{equation}

This generalizes trivially to 2 dimensions by being applied independently to both axes. \\\\ For the temporal gradient, as consecutive frames are sampled at unit time intervals (i.e. $\Delta t=1$), we use the first-order forward difference between \mbox{current frame $I_1$ and subsequent frame $I_2$:}
\begin{equation}
    I_t(x,y)=I_2(x,y)-I_1(x,y)
\end{equation}

The forward formulation is natural, given that the \hyperref[eq:ofce]{OFCE} relates the current and subsequent frames.

\subsubsection{Least-Squares Estimation}
\label{sec:2.2.3}
The \hyperref[eq:ofce]{OFCE} constitutes a single linear equation of two unknowns ($u,v$): this underdetermined system admits an infinite family of solutions, a fundamental ambiguity known in computer vision as the \textit{aperture problem}. To understand this, we rewrite the OFCE:
\begin{equation}
    {\underbrace{\begin{bmatrix}I_x(x,y)\\I_y(x,y)\end{bmatrix}}_{\nabla I(x,y)}}^T\begin{bmatrix}u(x,y)\\v(x,y)\end{bmatrix}+I_t(x,y)=0
\end{equation}

In $uv$-space, this defines a line with normal vector $\nabla I(x,y)$: we thus may only uniquely determine the component of $(u(x,y),v(x,y))$ parallel to $\nabla I(x,y)$---motion along the intensity gradient. This \textit{normal flow} is given by:
\begin{equation}
    f(x,y)_n=-\frac{I_t(x,y)}{\|\nabla I(x,y)\|}\cdot\frac{\nabla I(x,y)}{\|\nabla I(x,y)\|}
\end{equation}

However, the component orthogonal to $\nabla I$---motion along a line of invariant intensity---is entirely invisible within our aperture. Thus Lucas-Kanade's third assumption is \textit{spatial coherence}, that pixels near one another share uniform motion. For some odd $K$, we pad $I_1,I_2$ by thickness $(K-1)/2$ via Eq.~\eqref{eq:pad}. Then for each original pixel $\vec{x}=(x,y)\in\mathcal{D}$, we then construct a $K\times K$ pixel \textit{window}:
\begin{equation}
    \mathcal{W}_K(\vec{x})\triangleq\left\{(x+m,y+n)\mid m,n\in\left\{-\frac{K-1}{2},\cdots,\frac{K-1}{2}\right\}\right\}
\end{equation}

Assuming all pixels share motion $(u(\vec{x}),v(\vec{x}))$, we obtain $K^2$ equations for two unknowns:
\begin{equation}
\label{eq:lkmat}
    \underbrace{\begin{bmatrix}I_x(\vec{x}_1)&I_y(\vec{x}_1)\\\vdots&\vdots\\I_x(\vec{x}_{K^2})&I_y(\vec{x}_{K^2})\end{bmatrix}}_{\vec{\nabla I}(\vec{x})}\underbrace{\begin{bmatrix}u(\vec{x})\\v(\vec{x})\end{bmatrix}}_{\vec{f}(\vec{x})}+\underbrace{\begin{bmatrix}I_t(\vec{x}_1)\\\vdots\\I_t(\vec{x}_{K^2})\end{bmatrix}}_{\vec{I_t}(\vec{x})}=\vec{0}
\end{equation}

Here and henceforth $\vec{x}_1,\cdots,\vec{x}_{K^2}$ denote the members of $\mathcal{W}_K(\vec{x})$. If this image neighborhood is ``well-conditioned", the system produces a unique least-squares solution:
\begin{equation}
\begin{aligned}
    \vec{f}(\vec{x}) &= \underset{\vec{f}(\vec{x})}{\arg\min}(\vec{\nabla I}(\vec{x})\vec{f}(\vec{x})+\vec{I_t}(\vec{x}))^T(\vec{\nabla I}(\vec{x})\vec{f}(\vec{x})+\vec{I_t}(\vec{x})) \\&= -\left(\vec{\nabla I}(\vec{x})^T\vec{\nabla I}(\vec{x})\right)^{-1}\vec{\nabla I}(\vec{x})^T\vec{I_t}(\vec{x})
\end{aligned}
\end{equation}

However, this does not always occur. Consider the \textit{structure tensor}:
\begin{equation}
    \vec{\nabla I}(\vec{x})^T\vec{\nabla I}(\vec{x})=\sum_{\vec{y}\in\mathcal{W}_K(\vec{x})}\begin{bmatrix}I_x(\vec{y})^2&I_x(\vec{y})I_y(\vec{y})\\I_x(\vec{y})I_y(\vec{y})&I_y(\vec{y})^2\end{bmatrix}
\end{equation}

The invertibility of this matrix is a critical indicator of motion computability. If our region contains two-dimensional textures (e.g. corners, textured patterns), then its gradient vectors have varying directions: both eigenvalues are large, producing a well-conditioned flow estimate. If our region has only one-dimensional texture (e.g. a single edge, parallel lines), then all gradient vectors are nearly parallel (i.e. the aperture problem persists): one eigenvalue is near zero, and our estimate is ill-conditioned. If our region is textureless (e.g. a flat plane, uniformly colored object), gradient vectors are almost non-existent: both eigenvalues are near zero, and our estimate remains ill-conditioned.

\subsubsection{Weighted Lucas-Kanade}
\label{sec:2.2.4}
The standard Lucas-Kanade formulation (Eq.~\eqref{eq:lkmat}) treats $\vec{x}_1,\cdots,\vec{x}_{K^2}$ with equal importance, though an alternative, weighted scheme \cite{Lucas1981} is possible. Most often this is Gaussian weighting:
\begin{equation}
    w_{\sigma}(\Delta\vec{x})\triangleq\exp\left(-\frac{\|\Delta\vec{x}\|_2^2}{2\sigma^2}\right)
\end{equation}

\begin{equation}
\label{eq:lkwgt}
    \vec{W}(\vec{x})\triangleq\operatorname{diag}(w_{\sigma}(\vec{x}_1-\vec{x}),\cdots,w_{\sigma}(\vec{x}_{K^2}-\vec{x}))
\end{equation}

The modified system then produces a weighted least-squares solution:
\begin{equation}
\label{eq:lklsq}
\begin{aligned}
    \vec{f}(\vec{x}) &= \underset{\vec{f}(\vec{x})}{\arg\min}(\vec{\nabla I}(\vec{x})\vec{f}(\vec{x})+\vec{I_t}(\vec{x}))^T\vec{W}(\vec{x})(\vec{\nabla I}(\vec{x})\vec{f}(\vec{x})+\vec{I_t}(\vec{x})) \\&= -\left(\vec{\nabla I}(\vec{x})^T\vec{W}(\vec{x})\vec{\nabla I}(\vec{x})\right)^{-1}\vec{\nabla I}(\vec{x})^T\vec{W}(\vec{x})\vec{I_t}(\vec{x})
\end{aligned}
\end{equation}

The Gaussian weighting scheme serves three purposes. First, it reinforces the assumption of spatial coherence: pixels near window centers, most likely to share motion, are up-weighted, while more peripheral pixels, having possibly crossed object boundaries, are naturally down-weighted. Second, it improves robustness by suppressing high-frequency noise in gradient estimates. Third, and most subtly, it improves structure tensor conditioning: by effectively reducing window size, it prevents strong distant edges from artificially inflating $\lambda_{\max}$, while strengthening weak two-dimensional texture near window centers, increasing $\lambda_{\min}$, yielding a more stable structure tensor and flow estimate.

\subsubsection{Pyramidal Lucas-Kanade}
\label{sec:2.2.5}
Weighted Lucas-Kanade still relies heavily on the assumption of small motions underpinning the \hyperref[eq:bce]{BCE} Taylor expansion (Eq.~\eqref{eq:bcete}). When flow magnitude $\|\vec{f}(\vec{x})\|_2$ exceeds approximately one pixel, this linearization becomes invalid, and our least-squares estimate diverges. To overcome this, Lucas-Kanade is embedded within a \hyperref[eq:pyr]{Gaussian image pyramid}. By recursively subsampling the image by a factor of 2, a displacement of $2^{L-\ell}$ pixels in $I^{(L)}$ becomes just 1 pixel at coarsest level $I^{(1)}$. \\\\ The pyramidal Lucas-Kanade algorithm \cite{Bouguet2000} accepts a pixel's starting position $\vec{x}$ in $I_1$, and computes its ending position $\vec{x}'$ in $I_2$. It first downscales this starting position $\vec{x}_0^{(1)}\gets\vec{x}/2^{L-1}$, then computes flow increment $\vec{f}(\vec{x})^{(1)}$ by Eq.~\eqref{eq:lklsq}, and updates its position $\vec{x}'^{(1)}\gets\vec{x}^{(1)}+\vec{f}(\vec{x})^{(1)}$. Critically, it then propagates this position down to the next finer pyramid level: $\vec{x}^{(2)}\gets2\vec{x}'^{(1)}$. This iteration continues until final position $\vec{x}'^{(L)}$ is computed, upon which $\vec{f}(\vec{x})=\vec{x}'^{(L)}-\vec{x}$. \\\\ Estimation at all levels $\ell=1,\cdots,L$ shares window size $K$: larger windows at coarser levels allow our targeted large-scale motions to be captured, while smaller windows at finer levels allow small-scale motions to be reincorporated. Central to this algorithm is the \textit{warping} operation: at level $\ell$, with scaled starting position $\vec{x}_0^{(\ell)}\gets\vec{x}/2^{L-\ell}$ and current position $\vec{x}^{(\ell)}$ (either $2\vec{x}^{(\ell-1)}$ from the previous level, or $\vec{x}_0^{(\ell)}$ for $\ell=1$), rather than computing $I_t(\vec{y})$ as raw difference $I_2^{(\ell)}(\vec{y})-I_1^{(\ell)}(\vec{y})$, we instead warp $I_2$ by our current displacement:
\begin{equation}
    I_t(\vec{y})\gets I_2^{(\ell)}\left(\vec{y}+(\vec{x}^{(\ell)}-\vec{x}_0^{(\ell)})\right)-I_1^{(\ell)}(\vec{y})
\end{equation}

As $\vec{y}+(\vec{x}^{(\ell)}-\vec{x}_0^{(\ell)})$ is generally non-integer, we perform bilinear interpolation, defined for input $(i,j)\in\RR^2$ and map $g:\mathbb{Z}^2\to\mathcal{X}$ for any arbitrary $\mathcal{X}$ by:
\begin{equation}
\label{eq:interp}
\begin{aligned}
    \operatorname{INTERP}((i,j);\ g) &\triangleq (1-\underbrace{(i-\lfloor i\rfloor)}_{\alpha})(1-\underbrace{(j-\lfloor j\rfloor)}_{\beta})\ g(\lfloor i\rfloor,\lfloor j\rfloor) \\&+ \alpha(1-\beta)\ g(\lceil i\rceil,\lfloor j\rfloor) \\&+ (1-\alpha)\beta\ g(\lfloor i\rfloor,\lceil j\rceil) \\&+ \alpha\beta\ g(\lceil i\rceil,\lceil j\rceil)
\end{aligned}
\end{equation}

This transforms the problem at level $\ell$ from estimating total displacement to estimating only residual displacement remaining after current displacement $\vec{x}^{(\ell)}-\vec{x}_0^{(\ell)}$ is applied. As this current displacement brings $I_1$ and warped $I_2$ into approximate alignment, the linearization of the residual is more accurate than the linearization of the total. However, even with coarse-to-fine propagation, a single linear solve at each level is often insufficient. Near motion boundaries, spatial coherence fails: windows contain pixels belonging to distinct objects with varying motions: a single solve produces a compromised flow misaligning all surfaces, but after warping, pixels belonging to better-aligned surfaces contribute smaller residuals, while worse-aligned surfaces contribute larger residuals, allowing subsequent solves to generate corrections progressively shifting estimates toward regions' dominant motions. \\\\ We thus perform several iterations at each level: each iteration $n$ recomputes spatial and temporal gradients using its predecessor's displacement $\vec{x}'^{(\ell)}_{(n-1)}-\vec{x}^{(\ell)}_{(0)}$, and solves for a new residual increment. Iteration terminates upon either reaching maximum iterations $N$, or when residual increment magnitude $\left\|\vec{x}'^{(\ell)}_{(n)}-\vec{x}'^{(\ell)}_{(n-1)}\right\|_2$ falls below convergence tolerance $\epsilon$. \\\\ The complete Pyramidal Lucas-Kanade framework is given by Algorithm~\ref{alg:lk} in Appendix~\ref{sec:a}.

\subsection{Farnebäck Optical Flow}
\label{sec:2.3}

\subsubsection{Polynomial Approximations}
\label{sec:2.3.1}
The Lucas-Kanade method estimates optical flow by minimizing the brightness constancy residual under a first-order Taylor expansion: it is inherently gradient-based, and struggles in textureless regions where gradients vanish. Farnebäck proposed an alternative paradigm \cite{Farneback2003}, which avoids gradients entirely, and is centered on the key assumption that the intensity in a small neighborhood around a pixel $\vec{x}$ may be modeled by a quadratic polynomial:
\begin{equation}
\label{eq:fbpoly}
    I(\vec{x}+\Delta\vec{x})\approx\Delta\vec{x}^T\vec{A}(\vec{x})\Delta\vec{x}+\vec{b}(\vec{x})^T\Delta\vec{x}+c(\vec{x})
\end{equation}

Here symmetric $\vec{A}(\vec{x})\in\RR^{2\times2}$ captures local second-order intensity curvature, $\vec{b}(\vec{x})\in\RR^2$ captures gradient direction, and $c(\vec{x})$ is a scalar offset. \\\\ To estimate these coefficients $\vec{A}(\vec{x}),\vec{b}(\vec{x}),c(\vec{x})$, we perform weighted least-squares estimation over a local neighborhood. Namely, over window $\mathcal{W}_{K_p}(\vec{x})$, we minimize the weighted sum of squared residuals between the true intensities and their polynomial approximations:
\begin{equation}
\begin{aligned}
    & \vec{A}(\vec{x}),\vec{b}(\vec{x}),c(\vec{x})=\underset{\vec{A}(\vec{x}),\vec{b}(\vec{x}),c(\vec{x})}{\arg\min}\sum_{\vec{x}+\Delta\vec{x}\in\mathcal{W}_{K_p}(\vec{x})}w_{\sigma_p}(\Delta\vec{x})\aleph \\& \aleph=\left[I(\vec{x}+\Delta\vec{x})-(\Delta\vec{x}^T\vec{A}(\vec{x})\Delta\vec{x}+\vec{b}(\vec{x})^T\Delta\vec{x}+c(\vec{x}))\right]^2
\end{aligned}
\end{equation}

As the polynomial model is linear in its coefficients, we obtain a closed-form solution via the normal equations. For $\Delta\vec{x}=(dx,dy),\vec{A}(\vec{x})=\left[\begin{smallmatrix}a&b\\b&c\end{smallmatrix}\right],\vec{b}(\vec{x})=(d,e)^T,c(\vec{x})=f$, define:
\begin{equation}
\begin{aligned}
    \vec{\theta}(\vec{x}) &\triangleq (a,b,c,d,e,f)^T \\ \vec{\varphi}(\Delta\vec{x}) &\triangleq (dx^2,2dxdy,dy^2,dx,dy,1)^T
\end{aligned}
\end{equation}

Then the polynomial approximation of $I(\vec{x}+\Delta\vec{x})$ becomes $\vec{\varphi}(\Delta\vec{x})^T\vec{\theta}(\vec{x})$. Substituting this into our objective gives:
\begin{equation}
\label{eq:fbplsq}
\begin{aligned}
    \vec{\theta}(\vec{x}) &= \underset{\vec{\theta}(\vec{x})}{\arg\min}\sum_{\vec{x}+\Delta\vec{x}\in\mathcal{W}_{K_p}(\vec{x})}w_{\sigma_p}(\Delta\vec{x})\left(I(\vec{x}+\Delta\vec{x})-\vec{\varphi}(\Delta\vec{x})^T\vec{\theta}(\vec{x})\right)^2 \\&= {\underbrace{\left(\sum w_{\sigma_p}(\Delta\vec{x})\vec{\varphi}(\Delta\vec{x})\vec{\varphi}(\Delta\vec{x})^T\right)}_{\vec{G}}}^{-1}\left(\sum w_{\sigma_p}(\Delta\vec{x})\vec{\varphi}(\Delta\vec{x})I(\vec{x}+\Delta\vec{x})\right)
\end{aligned}
\end{equation}

From our solution $\vec{\theta}(\vec{x})$ we recover $\vec{A}(\vec{x}),\vec{b}(\vec{x}),c(\vec{x})$. As Gram matrix $\vec{G}$ does not depend on $\vec{x}$, it may be precomputed once for the entire image, rendering coefficient estimation highly efficient even when tracking many pixels.

\subsubsection{Approximation Translation}
\label{sec:2.3.2}
Consider two consecutive frames $I_1,I_2$, with respective coefficients $(\vec{A}_1(\vec{x}),\vec{b}_1(\vec{x}),c_1(\vec{x}))_{\vec{x}\in\mathcal{D}}$ and $(\vec{A}_2(\vec{x}),\vec{b}_2(\vec{x}),c_2(\vec{x}))_{\vec{x}\in\mathcal{D}}$. Farnebäck shares Lucas-Kanade's assumptions of brightness constancy and spatial coherence. Namely, given optical flow $\vec{f}(\vec{x})$, it assumes:
\begin{equation}
\label{eq:fbeq1}
    \forall\vec{y}=\vec{x}+\Delta\vec{x}\in\mathcal{W}_K(\vec{x})\qquad I_1(\vec{y})=I_2(\vec{y}+\vec{f}(\vec{x}))
\end{equation}

The approximation of $I_2(\vec{y}+\vec{f}(\vec{x}))$ is:
\begin{equation}
\begin{aligned}
    I_2(\vec{y}+\vec{f}(\vec{x})) &\approx (\Delta\vec{x}+\vec{f}(\vec{x}))^T\vec{A}_2(\vec{x})(\Delta\vec{x}+\vec{f}(\vec{x}))+\vec{b}_2(\vec{x})^T(\Delta\vec{x}+\vec{f}(\vec{x}))+c_2(\vec{x}) \\&= \left[\Delta\vec{x}^T\vec{A}_2(\vec{x})\Delta\vec{x}+(2\vec{A}_2(\vec{x})\vec{f}(\vec{x}))^T\Delta\vec{x}+\vec{f}(\vec{x})^T\vec{A}_2(\vec{x})\vec{f}(\vec{x})\right] \\&\quad+ \left[\vec{b}_2(\vec{x})^T\Delta\vec{x}+\vec{b}_2(\vec{x})^T\vec{f}(\vec{x})\right]+c_2(\vec{x}) \\&= \underbrace{\Delta\vec{x}^T\vec{A}_2(\vec{x})\Delta\vec{x}}_{\text{Quadratic}}+\underbrace{\left[(\vec{b}_2(\vec{x})+2\vec{A}_2(\vec{x})\vec{f}(\vec{x}))^T\Delta\vec{x}\right]}_{\text{Linear}} \\&\quad+ \underbrace{\left[\vec{f}(\vec{x})^T\vec{A}_2(\vec{x})\vec{f}(\vec{x})+\vec{b}_2(\vec{x})^T\vec{f}(\vec{x})+c_2(\vec{x})\right]}_{\text{Constant}}
\end{aligned}
\end{equation}

We equate the approximation of $I_1(\vec{y})$ to this expression, and drop constant (in $\vec{f}(\vec{x})$) terms for brevity, to obtain:
\begin{equation}
    \Delta\vec{x}^T\vec{A}_1(\vec{x})\Delta\vec{x}+\vec{b}_1(\vec{x})^T\Delta\vec{x}=\Delta\vec{x}^T\vec{A}_2(\vec{x})\Delta\vec{x}+(\vec{b}_2(\vec{x})+2\vec{A}_2(\vec{x})\vec{f}(\vec{x}))^T\Delta\vec{x}
\end{equation}

This produces two relations. The first asserts that approximated intensity curvature is invariant under translation. The second links the difference in linear coefficients directly to the optical flow:
\begin{equation}
\begin{aligned}
    \vec{A}_1(\vec{x}) &= \vec{A}_2(\vec{x}) \\ \vec{b}_1(\vec{x}) &= \vec{b}_2(\vec{x})+2\vec{A}_2(\vec{x})\vec{f}(\vec{x})
\end{aligned}
\end{equation}

As $\vec{A}_1(\vec{x}),\vec{A}_2(\vec{x})$ are often noisy and not perfectly equal, in practice we use their average. We also define the linear coefficient difference:
\begin{equation}
\begin{aligned}
    \vec{A}(\vec{x}) &\triangleq \frac{1}{2}(\vec{A}_1(\vec{x})+\vec{A}_2(\vec{x})) \\ \Delta\vec{b}(\vec{x}) &\triangleq \frac{1}{2}(\vec{b}_1(\vec{x})-\vec{b}_2(\vec{x}))
\end{aligned}
\end{equation}

These relations thus produce a linear system from which we compute our optical flow:
\begin{equation}
\label{eq:fbeq2}
    \vec{A}(\vec{x})\vec{f}(\vec{x})=\Delta\vec{b}(\vec{x})
\end{equation}

\subsubsection{Least-Squares Estimation}
\label{sec:2.3.3}
The system $\vec{A}(\vec{x})\vec{f}(\vec{x})=\Delta\vec{b}(\vec{x})$ provides a direct estimate of $\vec{f}(\vec{x})$: in practice, however, it is frequently ill-conditioned or singular. $\vec{A}(\vec{x})$ describes the local intensity curvature: if $\mathcal{W}_{K_p}(\vec{x})$ is textureless or contains only a single edge, $\vec{A}(\vec{x})$ is rank-deficient, and the system admits infinitely many solutions along the edge direction. This is precisely the aperture problem discussed in Section~\ref{sec:2.2.3}, now manifesting in the curvature matrix rather than the structure tensor. Even when $\vec{A}(\vec{x})$ is invertible, because coefficients are computed via least-squares fits over small windows, small intensity perturbations may significantly alter them, producing unstable flow estimates. \\\\ To overcome both issues, Farnebäck aggregates a system of constraints from a window $\mathcal{W}_K(\vec{x})$, exactly as in Lucas-Kanade ($I_1,I_2$ are again first padded, such that each pixel has a valid window):
\begin{equation}
\label{eq:fbmat}
    \underbrace{\begin{bmatrix}\vec{A}(\vec{x}_1)\\\vdots\\\vec{A}(\vec{x}_{K^2})\end{bmatrix}}_{\vec{\mathcal{A}}(\vec{x})}\vec{f}(\vec{x})=\underbrace{\begin{bmatrix}\Delta\vec{b}(\vec{x}_1)\\\vdots\\\Delta\vec{b}(\vec{x}_{K^2})\end{bmatrix}}_{\vec{\mathcal{B}}(\vec{x})}
\end{equation}

It then computes the least-squares solution:
\begin{equation}
\label{eq:fblsq}
\begin{aligned}
    \vec{f}(\vec{x}) &= \underset{\vec{f}(\vec{x})}{\arg\min}\sum_{\vec{y}\in\mathcal{W}_K(\vec{x})}w_{\sigma}(\vec{y}-\vec{x})\|\vec{A}(\vec{y})\vec{f}(\vec{x})-\Delta\vec{b}(\vec{y})\|^2 \\&= \left(\sum w_{\sigma}(\vec{y}-\vec{x})\vec{A}(\vec{y})^T\vec{A}(\vec{y})\right)^{-1}\left(\sum w_{\sigma}(\vec{y}-\vec{x})\vec{A}(\vec{y})^T\Delta\vec{b}(\vec{y})\right)
\end{aligned}
\end{equation}

Or equivalently in matrix form, we first compute modified weight matrix $\vec{\mathcal{W}}(\vec{x})$ as the Kronecker product of $\vec{W}(\vec{x})$ with $\vec{I}_2$, to cast each weight to both dimensions of its corresponding constraint:
\begin{equation}
\label{eq:fbwgt}
    \vec{\mathcal{W}}(\vec{x})\triangleq\vec{W}(\vec{x})\otimes\vec{I}_2=\begin{bmatrix}w_{\sigma}(\vec{x}_1-\vec{x})\vec{I}_2&\vec{O}&\cdots&\vec{O}\\\vec{O}&w_{\sigma}(\vec{x}_2-\vec{x})\vec{I}_2&\cdots&\vec{O}\\\vdots&\vdots&\ddots&\vdots\\\vec{O}&\vec{O}&\cdots&w_{\sigma}(\vec{x}_{K^2}-\vec{x})\vec{I}_2\end{bmatrix}
\end{equation}

The least-squares solution is thus:
\begin{equation}
\begin{aligned}
    \vec{f}(\vec{x}) &= \underset{\vec{f}(\vec{x})}{\arg\min}(\vec{\mathcal{A}}(\vec{x})\vec{f}(\vec{x})-\vec{\mathcal{B}}(\vec{x}))^T\vec{\mathcal{W}}(\vec{x})(\vec{\mathcal{A}}(\vec{x})\vec{f}(\vec{x})-\vec{\mathcal{B}}(\vec{x})) \\&= \left(\vec{\mathcal{A}}(\vec{x})^T\vec{\mathcal{W}}(\vec{x})\vec{\mathcal{A}}(\vec{x})\right)^{-1}\vec{\mathcal{A}}(\vec{x})^T\vec{\mathcal{W}}(\vec{x})\vec{\mathcal{B}}(\vec{x})
\end{aligned}
\end{equation}

\subsubsection{Pyramidal Farnebäck}
\label{sec:2.3.4}
Like Lucas-Kanade, by employing Taylor approximations of intensity surfaces, Farnebäck relies on the critical assumption of sufficiently small displacements. Namely, Eq.~\eqref{eq:fbeq1} is approximated by equating the approximations of $I_1(\vec{y})$ and $I_2(\vec{y}+\vec{f}(\vec{x}))$: if $\vec{f}(\vec{x})$ is overly large, then $\vec{A}_2(\vec{x}),\vec{b}_2(\vec{x})$ no longer accurately represent intensity structure at $\vec{y}+\vec{f}(\vec{x})$, rendering system Eq.~\eqref{eq:fbeq2} invalid. \\\\ This failure mode is entirely analogous to Lucas-Kanade's large motion failure mode: polynomial approximations are only locally accurate. To overcome this, Farnebäck is embedded within the Gaussian image pyramid framework \cite{Farneback2003}: large displacements in the original image become sub-pixel at coarser levels, ensuring coefficient matching remains locally valid, upon which a well-conditioned flow estimate is propagated down to finer levels. \\\\ Like \hyperref[sec:2.2.5]{pyramidal Lucas-Kanade}, the pyramidal Farnebäck algorithm accepts a pixel's starting position $\vec{x}$ in $I_1$, and computes its ending position $\vec{x}'$ in $I_2$. It first downscales this starting position to the coarsest level $\vec{x}^{(1)}_{(0)}\gets\vec{x}/2^{L-1}$, then iteratively computes flow increments $\vec{f}(\vec{x})^{(1)}_{(1)},\vec{f}(\vec{x})^{(1)}_{(2)},\cdots$ by Eq.~\eqref{eq:fblsq} with each increment updating its position $\vec{x}'^{(1)}_{(n)}\gets\vec{x}'^{(1)}_{(n-1)}+\vec{f}(\vec{x})^{(1)}_{(n)}$, until reaching either maximum iterations $N$ or convergence $\left\|\vec{x}'^{(1)}_{(n)}-\vec{x}'^{(1)}_{(n-1)}\right\|<\epsilon$, upon which the final position is propagated down to the next finer pyramid level $\vec{x}^{(2)}_{(0)}\gets2\vec{x}'^{(1)}_{(N)}$, and the process repeats until $\vec{x}'^{(L)}_{(N)}$. \\\\ Again central to this algorithm is the warping operation: as in \hyperref[sec:2.2.5]{pyramidal Lucas-Kanade}, we compute polynomial coefficients for $I_2$ at warped coordinates, in order to estimate only residual displacements (instead of total displacement), via bilinear interpolation, as defined in Eq.~\eqref{eq:interp}:
\begin{equation}
\begin{aligned}
    \vec{A}_2\left(\vec{y}+(\vec{x}'^{(\ell)}_{(n)}-\vec{x}^{(\ell)})\right) &= \operatorname{INTERP}\left(\vec{y}+(\vec{x}'^{(\ell)}_{(n-1)}-\vec{x}^{(\ell)});\vec{A}_2^{(\ell)}\right) \\ \vec{b}_2\left(\vec{y}+(\vec{x}'^{(\ell)}_{(n)}-\vec{x}^{(\ell)})\right) &= \operatorname{INTERP}\left(\vec{y}+(\vec{x}'^{(\ell)}_{(n-1)}-\vec{x}^{(\ell)});\vec{b}_2^{(\ell)}\right)
\end{aligned}
\end{equation}

The complete Pyramidal Farnebäck framework is given by Algorithm~\ref{alg:fb} in Appendix~\ref{sec:a}.

\subsection{Kalman Filtering}
\label{sec:2.4}
Our probabilistic motion framework will treat optical flow vectors as random variables and estimate their distributions. The Kalman Filter (KF) \cite{Kalman1960} provides such a framework for linear dynamical systems. It assumes a linear Gaussian state-space model, under which the posterior distribution at each timestep $t$ is characterized fully by its mean $\vec{x}_t$ and covariance $\vec{\Sigma}_t$. The filter yields the optimal mean-square error estimate through a recursive procedure: a prediction step propagating the state distributions through the system dynamics, and an update step incorporating the current measurement $\vec{z}_t$. For nonlinear measurement models, the Extended Kalman Filter (EKF) \cite{Jazwinski1970} and its iterated variant (IEKF) \cite{Bell1993} approximate the Gaussian posterior via linearization.

\subsubsection{Kalman Filter}
\label{sec:2.4.1}
Consider a discrete-time linear dynamical system with state vector $\vec{x}_t\in\RR^{d_x}$ at time $t$. This state evolves via a linear transition model, and produces measurements $\vec{z}_t\in\RR^{d_z}$ via a linear measurement model:
\begin{equation}
    \vec{x}_t=\vec{F}_t\vec{x}_{t-1}+\vec{w}_t,\quad\vec{w}_t\sim\mathcal{N}(\vec{0},\vec{Q}_t)
\end{equation}
\begin{equation}
    \vec{z}_t=\vec{H}_t\vec{x}_t+\vec{v}_t,\quad\vec{v}_t\sim\mathcal{N}(\vec{0},\vec{R}_t)
\end{equation}

Here we have transition matrix $\vec{F}_t\in\RR^{d_x\times d_x}$, measurement matrix $\vec{H}_t\in\RR^{d_z\times d_x}$, and independent zero-mean Gaussian noise processes $\vec{w}_t\in\RR^{d_x},\vec{v}_t\in\RR^{d_z}$ with respective covariances $\vec{Q}_t,\vec{R}_t$. \\\\ The \textit{prediction step} propagates the previous timestep's posterior mean $\vec{x}_{t-1}$ and covariance $\vec{\Sigma}_{t-1}$ through the transition model to produce the current timestep's prior mean and covariance:
\begin{equation}
\begin{aligned}
    \vec{x}_{t|t-1} &= \vec{F}_t\vec{x}_{t-1} \\ \vec{\Sigma}_{t|t-1} &= \vec{F}_t\vec{\Sigma}_{t-1}\vec{F}_t^T+\vec{Q}_{t}
\end{aligned}
\end{equation}

The \textit{update step} fuses these priors with measurement $\vec{z}_t$. It is commonly presented in \textit{covariance form}, which operates directly on state $\vec{x}_t$ and covariance $\vec{\Sigma}_t$, and requires inverting a $d_z\times d_z$ matrix. We instead present its dual \textit{information form} \cite{Mutambara1998}, which operates on \textit{information matrix} $\vec{Y}_t=\vec{\Sigma}_t^{-1}$ and \textit{information vector} $\vec{y}_t=\vec{Y}_t\vec{x}_t$, and performs $d_x\times d_x$ inversions, as our framework will have $d_x\ll d_z$. We construct $\vec{Y}_t,\vec{y}_t$ using our priors:
\begin{equation}
\begin{aligned}
    \vec{Y}_t &= \vec{\Sigma}_{t|t-1}^{-1} \\ \vec{y}_t &= \vec{Y}_t\vec{x}_{t|t-1}
\end{aligned}
\end{equation}

The posterior information matrix and vector are obtained by adding the measurement's contribution:
\begin{equation}
\label{eq:kfbal}
\begin{aligned}
\vec{Y}_t' &= \vec{Y}_t+\vec{H}_t^T\vec{R}_t^{-1}\vec{H}_t \\ \vec{y}_t' &= \vec{y}_t+\vec{H}_t^T\vec{R}_t^{-1}\vec{z}_t
\end{aligned}
\end{equation}

The posterior mean and covariance are then recovered:
\begin{equation}
\begin{aligned}
    \vec{\Sigma}_t &= \left(\vec{Y}_t'\right)^{-1} \\ \vec{x}_t &= \vec{\Sigma}_t\vec{y}_t'
\end{aligned}
\end{equation}

\subsubsection{Extended Kalman Filter}
\label{sec:2.4.2}
Consider a linear dynamical system with a general (nonlinear) measurement model:
\begin{equation}
    \vec{x}_t=\vec{F}_t\vec{x}_{t-1}+\vec{w}_t,\quad\vec{w}_t\sim\mathcal{N}(\vec{0},\vec{Q}_t)
\end{equation}
\begin{equation}
    \vec{z}_t=\vec{h}(\vec{x}_t)+\vec{v}_t,\quad\vec{v}_t\sim\mathcal{N}(\vec{0},\vec{R}_t)
\end{equation}

The prediction step is unchanged, but the update step becomes invalid. The Extended Kalman Filter (EKF) addresses this by linearizing the measurement function $\vec{h}$ via a first-order Taylor expansion around prior mean $\vec{x}_{t|t-1}$:
\begin{equation}
    \vec{h}(\vec{x}_t)\approx\vec{h}(\vec{x}_{t|t-1})+\underbrace{\left.\frac{\partial\vec{h}(\vec{x})}{\partial\vec{x}}\right|_{\vec{x}=\vec{x}_{t|t-1}}}_{\vec{H}_t}(\vec{x}_t-\vec{x}_{t|t-1})
\end{equation}

Substituting this into our measurement model yields:
\begin{equation}
    \vec{z}_t\approx\vec{h}(\vec{x}_{t|t-1})+\vec{H}_t(\vec{x}_t-\vec{x}_{t|t-1})+\vec{v}_t
\end{equation}

We rearrange to obtain an approximate linear measurement model:
\begin{equation}
    \underbrace{\vec{z}_t-\vec{h}(\vec{x}_{t|t-1})+\vec{H}_t\vec{x}_{t|t-1}}_{\vec{z}^{(t)}}\approx\vec{H}_t\vec{x}_t+\vec{v}_t
\end{equation}

The update step is then conducted using this linearized Jacobian $\vec{H}_t$ and \textit{pseudomeasurement} $\vec{z}^{(t)}$.

\subsubsection{Iterated Extended Kalman Filter}
\label{sec:2.4.3}
The EKF performs this linearized update once per timestep, at prior mean $\vec{x}_{t|t-1}$. However, when measurement function $\vec{h}$ exhibits strong curvature, or when prior covariance $\vec{\Sigma}_{t|t-1}$ is large, linearization error may be substantial and degrade estimation quality. The iterated EKF (IEKF) computes a sequence of increasingly accurate linearizations at successive posterior mean estimates. \\\\ The prediction step is unchanged. We begin our update step by initializing:
\begin{equation}
    \vec{x}^{(t)}_{(0)}=\vec{x}_{t|t-1}\qquad\vec{\Sigma}^{(t)}_{(0)}=\vec{\Sigma}_{t|t-1}
\end{equation}
\begin{equation}
    \vec{Y}^{(t)}_{(0)}=\left(\vec{\Sigma}^{(t)}_{(0)}\right)^{-1}\qquad\vec{y}^{(t)}_{(0)}=\vec{Y}^{(t)}_{(0)}\vec{x}^{(t)}_{(0)}
\end{equation}

At iteration $n$, we linearize our measurement model around our previous posterior mean estimate $\vec{x}^{(t)}_{(n-1)}$ to obtain Jacobian $\vec{H}^{(t)}_{(n)}$ and pseudomeasurement $\vec{z}^{(t)}_{(n)}$. We then add this pseudomeasurement's contribution to $\vec{Y}^{(t)}_{(0)},\vec{y}^{(t)}_{(0)}$ via Eq.~\eqref{eq:kfbal} to obtain $\vec{Y}^{(t)}_{(n)},\vec{y}^{(t)}_{(n)}$, before recovering $\vec{x}^{(t)}_{(n)},\vec{\Sigma}^{(t)}_{(n)}$. Iteration terminates upon either reaching maximum iterations $N$ or when the change in estimates $\left\|\vec{x}^{(t)}_{(n)}-\vec{x}^{(t)}_{(n-1)}\right\|_2$ falls below convergence tolerance $\epsilon$. \\\\ The IEKF may be viewed as a Gauss-Newton procedure for maximum a posteriori estimation. Each iteration linearizes the measurement function at the current mean estimate, producing a quadratic approximation to the log-posterior and moving the estimate toward a local optimum. \\\\ Critically, we perform an additional iteration ($n=N+1$) to exclusively update the posterior covariance estimate $\vec{\Sigma}^{(t)}_{(N+1)}$, but not the mean estimate, which remains at $\vec{x}^{(t)}_{(N)}$. This is because each covariance $\vec{\Sigma}^{(t)}_{(n)}$ is computed via Jacobian $\vec{H}^{(t)}_{(n)}$ from approximation about previous estimate $\vec{x}^{(t)}_{(n-1)}$: thus to compute the covariance at the final estimate $\vec{x}^{(t)}_{(N)}$ requires $\vec{H}^{(t)}_{(N+1)}$. \\\\ The complete IEKF framework is given by Algorithm~\ref{alg:iekf} in Appendix~\ref{sec:a}.

\section{Methodology}
\label{sec:3}
Building upon our now established mathematical foundations, we now discuss our novel probabilistic optical flow framework. This section discusses how optical flow estimation is an inherently uncertain procedure, and how classical algorithms like Lucas-Kanade and Farnebäck fail to address or quantify the key notion of confidence in estimation quality, then shows how a probabilistic view of motions as random variables allows for distributions over optical flows to naturally encode such notions, and how iterated extended Kalman filtering meshes naturally with pyramidal Lucas-Kanade and Farnebäck to produce uncertainty-aware motion algorithms, before describing these algorithms' major advantages over their classical counterparts, and giving their precise mathematical formulations.

\subsection{Motivation}
\label{sec:3.1}
Optical flow estimation is a fundamentally ill-posed inverse problem. Given two consecutive frames $I_1,I_2$, it seeks to recover the motion field that relates them: the brightness constancy assumption provides a single constraint per pixel $\vec{x}$ for two unknown flow components, rendering the system underdetermined and subject to the aperture problem. The Lucas-Kanade and Farnebäck methods address this ambiguity through spatial coherence, aggregating constraints from local neighborhood $\mathcal{W}_K(\vec{x})$ to obtain least-squares solution $\vec{f}(\vec{x})$. This solution, however, produces a point estimate, i.e. a single fixed vector $\vec{f}(\vec{x})$, with no information regarding its reliability.

The eigenvalues of the structure tensor (Lucas-Kanade: $\vec{\nabla I}(\vec{x})^T\vec{W}(\vec{x})\vec{\nabla I}(\vec{x})$ by Eq.~\eqref{eq:lklsq}, Farnebäck: $\vec{\mathcal{A}}(\vec{x})^T\vec{\mathcal{W}}(\vec{x})\vec{\mathcal{A}}(\vec{x})$  by Eq.~\eqref{eq:fblsq}) provide some indication of system conditioning, but they do not propagate through the tracking pipeline, nor do they provide a meaningful description of the quality of the resulting estimate itself. A pixel in a textured region may yield a well-conditioned tensor, but its estimate may be corrupted by brightness constancy violations (luminance changes due to shadows, specular reflections, camera exposure changes) and interpolation artifacts: conversely, a pixel in a one-dimensionally textured region with an ill-conditioned tensor may still produce an accurate estimate if its true motion aligns with the intensity gradient, i.e. is normal to the texture.

This lack of genuine quality quantification is particularly problematic in real-world tracking scenarios, vulnerable to countless sources of uncertainty: sensor readout noise, quantization errors, occlusions, non-Lambertian surfaces, camera focus blur, etc., all introduce stochastic perturbations in image intensities, tampering with estimated flows. Moreover, real-world image sequences rarely perfectly obey brightness constancy, many real-world objects without strong two-dimensional texture (even as simple as walls) reignite the aperture problem, and windows $\mathcal{W}_K(\vec{x})$ are bound to contain objects with contrastive motions, collapsing spatial coherence.

Classical algorithms like Lucas-Kanade and Farnebäck attempt to mitigate these challenges through robust weighting, pyramidal refinement, and meticulous parameter tuning. The fundamental issue ignored, however, is this: with so many failure modes, each spatially varying, data-dependent, and difficult to predict, \underline{motion models must believe in themselves}. Namely, they must be able to describe how confident they are in each estimate. Thus deterministic estimation alone is insufficient: a point estimate $\vec{f}(\vec{x})$ is accompanied by no confidence information. Such algorithms return this single vector, leaving downstream users to infer reliability from structure tensor eigenvalues and other indirect heuristics, at best proxies and at worst misleading. What is ideal is an intuitive, interpretable confidence measure, that captures the combined effects of all sources of error, quantifies the uncertainty inherent in each resultant estimate, and propagates naturally through tracking algorithms.

Our novel optical flow framework adopts a probabilistic perspective, treating each flow as a random variable rather than a deterministic quantity. Specifically, we assume it to be distributed according to a Gaussian: $\widetilde{\vec{f}(\vec{x})}\sim\mathcal{N}(\vec{f}(\vec{x}),\vec{\Sigma}(\vec{x}))$. Notice we now denote this flow as $\widetilde{\vec{f}(\vec{x})}$ (largely unimportant), and use $\vec{f}(\vec{x})$ to represent its mean: our framework will estimate these Gaussian parameters $\vec{f}(\vec{x}),\vec{\Sigma}(\vec{x})$, to describe the posterior distribution for $\widetilde{\vec{f}(\vec{x})}$. As it is Gaussian, the maximum a posteriori estimate for $\widetilde{\vec{f}(\vec{x})}$ is $\vec{f}(\vec{x})$ itself. Thus we will be chiefly concerned with the estimation of mean $\vec{f}(\vec{x})$, which we denote as such to remain consistent with previous discussions of estimation.

Of central importance is covariance $\vec{\Sigma}(\vec{x})$, which satisfies precisely our requirements of a confidence measure. First, it intuitively quantifies an algorithm's confidence in its estimated mean $\vec{f}(\vec{x})$ as the concentration of density $p\left(\widetilde{\vec{f}(\vec{x})}\mid\vec{f}(\vec{x}),\vec{\Sigma}(\vec{x})\right)$ around $\vec{f}(\vec{x})$: a small $\vec{\Sigma}(\vec{x})$, i.e. a highly concentrated density, indicates the true $\widetilde{\vec{f}(\vec{x})}$ is likely close to $\vec{f}(\vec{x})$, thus the algorithm is ``more confident" in $\vec{f}(\vec{x})$, while a large $\vec{\Sigma}(\vec{x})$, i.e. a diffuse density, indicates a larger chance for $\widetilde{\vec{f}(\vec{x})}$ to deviate significantly from $\vec{f}(\vec{x})$, thus the algorithm is ``less confident" in it.

Second, it is directly interpretable: its eigenvalues and eigenvectors describe the magnitude and orientation of the ``uncertainty ellipsoid" in flow space: suppose a Lucas-Kanade-like algorithm, which produces $\vec{f}(\vec{x}),\vec{\Sigma}(\vec{x})$, encounters a region with strong one-dimensional texture, where (recall from section~\ref{sec:2.2.3}) only the component of $\vec{f}(\vec{x})$ normal to the texture is stably computable, while the component parallel is ill-conditioned: then the dominant eigenvector (corresponding to the direction of maximal uncertainty) will be near-parallel to the texture, with a large eigenvalue, while the minor eigenvector (corresponding to the less uncertain, i.e. more confident direction) will be near-orthogonal to the texture, with a small eigenvalue. Third, by employing probabilistic modeling, we inherently capture all sources of estimation error (noise, artifacts, etc.) through $\vec{\Sigma}(\vec{x})$. Fourth, $\vec{\Sigma}(\vec{x})$ propagates through tracking algorithms via Kalman filtering, as described in Section~\ref{sec:2.4}.

Namely, we discuss two novel algorithms, which respectively combine iterated extended Kalman Filters (IEKFs) with Lucas-Kanade- and Farnebäck-inspired measurement models: the former employs intensity derivatives $I(\vec{x}),\nabla I(\vec{x}),\nabla^2I(\vec{x})$, while the latter employs polynomial coefficients $c(\vec{x}),\vec{b}(\vec{x}),\vec{A}(\vec{x})$. The IEKF here is a natural framework: its recursive state estimation across timesteps $t=1,\cdots,T$ aligns with recursive flow estimation across \hyperref[eq:pyr]{Gaussian image pyramid} levels $\ell=1,\cdots,L$, its iterative linearization accommodates nonlinear measurement models, it soundly propagates estimated mean $\vec{f}(\vec{x})^{(\ell)}$ and covariance $\vec{\Sigma}(\vec{x})^{(\ell)}$, allowing covariances to reflect changing confidences in updating means, and it will require only $2\times 2$ matrix inversions, rendering it efficient even when accumulating constraints from many pixels.

In addition to confidence quantification, by employing Kalman filtering, our framework is equipped with another major advantage: a principled mechanism for balancing prior predictions for $\vec{f}(\vec{x}),\vec{\Sigma}(\vec{x})$ with adjustments suggested by measurement data. Recall Eq.~\eqref{eq:kfbal} (or Algorithm~\ref{alg:iekf} lines 9-10), where the posterior information matrix and vector $\vec{Y}_t',\vec{y}_t'$ are the sums of their prior counterparts $\vec{Y}_t,\vec{y}_t$ with measurement contributions $\vec{H}_t^T\vec{R}_t^{-1}\vec{H}_t$ and $\vec{H}_t^T\vec{R}_t^{-1}\vec{z}_t$. We will soon see precisely how our framework defines these $\vec{H}_t,\vec{R}_t,\vec{z}_t$: universally, for pixels where intensities provide strong signal, these measurement contributions will outweigh $\vec{Y}_t,\vec{y}_t$, while for pixels where intensities fail (akin to Lucas-Kanade at textureless regions), these contributions will be near zero. Instead of being unable to provide stable estimates, however, our $\vec{Y}_t',\vec{y}_t'$ will ``default" to $\vec{Y}_t,\vec{y}_t$. Namely, \underline{Kalman filtering balances priors and data}, a behavior impossible in deterministic algorithms, where no priors are present and only data is available. Thus our framework is adaptive, able to ``adjust" according to the quality of incoming measurement data, enabling robust motion tracking in real-world scenarios, where data quality varies dramatically across space and time.

\subsection{Probabilistic Lucas-Kanade}
\label{sec:3.2}

\subsubsection{Intensity Model}
\label{sec:3.2.1}
Recall from Eqs.~\eqref{eq:lkmat},~\eqref{eq:lklsq} that Lucas-Kanade computes flow $\vec{f}(\vec{x})$ by minimizing a weighted sum of squared \hyperref[eq:ofce]{OFCE} residuals:
\begin{equation}
    \vec{f}(\vec{x})=\underset{\vec{f}(\vec{x})}{\arg\min}\sum_{\vec{y}\in\mathcal{W}_K(\vec{x})}w_{\sigma}(\vec{y}-\vec{x})[\nabla I(\vec{y})^T\vec{f}(\vec{x})-I_t(\vec{y})]^2
\end{equation}

This formulation views them as deterministic values to be minimized. Our first method, which we name \textit{probabilistic Lucas-Kanade (PLK)-Intensity}, composes brightness constancy (\hyperref[eq:bce]{BCE}) residuals to form a Kalman measurement function. Namely, for mean $\vec{f}(\vec{x})$ of $\widetilde{\vec{f}(\vec{x})}$, as in Section~\ref{sec:3.1}, we propose the following Kalman transition and measurement models:
\begin{equation}
\label{eq:plktran}
    \vec{f}(\vec{x})^{(\ell)}=2\vec{f}(\vec{x})^{(\ell-1)}+\vec{w}(\vec{x})^{(\ell)},\quad\vec{w}(\vec{x})^{(\ell)}\sim\mathcal{N}(\vec{0},\omega\vec{I}_2)
\end{equation}
\begin{equation}
\label{eq:plkmeas}
    \vec{0}=\underbrace{\begin{bmatrix}I_2^{(\ell)}\left(\vec{x}_1+\vec{f}(\vec{x})^{(\ell)}\right)-I_1^{(\ell)}(\vec{x}_1)\\\vdots\\I_2^{(\ell)}\left(\vec{x}_{K^2}+\vec{f}(\vec{x})^{(\ell)}\right)-I_1^{(\ell)}(\vec{x}_{K^2})\end{bmatrix}}_{\vec{h}\left(\vec{f}(\vec{x})^{(\ell)}\right)\in\RR^{K^2}}+\vec{v}(\vec{x})^{(\ell)},\quad\vec{v}(\vec{x})^{(\ell)}\sim\mathcal{N}(\vec{0},\vec{R}(\vec{x})^{(\ell)})
\end{equation}

Whereas Kalman filtering is traditionally applied to estimate states transitioning across timesteps $t$, here our flows $\vec{f}(\vec{x})^{(\ell)}$ propagate through Gaussian image pyramid levels $\ell$.

Our transition model captures the relationship between flow estimates at consecutive pyramid levels. Recall from Section~\ref{sec:2.2.5} that each pyramid level is obtained by subsampling the image by a factor of 2: our coarser level's posterior flow estimate of $\vec{f}(\vec{x})^{(\ell-1)}$ pixels thus corresponds to $2\vec{f}(\vec{x})^{(\ell-1)}$ pixels at our finer level, to which $\vec{f}(\vec{x})^{(\ell-1)}$ propagates as the prior prediction for $\vec{f}(\vec{x})^{(\ell)}$. This transition, however, is inherently imperfect, $\vec{f}(\vec{x})^{(\ell-1)}$ having been computed as a mere approximation for true flow $\widetilde{\vec{f}(\vec{x})}$ at final level $L$, from lower-resolution versions of images $I_1,I_2$ necessarily lacking fine details (e.g. high-frequency textures, sub-pixel motions, etc.) for which $\vec{f}(\vec{x})^{(\ell-1)}$ cannot have accounted. Moreover, recall from Section~\ref{sec:2.2.5} that all levels $\ell=1,\cdots,L$ share window size $K$: thus larger windows at coarser levels are likely to capture distinct objects with contrastive motions, resulting in estimates $\vec{f}(\vec{x})^{(\ell-1)}$ as averaged aggregates irrepresentative of any ``true" motion, and not serving as meaningful predictions for estimates $\vec{f}(\vec{x})^{(\ell)}$ at finer levels, whose windows, now smaller, allow for different motion patterns to be more properly separated.

This justifies the introduction of transition noise $\vec{w}(\vec{x})^{(\ell)}$, which we model as an isotropic zero-mean Gaussian with covariance $\omega\vec{I}_2$. First, we have no reason to expect propagation $\vec{f}(\vec{x})^{(\ell-1)}\to\vec{f}(\vec{x})^{(\ell)}$ to be systematically biased in any particular direction, thus to reduce complexity we employ an isotropic covariance. Second, we have no reason to expect this propagation to be systematically more susceptible to error at any particular level $\ell$ or pixel $\vec{x}$ given that each level is constructed by an identical subsampling procedure, and that objects, i.e. motion patterns, are expected a priori to exist at all scales and locations in our original images, thus to further simplify we employ a single level- and input-independent scalar $\omega$. This parameter may be set empirically: a larger $\omega$ ``allows for more deviation" in estimate upsampling, indicating distrust in coarser-level signals, and is natural if images $I_1,I_2$ contain ample high-frequency textures that are lost during upsampling, or if several adjacent contrastive motions exist that cause meaningless aggregates, etc., while a smaller $\omega$ indicates a trustworthy Gaussian image pyramid construction, and is natural for $I_1,I_2$ with weak or low-frequency textures and smooth or coherent motions.

A subtle but important distinction concerns the interpretation of our measurement model. In the standard Kalman filtering framework, measurement model $\vec{z}_t=\vec{h}(\vec{x}_t)+\vec{v}_t$ posits a relationship between $\vec{x}_t$ and $\vec{z}_t$ through $\vec{h}$: it computes $\vec{h}(\vec{x})$ as the expected value of $\vec{z}_t$, then considers its actual, externally observed, value $\vec{z}_t$, existent and deduced independently of $\vec{x}_t$, before attributing any discrepancy between the two to noise $\vec{v}_t$. Our measurement, however, is the collection of brightness constancy residuals across $\mathcal{W}_K(\vec{x})$: its expected value is $\vec{z}^{(\ell)}\triangleq\vec{0}$, while its observed value is $\vec{h}\left(\vec{f}(\vec{x})^{(\ell)}\right)$. Namely, we are required to reverse our two terms: in standard Kalman filtering, expected $\vec{h}(\vec{x}_t)$ depends on state $\vec{x}_t$, while observed $\vec{z}_t$ is independent of $\vec{x}_t$, while in our model, expected residual $\vec{0}$ is independent of $\vec{f}(\vec{x})^{(\ell)}$, while observed residuals $\vec{h}\left(\vec{f}(\vec{x})^{(\ell)}\right)$ depends on $\vec{f}(\vec{x})^{(\ell)}$. Theoretically, our construction slightly abuses notation: the Kalman measurement model represents likelihood $p(\vec{z}_t\mid\vec{x}_t)=\mathcal{N}(\vec{z}_t\mid\vec{h}(\vec{x}_t),\vec{R}_t)$, a distribution for $\vec{z}_t$, thus it is valid only when $\vec{z}_t$ is a random quantity. Practically, however, because observation $\vec{z}_t$ is treated as a constant once measured, in order to construct posterior $p\left(\vec{x}_t\mid(\vec{z}_t)_{t=1}^T\right)\propto p(\vec{z}_t\mid\vec{x}_t)p\left(\vec{x}_t\mid(\vec{z}_t)_{t=1}^{T-1}\right)$ as the Gauss-Newton optimization objective, as discussed in Section~\ref{sec:2.4.3}, using constant $\vec{0}$ is qualitatively identical. Thus $\vec{h}$, our sole function of $\vec{f}(\vec{x})^{(\ell)}$, is our most natural measurement function.

We have defined our measurement model via brightness constancy residuals, first in order to encode classical Lucas-Kanade's brightness constancy assumption, and second because linearization of $\vec{h}\left(\vec{f}(\vec{x})^{(\ell)}\right)$ will produce the intensity gradients with whose usage and interpretation we have become familiar. Our model attributes any deviation of observed brightness constancy residuals $\vec{h}\left(\vec{f}(\vec{x})^{(\ell)}\right)\in\RR^{K^2}$ from their expectation $\vec{0}$ to random measurement noise $\vec{v}(\vec{x})^{(\ell)}$, which we model as a zero-mean Gaussian with covariance $\vec{R}(\vec{x})^{(\ell)}$. Notice that, as opposed to our previous $\omega\vec{I}_2$, this noise covariance is, as we will justify in Section~\ref{sec:3.2.3}, level- and input-dependent, i.e. anisotropic.

Finally, as we have seen in Sections~\ref{sec:2.2.5} and ~\ref{sec:2.3.4}, because $\vec{f}(\vec{x})^{(\ell)}$ is generally non-integer, we compute $\vec{h}\left(\vec{f}(\vec{x})^{(\ell)}\right)$ via bilinear interpolation, as defined in Eq.~\eqref{eq:interp}:
\begin{equation}
\label{eq:plkitemp}
    I_2^{(\ell)}(\vec{y}+\vec{f}(\vec{x}))=\operatorname{INTERP}\left(\vec{y}+\vec{f}(\vec{x});I_2^{(\ell)}\right)
\end{equation}

\subsubsection{Measurement Linearization}
\label{sec:3.2.2}
To apply iteration $n$ of our IEKF update step, as in Section~\ref{sec:2.4.3}, we linearize our measurement function $\vec{h}$ with respect to previous state estimate $\vec{f}(\vec{x})^{(\ell)}_{(n-1)}$. Differentiating the $i$th component of $\vec{h}$ gives:
\begin{equation}
\begin{aligned}
    \left.\frac{\partial h(\vec{f}(\vec{x}))_i}{\partial\vec{f}(\vec{x})}\right|_{\vec{f}(\vec{x})=\vec{f}(\vec{x})^{(\ell)}_{(n-1)}} &= \left.\frac{\partial}{\partial\vec{f}(\vec{x})}\left[I_2^{(\ell)}(\vec{x}_i+\vec{f}(\vec{x}))-I_1(\vec{x}_i)\right]\right|_{\vec{f}(\vec{x})=\vec{f}(\vec{x})^{(\ell)}_{(n-1)}} \\&= \left.\left(\frac{\partial}{\partial u(\vec{x})}I_2^{(\ell)}(\vec{x}_i+\vec{f}(\vec{x})),\frac{\partial}{\partial v(\vec{x})}I_2^{(\ell)}(\vec{x}_i+\vec{f}(\vec{x}))\right)\right|_{\vec{f}(\vec{x})=\vec{f}(\vec{x})^{(\ell)}_{(n-1)}} \\&= \left(I_{2x}^{(\ell)}\left(\vec{x}_i+\vec{f}(\vec{x})^{(\ell)}_{(n-1)}\right),I_{2y}^{(\ell)}\left(\vec{x}_i+\vec{f}(\vec{x})^{(\ell)}_{(n-1)}\right)\right) \\&= \nabla I_2^{(\ell)}\left(\vec{x}_i+\vec{f}(\vec{x})^{(\ell)}_{(n-1)}\right)^T
\end{aligned}
\end{equation}

Here $I_{2x},I_{2y}$ are shorthand for $(I_2)_x,(I_2)_y$, the horizontal and vertical gradient components of the second image. Specifically, we have these gradients evaluated at warped coordinates $\vec{x}_i+\vec{f}(\vec{x})^{(\ell)}_{(n-1)}$:
\begin{equation}
\label{eq:plkispat}
\begin{aligned}
    I_{2x}^{(\ell)}(\vec{x}_i+\vec{f}(\vec{x})) &= \operatorname{INTERP}\left(\vec{x}_i+\vec{f}(\vec{x});I_{2x}^{(\ell)}\right) \\ I_{2y}^{(\ell)}(\vec{x}_i+\vec{f}(\vec{x})) &= \operatorname{INTERP}\left(\vec{x}_i+\vec{f}(\vec{x});I_{2y}^{(\ell)}\right)
\end{aligned}
\end{equation}

Stacking these gradients over the full window $\mathcal{W}_K(\vec{x})$, i.e. all components of $\vec{h}$, gives our measurement Jacobian, $\vec{H}(\vec{x})^{(\ell)}_{(n)}\in\RR^{K^2\times 2}$:
\begin{equation}
\begin{aligned}
    \vec{H}(\vec{x})^{(\ell)}_{(n)} &= \left.\frac{\partial\vec{h}(\vec{f}(\vec{x}))}{\partial\vec{f}(\vec{x})}\right|_{\vec{f}(\vec{x})=\vec{f}(\vec{x})^{(\ell)}_{(n-1)}} \\&= \begin{bmatrix}I_{2x}^{(\ell)}\left(\vec{x}_1+\vec{f}(\vec{x})^{(\ell)}_{(n-1)}\right)&I_{2y}^{(\ell)}\left(\vec{x}_1+\vec{f}(\vec{x})^{(\ell)}_{(n-1)}\right)\\\vdots&\vdots\\I_{2x}^{(\ell)}\left(\vec{x}_{K^2}+\vec{f}(\vec{x})^{(\ell)}_{(n-1)}\right)&I_{2y}^{(\ell)}\left(\vec{x}_{K^2}+\vec{f}(\vec{x})^{(\ell)}_{(n-1)}\right)\end{bmatrix} \\&= \begin{bmatrix}\nabla I_2^{(\ell)}\left(\vec{x}_1+\vec{f}(\vec{x})^{(\ell)}_{(n-1)}\right)^T\\\vdots\\\nabla I_2^{(\ell)}\left(\vec{x}_{K^2}+\vec{f}(\vec{x})^{(\ell)}_{(n-1)}\right)^T\end{bmatrix}
\end{aligned}
\end{equation}

We have encountered our familiar spatial intensity gradients, though with a key distinction. Classical Lucas-Kanade, for each level $\ell$, computes its spatial gradients on first image $I_1$ at fixed window coordinates $\vec{y}\in\mathcal{W}_K(\vec{x})$, as in Algorithm~\ref{alg:lk} line 12, and holds them constant throughout intra-level iterations $n=1,\cdots,N$, a pragmatic choice which avoids repetitive recomputation, whereas our Jacobian gradients, by nature of our measurement model construction, must be computed on second image $I_2$, and recomputed with each iteration $n$ at warped coordinates $\vec{y}+\vec{f}(\vec{x})^{(\ell)}_{(n-1)}$. In practice, to improve robustness against slight intensity changes, we average gradients across both image frames:
\begin{equation}
\label{eq:plkigrad}
    \nabla I^{(\ell)}(\vec{y};\vec{f}(\vec{x}))\triangleq\frac{1}{2}\left(\nabla I_1^{(\ell)}(\vec{y})+\nabla I_2^{(\ell)}(\vec{y}+\vec{f}(\vec{x}))\right)
\end{equation}

With our Jacobian, we perform our measurement linearization:
\begin{equation}
    \vec{h}\left(\vec{f}(\vec{x})^{(\ell)}_{(n)}\right)\approx\vec{h}\left(\vec{f}(\vec{x})^{(\ell)}_{(n-1)}\right)+\vec{H}(\vec{x})^{(\ell)}_{(n)}\left(\vec{f}(\vec{x})^{(\ell)}_{(n)}-\vec{f}(\vec{x})^{(\ell)}_{(n-1)}\right)
\end{equation}

Substituting this into our measurement model and rearranging yields:
\begin{equation}
    -\vec{h}\left(\vec{f}(\vec{x})^{(\ell)}_{(n-1)}\right)+\vec{H}(\vec{x})^{(\ell)}_{(n)}\vec{f}(\vec{x})^{(\ell)}_{(n-1)}\approx\vec{H}(\vec{x})^{(\ell)}_{(n)}\vec{f}(\vec{x})^{(\ell)}_{(n)}+\vec{v}^{(\ell)}
\end{equation}

We have produced an approximate linear measurement model for our current iterate. As in Section~\ref{sec:2.4.2} (or Algorithm~\ref{alg:iekf} line 8), this left-hand side becomes our pseudomeasurement $\vec{z}(\vec{x})^{(\ell)}_{(n)}\in\RR^{K^2}$, given explicitly by:
\begin{equation}
\label{eq:plkimeas}
\begin{aligned}
    \vec{z}(\vec{x})^{(\ell)}_{(n)} &= \begin{bmatrix}I_1^{(\ell)}(\vec{x}_1)-I_2^{(\ell)}\left(\vec{x}_1+\vec{f}(\vec{x})^{(\ell)}_{(n-1)}\right)\\\vdots\\I_1^{(\ell)}(\vec{x}_{K^2})-I_2^{(\ell)}\left(\vec{x}_{K^2}+\vec{f}(\vec{x})^{(\ell)}_{(n-1)}\right)\end{bmatrix} \\&+ \begin{bmatrix}\nabla I^{(\ell)}\left(\vec{x}_1;\vec{f}(\vec{x})^{(\ell)}_{(n-1)}\right)^T\vec{f}(\vec{x})^{(\ell)}_{(n-1)}\\\vdots\\\nabla I^{(\ell)}\left(\vec{x}_{K^2};\vec{f}(\vec{x})^{(\ell)}_{(n-1)}\right)^T\vec{f}(\vec{x})^{(\ell)}_{(n-1)}\end{bmatrix}
\end{aligned}
\end{equation}

\subsubsection{Noise Covariance Estimation}
\label{sec:3.2.3}
Of chief relevance to the design of a truly adaptive IEKF framework is the estimation of measurement noise covariance $\vec{R}(\vec{x})^{(\ell)}$ in Eq.~\eqref{eq:plkmeas}. Lucas-Kanade treats all pixels in window $\mathcal{W}_K(\vec{x})$ as equally reliable a priori, which ignores that the difficulties of flow estimation---as described in Section~\ref{sec:3.1}, and originally motivating probabilistic considerations altogether---exhibit different behaviors at different pixels. An object captured in $\mathcal{W}_K(\vec{x})$ may become only partially shaded or occluded by another object, and we should trust more closely those pixels unaffected. Motion blur by a foreground object prompts us to focus consideration on more stable background pixels, and sensor noise certainly will vary, often drastically, from each pixel to its neighbors. In all these cases, intensity gradients may remain high: such phenomena motivate an additional data-driven mechanism to evaluate each pixel's trustworthiness. Thus, we dynamically assign, for each pixel, a noise variance based directly in its measurement residual:
\begin{equation}
\label{eq:plkivar}
    \sigma_R(\vec{y})^{(\ell)}_{(n)}\triangleq\max\left(\left|I_1^{(\ell)}(\vec{y})-I_2^{(\ell)}\left(\vec{y}+\vec{f}(\vec{x})^{(\ell)}_{(n-1)}\right)\right|^{\gamma},10^{-6}\right)
\end{equation}

A pixel $\vec{y}$ exhibiting a large intensity difference after warping is a large violation of brightness constancy, our entire measurement model's foundational assumption, prompting its ``contribution" to the posterior information data terms to be discarded---recall from Eq.~\eqref{eq:kfbal} that these terms are, in simplified notation, $\vec{H}^T\vec{R}^{-1}\vec{H}$ and $\vec{H}^T\vec{R}^{-1}\vec{z}$. Our assignment $\sigma_R(\vec{y})$, to such a pixel is then large, meaning when inverted in $\vec{R}^{-1}$, its contribution will be small. This effectively masks out massive-violation pixels, preventing them from poisoning posterior estimates. Conversely, a small residual, i.e. adherence to brightness constancy, allows a pixel, through its large $\sigma_R(\vec{y})^{-1}$, to strongly inform the update. To prevent division by zero for pixels with perfect brightness constancy, we maintain a uniform lower bound on $\sigma_R(\vec{y})$ of $10^{-6}$: tuning this specific value has negligible effect on results.

We generalize our residual by introducing exponent $\gamma$, allowing manual tuning of the aggressiveness of our penalty. Setting $\gamma=0$ reduces noise covariance to its classical, non-adaptive (though still weighted) formulation. $\gamma=1$ penalizes minor blurs and noise, but is often insufficient to suppress massive violations by occlusions and specular reflections. $\gamma\ge2$ aggressively punishes and rejects pixels with large residuals, but may risk discarding valid measurements with minor deviations.

The precise choice of $\gamma$ is highly influential on posterior results, and its optimal setting is highly image-dependent. For scenes containing rigid bodies traversing across sharp, high-frequency occlusion boundaries, an aggressive $\gamma$ excels, identifying and discarding occluded pixels while retaining their trackable counterparts. However, scenes with low-frequency lighting changes, atmospheric effects (e.g. fog), and complex non-rigid motions violate brightness constancy uniformly over large contiguous regions. In such cases, large $\gamma$ values uniformly amplify $\sigma_R(\vec{y})$ across $\vec{y}\in\mathcal{W}_K(\vec{x})$, risking our IEKF completely rejecting incoming image data for $\mathcal{W}_K(\vec{x})$ and defaulting exclusively to its priors. In particular, these priors are often uninformative in these highly dynamic scenes, causing estimates to stagnate and errors to compound. Thus it is not always detrimental to set $\gamma=0$, requiring our IEKF to duly consider its data. While the measurement noise covariance $\vec{R}$ is somewhat compromised, the now accepted intensity gradients $\vec{H}$ allow the IEKF to continue tracking overarching structural motions.

With each pixel's variance assigned, we assemble the complete $\vec{R}(\vec{x})^{(\ell)}\in\RR^{K^2\times K^2}$:
\begin{equation}
    \vec{R}(\vec{x})^{(\ell)}_{(n)}\triangleq\operatorname{diag}\left(\sigma_R(\vec{x}_1)^{(\ell)}_{(n)},\cdots,\sigma_R(\vec{x}_{K^2})^{(\ell)}_{(n)}\right)\odot\vec{W}(\vec{x})
\end{equation}

Here $\odot$ denotes element-wise multiplication. This construction employs one further assumption, that measurement noise at distinct pixels is independent. This is somewhat naïve, as shaded, occluded, and motion-blurred pixels typically occupy contiguous regions, but it allows us to massively simplify posterior information calculations, as detailed in Section~\ref{sec:3.2.4}, to ensure computational tractability. We then preserve Lucas-Kanade's weighting, as justified in Section~\ref{sec:2.2.4}, by scaling our diagonal noise covariance matrix by $\vec{W}(\vec{x})$, from Eq.~\eqref{eq:lkwgt}.

\subsubsection{Information Accumulation}
\label{sec:3.2.4}
Recall that per our update step, we must compute our posterior information matrix and vector, as in Eq.~\eqref{eq:kfbal} (or Algorithm~\ref{alg:iekf} lines 9-10):
\begin{equation}
\begin{aligned}
    \vec{Y}(\vec{x})^{(\ell)}_{(n)} &\gets \vec{Y}(\vec{x})^{(\ell)}_{(0)}+\left(\vec{H}(\vec{x})^{(\ell)}_{(n)}\right)^T\left(\vec{R}(\vec{x})^{(\ell)}_{(n)}\right)^{-1}\vec{H}(\vec{x})^{(\ell)}_{(n)} \\ \vec{y}(\vec{x})^{(\ell)}_{(n)} &\gets \vec{y}(\vec{x})^{(\ell)}_{(0)}+\left(\vec{H}(\vec{x})^{(\ell)}_{(n)}\right)^T\left(\vec{R}(\vec{x})^{(\ell)}_{(n)}\right)^{-1}\vec{z}(\vec{x})^{(\ell)}_{(n)}
\end{aligned}
\end{equation}

By now it becomes clear why in Section~\ref{sec:2.4.1} we chose the information-form presentation of Kalman filtering: when we then invert our posterior information matrix $\vec{Y}(\vec{x})^{(\ell)}_{(n)}$ to recover posterior covariance $\vec{\Sigma}(\vec{x})^{(\ell)}_{(n)}$, as in Section 2.4.1 (or Algorithm~\ref{alg:iekf} line 11), this inversion is of dimension:
\begin{equation}
    \operatorname{dim}\left(\left(\vec{H}(\vec{x})^{(\ell)}_{(n)}\right)^T\left(\vec{R}(\vec{x})^{(\ell)}_{(n)}\right)^{-1}\vec{H}(\vec{x})^{(\ell)}_{(n)}\right)=\left(2\times K^2\right)\left(K^2\times K^2\right)\left(K^2\times2\right)=\left(2\times2\right)
\end{equation}

Namely, this inversion is of state dimension $d_x\times d_x=2\times2$, whereas a standard-form procedure would require a much more computationally costly inversion of measurement dimension $d_z\times d_z=K^2\times K^2$, though it still requires the intermediate construction and inversion of $K^2\times K^2$ matrix $\vec{R}(\vec{x})^{(\ell)}_{(n)}$. We may avoid this expense, however, by exploiting this matrix's diagonal structure. \\\\ For general matrices $\vec{A}\in\RR^{M\times P},\vec{B}\in\RR^{P\times P},\vec{C}\in\RR^{P\times N}$, we may equate:
\begin{equation}
    (abc)_{mn}=\sum_{p=1}^P\sum_{p'=1}^Pa_{mp}b_{pp'}c_{p'n}\xlongequal{\text{$\vec{B}$ diagonal}}\sum_{p=1}^Pa_{mp}b_{pp}c_{pn}
\end{equation}

Thus, denoting column $p$ of $\vec{A}$ as $\vec{a}_p$, and row $p$ of $\vec{C}$ as $\vec{c}_p^T$, the full matrix product is given by:
\begin{equation}
    \vec{A}\vec{B}\vec{C}\xlongequal{\text{$\vec{B}$ diagonal}}\sum_{p=1}^Pb_{pp}\vec{a}_{p}\vec{c}_p^T
\end{equation}

Our accumulation has $\vec{A}=\left(\vec{H}(\vec{x})^{(\ell)}_{(n)}\right)^T,\vec{B}=\left(\vec{R}(\vec{x})^{(\ell)}_{(n)}\right)^{-1}$, and $\vec{C}$ either $\vec{H}(\vec{x})^{(\ell)}_{(n)}$ or $\vec{z}(\vec{x})^{(\ell)}_{(n)}$: the former case gives $\vec{a}_p\vec{B}_{pp}\vec{c}_p^T\in\RR^{2\times2}$, while the latter gives $\vec{a}_pc_p\in\RR^{2\times1}$. We have thus transformed our previous accumulation, computing a product involving a $K^2\times K^2$ matrix, into the accumulations below, requiring only $2\times2$ matrices or $2\times1$ vectors: designing diagonal measurement noise covariance reduces our accumulation's computation time complexity from $\mathcal{O}(K^6)$ to $\mathcal{O}(K^2)$. \\\\ Our posterior information matrix becomes:
\begin{equation}
\begin{gathered}
    \vec{Y}(\vec{x})^{(\ell)}_{(n)}\gets\vec{Y}(\vec{x})^{(\ell)}_{(0)}+\sum_{\vec{y}\in\mathcal{W}_K(\vec{x})}w_{\sigma}(\vec{y}-\vec{x})\left(\sigma_R(\vec{y})^{(\ell)}_{(n)}\right)^{-1}\aleph\\\aleph=\nabla I^{(\ell)}\left(\vec{y};\vec{f}(\vec{x})^{(\ell)}_{(n-1)}\right)\nabla I^{(\ell)}\left(\vec{y};\vec{f}(\vec{x})^{(\ell)}_{(n-1)}\right)^T
\end{gathered}
\end{equation}

Our posterior information vector becomes:
\begin{equation}
\begin{gathered}
    \vec{y}(\vec{x})^{(\ell)}_{(n)}\gets\vec{y}(\vec{x})^{(\ell)}_{(0)}+\sum_{\vec{y}\in\mathcal{W}_K(\vec{x})}w_{\sigma}(\vec{y}-\vec{x})\left(\sigma_R(\vec{y})^{(\ell)}_{(n)}\right)^{-1}\nabla I^{(\ell)}\left(\vec{y};\vec{f}(\vec{x})^{(\ell)}_{(n-1)}\right)\aleph\\\aleph=\left(I_1^{(\ell)}(\vec{y})-I_2^{(\ell)}\left(\vec{y}+\vec{f}(\vec{x})^{(\ell)}_{(n-1)}\right)+\nabla I^{(\ell)}\left(\vec{y};\vec{f}(\vec{x})^{(\ell)}_{(n-1)}\right)^T\vec{f}(\vec{x})^{(\ell)}_{(n-1)}\right)
\end{gathered}
\end{equation}

\subsubsection{Gradient Model}
\label{sec:3.2.5}
We have discussed previously in Section~\ref{sec:3.1} that, due to shadows, reflections, camera exposure changes, etc., the brightness constancy assumption, upon which Lucas-Kanade depends, is rarely perfectly valid. An alternative assumption, introduced by Brox et al. \cite{Brox2004}, is that of \textit{gradient constancy}: while an object's brightness may change frequently as it moves, rarely will its intrinsic shape and composition, i.e. its texture, edges, and corners, act likewise. Thus, rather than $I_1(\vec{x})=I_2(\vec{x}+\vec{f}(\vec{x}))$, we may assume:
\begin{equation}
    \nabla I_1(\vec{x})=\nabla I_2(\vec{x}+\vec{f}(\vec{x}))
\end{equation}

This assumption is intuitively somewhat superior, as it acts on objects' innate structures while being illumination-invariant: it focuses on high texture structures, which through their large gradients are precisely those regions most well-posed for motion recovery, while discounting flat, uniform regions, where brightness changes most commonly induce spurious motions. We thus introduce our second method, \textit{PLK-Gradient}, whose transition model is unchanged from its predecessor in Section~\ref{sec:3.2.1}, but whose measurement model uses our new assumption:
\begin{equation}
    \vec{f}(\vec{x})^{(\ell)}=2\vec{f}(\vec{x})^{(\ell-1)}+\vec{w}(\vec{x})^{(\ell)},\quad\vec{w}(\vec{x})^{(\ell)}\sim\mathcal{N}(\vec{0},\omega\vec{I}_2)
\end{equation}
\begin{equation}
    \vec{0}=\underbrace{\begin{bmatrix}\nabla I_2^{(\ell)}\left(\vec{x}_1+\vec{f}(\vec{x})^{(\ell)}\right)-\nabla I_1^{(\ell)}(\vec{x}_1)\\\vdots\\\nabla I_2^{(\ell)}\left(\vec{x}_{K^2}+\vec{f}(\vec{x})^{(\ell)}\right)-\nabla I_1^{(\ell)}(\vec{x}_{K^2})\end{bmatrix}}_{\vec{h}\left(\vec{f}(\vec{x})^{(\ell)}\right)\in\RR^{2K^2}}+\vec{v}(\vec{x})^{(\ell)},\quad\vec{v}(\vec{x})^{(\ell)}\sim\mathcal{N}(\vec{0},\vec{R}(\vec{x})^{(\ell)})
\end{equation}

The linearization, noise covariance estimation, and information accumulation of PLK-Gradient are entirely analogous to PLK-Intensity, with one major change: since our measurement consists of gradients, its Jacobian consists of image Hessians. Mirroring Eq.~\eqref{eq:plkitemp}, we compute $\vec{h}\left(\vec{f}(\vec{x})^{(\ell)}\right)$, via bilinear interpolation:
\begin{equation}
\label{eq:plkgtemp}
    \nabla I_2^{(\ell)}(\vec{y}+\vec{f}(\vec{x}))=\operatorname{INTERP}\left(\vec{y}+\vec{f}(\vec{x});\nabla I_2^{(\ell)}\right)
\end{equation}

To improve robustness, we average Hessians across both image frames, analogous to Farnebäck's averaging of $\vec{A}(\vec{x})$ in Section~\ref{sec:2.3.2}:
\begin{equation}
\label{eq:plkgspat}
\begin{gathered}
    I_{xx}(\vec{y};\vec{f}(\vec{x}))\triangleq\frac{1}{2}((I_1)_{xx}(\vec{y})+\operatorname{INTERP}(\vec{y}+\vec{f}(\vec{x});(I_2)_{xx}))\\I_{xy}(\vec{y};\vec{f}(\vec{x}))\triangleq\frac{1}{2}((I_1)_{xy}(\vec{y})+\operatorname{INTERP}(\vec{y}+\vec{f}(\vec{x});(I_2)_{xy}))\\I_{yy}(\vec{y};\vec{f}(\vec{x}))\triangleq\frac{1}{2}((I_1)_{yy}(\vec{y})+\operatorname{INTERP}(\vec{y}+\vec{f}(\vec{x});(I_2)_{yy}))
\end{gathered}
\end{equation}

Thus, denoting the averaged Hessian by $\nabla^2I$, our Jacobian $\vec{H}(\vec{x})^{(\ell)}_{(n)}\in\RR^{2K^2\times2}$ is given by:
\begin{equation}
\label{eq:plkggrad}
\begin{aligned}
    \vec{H}(\vec{x})^{(\ell)}_{(n)} &= \begin{bmatrix}I_{xx}^{(\ell)}\left(\vec{x}_1;\vec{f}(\vec{x})^{(\ell)}_{(n-1)}\right)&I_{xy}^{(\ell)}\left(\vec{x}_1;\vec{f}(\vec{x})^{(\ell)}_{(n-1)}\right)\\I_{xy}^{(\ell)}\left(\vec{x}_1;\vec{f}(\vec{x})^{(\ell)}_{(n-1)}\right)&I_{yy}^{(\ell)}\left(\vec{x}_1;\vec{f}(\vec{x})^{(\ell)}_{(n-1)}\right)\\\vdots&\vdots\\I_{xx}^{(\ell)}\left(\vec{x}_{K^2};\vec{f}(\vec{x})^{(\ell)}_{(n-1)}\right)&I_{xy}^{(\ell)}\left(\vec{x}_{K^2};\vec{f}(\vec{x})^{(\ell)}_{(n-1)}\right)\\I_{xy}^{(\ell)}\left(\vec{x}_{K^2};\vec{f}(\vec{x})^{(\ell)}_{(n-1)}\right)&I_{yy}^{(\ell)}\left(\vec{x}_{K^2};\vec{f}(\vec{x})^{(\ell)}_{(n-1)}\right)\end{bmatrix} \\&= \begin{bmatrix}\nabla^2I^{(\ell)}\left(\vec{x}_1;\vec{f}(\vec{x})^{(\ell)}_{(n-1)}\right)\\\vdots\\\nabla^2I^{(\ell)}\left(\vec{x}_{K^2};\vec{f}(\vec{x})^{(\ell)}_{(n-1)}\right)\end{bmatrix}
\end{aligned}
\end{equation}

From this follows, as in Section~\ref{sec:3.2.2}, our pseudomeasurement $\vec{z}(\vec{x})^{(\ell)}_{(n)}\in\RR^{2K^2}$:
\begin{equation}
\label{eq:plkgmeas}
\begin{aligned}
    \vec{z}(\vec{x})^{(\ell)}_{(n)} &= \begin{bmatrix}\nabla I_1^{(\ell)}(\vec{x}_1)-\nabla I_2^{(\ell)}\left(\vec{x}_1+\vec{f}(\vec{x})^{(\ell)}_{(n-1)}\right)\\\vdots\\\nabla I_1^{(\ell)}(\vec{x}_{K^2})-\nabla I_2^{(\ell)}\left(\vec{x}_{K^2}+\vec{f}(\vec{x})^{(\ell)}_{(n-1)}\right)\end{bmatrix} \\&+ \begin{bmatrix}\nabla^2I^{(\ell)}\left(\vec{x}_1;\vec{f}(\vec{x})^{(\ell)}_{(n-1)}\right)\vec{f}(\vec{x})^{(\ell)}_{(n-1)}\\\vdots\\\nabla^2I^{(\ell)}\left(\vec{x}_{K^2};\vec{f}(\vec{x})^{(\ell)}_{(n-1)}\right)\vec{f}(\vec{x})^{(\ell)}_{(n-1)}\end{bmatrix}
\end{aligned}
\end{equation}

We continue to use our noise covariance heuristic from Section~\ref{sec:3.2.3}, but now with gradient constancy residuals. Moreover, we assign separate, independent values for each horizontal and vertical gradient component, as we have seen texture is highly directional. If an object translates behind a vertical occlusion boundary, then horizontal gradient constancy is aggressively violated as texture appears or disappears, necessitating a high $\sigma_{Rx}$. However, vertical gradient constancy may remain perfectly valid, prompting a low $\sigma_{Ry}$. Thus we decouple $\sigma_{Rx}$ and $\sigma_{Ry}$, allowing the successful discarding of corrupted measurements along one direction, and the simultaneous extraction of valid information along the other:
\begin{equation}
\label{eq:plkgvar}
\begin{gathered}
    \sigma_{Rx}(\vec{y})^{(\ell)}_{(n)}\triangleq\max\left(\left|I_{1x}^{(\ell)}(\vec{y})-I_{2x}^{(\ell)}\left(\vec{y}+\vec{f}(\vec{x})^{(\ell)}_{(n-1)}\right)\right|^{\gamma},10^{-6}\right)\\\sigma_{Ry}(\vec{y})^{(\ell)}_{(n)}\triangleq\max\left(\left|I_{1y}^{(\ell)}(\vec{y})-I_{2y}^{(\ell)}\left(\vec{y}+\vec{f}(\vec{x})^{(\ell)}_{(n-1)}\right)\right|^{\gamma},10^{-6}\right)
\end{gathered}
\end{equation}

Recalling $\vec{\mathcal{W}}(\vec{x})$ from Eq.~\eqref{eq:fbwgt}, we form our measurement noise covariance:
\begin{equation}
    \vec{R}(\vec{x})^{(\ell)}_{(n)}\triangleq\operatorname{diag}\left(\sigma_{Rx}(\vec{x}_1)^{(\ell)}_{(n)},\sigma_{Ry}(\vec{x}_1)^{(\ell)}_{(n)},\cdots,\sigma_{Rx}(\vec{x}_{K^2})^{(\ell)}_{(n)},\sigma_{Ry}(\vec{x}_{K^2})^{(\ell)}_{(n)}\right)\odot\vec{\mathcal{W}}(\vec{x})
\end{equation}

Again $\odot$ denotes element-wise multiplication. Analogously to Section~\ref{sec:3.2.4}, we now exploit the diagonal structure of $\vec{R}(\vec{x})$. For general matrices $\vec{A}\in\RR^{2M\times2P},\vec{B}\in\RR^{2P\times2P},\vec{C}\in\RR^{2P\times2N}$, partitioned into $2\times2$ blocks $\vec{A}_{mp},\vec{B}_{pp'},\vec{C}_{p'n}\in\RR^{2\times2}$, we may equate:
\begin{equation}
    (\vec{A}\vec{B}\vec{C})_{mn}=\sum_{p=1}^P\sum_{p'=1}^P\vec{A}_{mp}\vec{B}_{pp'}\vec{C}_{p'n}\xlongequal{\text{$\vec{B}$ block-diagonal}}\sum_{p=1}^P\vec{A}_{mp}\vec{B}_{pp}\vec{C}_{pn}
\end{equation}

Thus, denoting block column $p$ of $\vec{A}$ as $\vec{A}_p\in\RR^{2M\times2}$, and block row $p$ of $\vec{C}$ as $\vec{C}_p\in\RR^{2\times2N}$, the full matrix product is given by:
\begin{equation}
    \vec{A}\vec{B}\vec{C}\xlongequal{\text{$\vec{B}$ block-diagonal}}\sum_{p=1}^P\vec{A}_p\vec{B}_{pp}\vec{C}_p
\end{equation}

Our accumulation has $\vec{A}=\left(\vec{H}(\vec{x})^{(\ell)}_{(n)}\right)^T,\vec{B}=\left(\vec{R}(\vec{x})^{(\ell)}_{(n)}\right)^{-1}$, and $\vec{C}$ either $\vec{H}(\vec{x})^{(\ell)}_{(n)}$ or $\vec{z}(\vec{x})^{(\ell)}_{(n)}$: the former case gives $\vec{A}_p\vec{B}_{pp}\vec{C}_p\in\RR^{2\times2}$, while the latter gives $\vec{A}_p\vec{B}_{pp}\vec{c}_p\in\RR^{2\times1}$. We have again reduced our accumulations in dimension from $2K^2\times2K^2$ to $2\times2$ and $2\times1$. \\\\ Our posterior information matrix becomes:
\begin{equation}
\begin{gathered}
    \vec{Y}(\vec{x})^{(\ell)}_{(n)}\gets\vec{Y}(\vec{x})^{(\ell)}_{(0)}+\sum_{\vec{y}\in\mathcal{W}_K(\vec{x})}w_{\sigma}(\vec{y}-\vec{x})\operatorname{diag}\left(\sigma_{Rx}(\vec{y})^{(\ell)}_{(n)},\sigma_{Ry}(\vec{y})^{(\ell)}_{(n)}\right)^{-1}\aleph\\\aleph=\nabla^2I^{(\ell)}\left(\vec{y};\vec{f}(\vec{x})^{(\ell)}_{(n-1)}\right)\nabla^2I^{(\ell)}\left(\vec{y};\vec{f}(\vec{x})^{(\ell)}_{(n-1)}\right)
\end{gathered}
\end{equation}

Our posterior information vector becomes:
\begin{equation}
\begin{gathered}
    \vec{y}(\vec{x})^{(\ell)}_{(n)}\gets\vec{y}(\vec{x})^{(\ell)}_{(0)}+\sum_{\vec{y}\in\mathcal{W}_K(\vec{x})}w_{\sigma}(\vec{y}-\vec{x})\operatorname{diag}\left(\sigma_{Rx}(\vec{y})^{(\ell)}_{(n)},\sigma_{Ry}(\vec{y})^{(\ell)}_{(n)}\right)^{-1}\aleph\\\aleph=\nabla^2I^{(\ell)}\left(\vec{y};\vec{f}(\vec{x})^{(\ell)}_{(n-1)}\right)\left(\nabla I_1^{(\ell)}(\vec{y})-\nabla I_2^{(\ell)}\left(\vec{y}+\vec{f}(\vec{x})^{(\ell)}_{(n-1)}\right)+\beth\right)\\\beth=\nabla^2I^{(\ell)}\left(\vec{y};\vec{f}(\vec{x})^{(\ell)}_{(n-1)}\right)\vec{f}(\vec{x})^{(\ell)}_{(n-1)}
\end{gathered}
\end{equation}

\subsubsection{Combined Model}
\label{sec:3.2.6}
Our first two methods, PLK-Intensity and PLK-Gradient, respectively employ zeroth- and first-order constancy constraints, neither of which is without drawbacks. First, illumination changes occur both systematically (by image conditions) as large low-frequency (not drastically spatially varying) $S(\vec{x})$, and stochastically (by sensor noise) as small high-frequency (rapidly oscillating) $\epsilon(\vec{x})$: thus brightness constancy struggles with low-frequency changes ($S(\vec{x})\gg\epsilon(\vec{x})$), while gradient constancy struggles with high-frequency changes ($\nabla\epsilon(\vec{x})\gg\nabla S(\vec{x})$). Second, both are susceptible to the aperture problem---the former in regions without bidirectional texture, as discussed in Section~\ref{sec:2.2.3}, and the latter in regions of smooth intensity variation, i.e. near-zero Hessians. Third, while gradients offer more information than intensities, their landscape is much more non-convex, risking the collapse of imperfect prior predictions toward spurious local minima. \\\\ Thus, we introduce our third method, \textit{PLK-Combined}, which simply concatenates the measurement models of PLK-Intensity and PLK-Gradient:
\begin{equation}
    \vec{f}(\vec{x})^{(\ell)}=2\vec{f}(\vec{x})^{(\ell-1)}+\vec{w}(\vec{x})^{(\ell)},\quad\vec{w}(\vec{x})^{(\ell)}\sim\mathcal{N}(\vec{0},\omega\vec{I}_2)
\end{equation}
\begin{equation}
    \vec{0}=\underbrace{\begin{bmatrix}I_2^{(\ell)}\left(\vec{x}_1+\vec{f}(\vec{x})^{(\ell)}\right)-I_1^{(\ell)}(\vec{x}_1)\\\vdots\\I_2^{(\ell)}\left(\vec{x}_{K^2}+\vec{f}(\vec{x})^{(\ell)}\right)-I_1^{(\ell)}(\vec{x}_{K^2})\\\nabla I_2^{(\ell)}\left(\vec{x}_1+\vec{f}(\vec{x})^{(\ell)}\right)-\nabla I_1^{(\ell)}(\vec{x}_1)\\\vdots\\\nabla I_2^{(\ell)}\left(\vec{x}_{K^2}+\vec{f}(\vec{x})^{(\ell)}\right)-\nabla I_1^{(\ell)}(\vec{x}_{K^2})\end{bmatrix}}_{\vec{h}\left(\vec{f}(\vec{x})^{(\ell)}\right)\in\RR^{3K^2}}+\vec{v}(\vec{x})^{(\ell)},\quad\vec{v}(\vec{x})^{(\ell)}\sim\mathcal{N}(\vec{0},\vec{R}(\vec{x})^{(\ell)})
\end{equation}

This simple construction allows information accumulation to be decomposed between brightness and gradient constancy components: $\vec{Y}'\gets\vec{Y}+\vec{H}^T_{\text{bright}}\vec{R}^{-1}_{\text{bright}}\vec{H}_{\text{bright}}+\vec{H}^T_{\text{grad}}\vec{R}^{-1}_{\text{grad}}\vec{H}_{\text{grad}}$, and likewise for $\vec{y}$. All other details follow directly from the previous sections.

\subsubsection{Summary}
\label{sec:3.2.7}
We have presented three variants of a novel optical flow framework, originating from Lucas and Kanade's least-squares approach, and rooted in their usage of image intensity derivatives: PLK-Intensity, which utilizes zeroth-order brightness constancy constraints and first-order gradients, PLK-Gradient, which employs analogous first-order gradient constancy constraints and second-order Hessians, and PLK-Combined, a concatenation of both approaches. Each varies only in its measurement model construction, but all adhere exactly to the IEKF execution framework, as outlined in Section~\ref{sec:2.4.3}. Each should outperform its predecessor, at the expense of additional computation time and memory. We will later see empirical results for PLK-Combined. \\\\ The complete method, with all three variants, is given by Algorithm~\ref{alg:plk} in Appendix~\ref{sec:a}.

\subsection{Probabilistic Farnebäck}
\label{sec:3.3}
Recall from Eqs.~\eqref{eq:fbpoly},\eqref{eq:fbplsq} that Farnebäck, rather than working directly with image intensities and gradients, approximates intensities via local quadratic polynomials centered at some $\vec{x}$:
\begin{equation}
    I(\vec{x}+\Delta\vec{x})\approx\Delta\vec{x}^T\vec{A}(\vec{x})\Delta\vec{x}+\vec{b}(\vec{x})^T\Delta\vec{x}+c(\vec{x})
\end{equation}

Upon obtaining these coefficients, Farnebäck estimates flows via weighted least-squares, similarly to Lucas-Kanade. To adapt a probabilistic approach, we directly copy our three PLK measurement models from Section~\ref{sec:3.2}, but we first apply this approximation, before computing Jacobians, pseudomeasurements, etc., to produce three analogous variants: \textit{probabilistic Farnebäck (PFB)-Intensity}, \textit{PFB-Gradient}, and \textit{PFB-Combined}. For simplicity we omit superscript $^{(\ell)}$ from $\vec{A},\vec{b},c$.

\subsubsection{Intensity Model}
\label{sec:3.3.1}
For our first model, PFB-Intensity, at IEKF iteration $n$, and each $\vec{y}\in\mathcal{W}_K(\vec{x})$, we compute Farnebäck approximations of $I_1^{(\ell)}$ and $I_2^{(\ell)}$, respectively centered at $\vec{y}$ and $\widetilde{\vec{y}}=\vec{y}+\vec{f}(\vec{x})^{(\ell)}_{(n-1)}$. These give:
\begin{equation}
\begin{aligned}
    I_1(\vec{y}) &\approx c_1(\vec{y}) \\ I_2\left(\vec{y}+\vec{f}(\vec{x})\right) &= I_2\Big(\widetilde{\vec{y}}+\smash{\underbrace{\left(\vec{f}(\vec{x})-\vec{f}(\vec{x})^{(\ell)}_{(n-1)}\right)}_{\Delta\vec{f}(\vec{x})}}\Big) \\[1.5em]&\approx \Delta\vec{f}(\vec{x})^T\vec{A}_2(\widetilde{\vec{y}})\Delta\vec{f}(\vec{x})+\vec{b}_2(\widetilde{\vec{y}})^T\Delta\vec{f}(\vec{x})+c_2(\widetilde{\vec{y}})
\end{aligned}
\end{equation}

In practice, we precompute $\vec{A}_2(\vec{y}),\vec{b}_2(\vec{y}),c_2(\vec{y})$ for all pixels $\vec{y}\in\mathcal{D}$, then interpolate:
\begin{equation}
\label{eq:pfbispat}
    \vec{A}_2(\widetilde{\vec{y}})=\operatorname{INTERP}(\widetilde{\vec{y}};\vec{A}_2)\qquad\vec{b}_2(\widetilde{\vec{y}})=\operatorname{INTERP}(\widetilde{\vec{y}};\vec{b}_2)\qquad c_2(\widetilde{\vec{y}})=\operatorname{INTERP}(\widetilde{\vec{y}};c_2)
\end{equation}

We assemble these approximations across all $\vec{y}\in\mathcal{W}_K(\vec{x})$ to produce $\vec{h}(\vec{f}(\vec{x}))\in\RR^{K^2}$:
\begin{equation}
    \vec{h}(\vec{f}(\vec{x}))\triangleq\begin{bmatrix}\Delta\vec{f}(\vec{x})^T\vec{A}_2(\widetilde{\vec{x}}_1)\Delta\vec{f}(\vec{x})+\vec{b}_2(\widetilde{\vec{x}}_1)^T\Delta\vec{f}(\vec{x})+c_2(\widetilde{\vec{x}}_1)-c_1(\vec{x}_1)\\\vdots\\\Delta\vec{f}(\vec{x})^T\vec{A}_2(\widetilde{\vec{x}}_{K^2})\Delta\vec{f}(\vec{x})+\vec{b}_2(\widetilde{\vec{x}}_{K^2})^T\Delta\vec{f}(\vec{x})+c_2(\widetilde{\vec{x}}_{K^2})-c_1(\vec{x}_{K^2})\end{bmatrix}
\end{equation}

We then copy our Kalman transition and measurement models from PLK-Intensity in Section~\ref{sec:3.1}, namely Eqs.~\eqref{eq:plktran},~\eqref{eq:plkmeas}, except we replace $I_1^{(\ell)}(\vec{y})$ and $I_2\left(\vec{y}+\vec{f}(\vec{x})^{(\ell)}\right)$ with their approximations:
\begin{equation}
    \vec{f}(\vec{x})^{(\ell)}=2\vec{f}(\vec{x})^{(\ell-1)}+\vec{w}(\vec{x})^{(\ell)},\quad\vec{w}(\vec{x})^{(\ell)}\sim\mathcal{N}(\vec{0},\omega\vec{I}_2)
\end{equation}
\begin{equation}
    \vec{0}=\vec{h}\left(\vec{f}(\vec{x})^{(\ell)}\right)+\vec{v}(\vec{x})^{(\ell)},\quad\vec{v}(\vec{x})^{(\ell)}\sim\mathcal{N}(\vec{0},\vec{R}(\vec{x})^{(\ell)})
\end{equation}

We now linearize our measurement by computing its Jacobian, as in Section~\ref{sec:3.2.2}. Differentiating the $i$th component of $\vec{h}$ gives:
\begin{equation}
    \left.\frac{\partial h(\vec{f}(\vec{x}))_i}{\partial\vec{f}(\vec{x})}\right|_{\vec{f}(\vec{x})=\vec{f}(\vec{x})^{(\ell)}_{(n-1)}}=\left.2\vec{A}_2(\widetilde{\vec{x}}_i)\Delta\vec{f}(\vec{x})+\vec{b}_2(\widetilde{\vec{x}}_i)\right|_{\vec{f}(\vec{x})=\vec{f}(\vec{x})^{(\ell)}_{(n-1)}}=\vec{b}_2(\widetilde{\vec{x}}_i)
\end{equation}

This is because $\vec{f}(\vec{x})^{(\ell)}_{(n-1)}-\vec{f}(\vec{x})^{(\ell)}_{(n-1)}$ eliminates the first term. Moreover, it gives:
\begin{equation}
    h\left(\vec{f}(\vec{x})^{(\ell)}_{(n-1)}\right)_i=c_2(\widetilde{\vec{x}}_i)-c_1(\vec{x}_i)
\end{equation}

Thus the $i$th component of our pseudomeasurement is:
\begin{equation}
\label{eq:pfbimeas}
    \left(z(\vec{x})^{(\ell)}_{(n)}\right)_i=c_1(\vec{x}_i)-c_2(\widetilde{\vec{x}}_i)+\vec{b}_2(\widetilde{\vec{x}}_i)^T\vec{f}(\vec{x})^{(\ell)}_{(n-1)}
\end{equation}

One will notice the similarity of this construction to the PLK-Intensity pseudomeasurement of Eq.~\eqref{eq:plkimeas}: in fact, we have constructed our measurement model precisely so that this would occur, i.e. that we can use $c$ and $b$ to replace $I$ and $\nabla I$, which aligns with their positions as zeroth- and first-order terms of Farnebäck's Taylor expansion. In a sense, $I,\nabla I$, and (later we will see) $\nabla^2I$, are ``isomorphic" to $c,\vec{b}$, and (later we will see) $\vec{A}$. Again in practice we average, as in Section~\ref{sec:3.2.2}, using, instead of $\vec{b}_2$:
\begin{equation}
\label{eq:pfbigrad}
    \vec{b}(\vec{y};\vec{f}(\vec{x}))\triangleq\frac{1}{2}(\vec{b}_1(\vec{y})+\vec{b}_2(\vec{y}+\vec{f}(\vec{x})))
\end{equation}

We construct measurement noise covariance $\vec{R}(\vec{x})^{(\ell)}$ exactly as we did in Section~\ref{sec:3.2.3}, namely Eq.~\eqref{eq:plkivar}, except, keeping with our Farnebäck coefficients, we replace $I$ with $c$, giving:
\begin{equation}
\label{eq:pfbivar}
    \sigma_R(\vec{y})^{(\ell)}_{(n)}\triangleq\max\left(\left|c_1(\vec{y})-c_2(\widetilde{\vec{y}})\right|^{\gamma},10^{-6}\right)
\end{equation}
\begin{equation}
    \vec{R}(\vec{x})^{(\ell)}_{(n)}\triangleq\operatorname{diag}\left(\sigma_R(\vec{x}_1)^{(\ell)}_{(n)},\cdots,\sigma_R(\vec{x}_{K^2})^{(\ell)}_{(n)}\right)\odot\vec{W}(\vec{x})
\end{equation}

This diagonal structure allows for the pixelwise information accumulation of Section~\ref{sec:3.2.4}. \\\\ Our posterior information matrix becomes:
\begin{equation}
\begin{gathered}
    \vec{Y}(\vec{x})^{(\ell)}_{(n)}\gets\vec{Y}(\vec{x})^{(\ell)}_{(0)}+\sum_{\vec{y}\in\mathcal{W}_K(\vec{x})}w_{\sigma}(\vec{y}-\vec{x})\left(\sigma_R(\vec{y})^{(\ell)}_{(n)}\right)^{-1}\aleph\\\aleph=\vec{b}\left(\vec{y};\vec{f}(\vec{x})^{(\ell)}_{(n-1)}\right)\vec{b}\left(\vec{y};\vec{f}(\vec{x})^{(\ell)}_{(n-1)}\right)^T
\end{gathered}
\end{equation}

Our posterior information vector becomes:
\begin{equation}
\begin{gathered}
    \vec{y}(\vec{x})^{(\ell)}_{(n)}\gets\vec{y}(\vec{x})^{(\ell)}_{(0)}+\sum_{\vec{y}\in\mathcal{W}_K(\vec{x})}w_{\sigma}(\vec{y}-\vec{x})\left(\sigma_R(\vec{y})^{(\ell)}_{(n)}\right)^{-1}\vec{b}\left(\vec{y};\vec{f}(\vec{x})^{(\ell)}_{(n-1)}\right)\aleph\\\aleph=\left(c_1(\vec{y})-c_2\left(\vec{y}+\vec{f}(\vec{x})^{(\ell)}_{(n-1)}\right)+\vec{b}\left(\vec{y};\vec{f}(\vec{x})^{(\ell)}_{(n-1)}\right)^T\vec{f}(\vec{x})^{(\ell)}_{(n-1)}\right)
\end{gathered}
\end{equation}

\subsubsection{Gradient Model}
\label{sec:3.3.2}
With the same motivation as in Section~\ref{sec:3.2.5}, we introduce our second Farnebäck model, PFB-Gradient. Upon obtaining coefficients $\vec{A},\vec{b},c$ via intensity $I$, we approximate gradient $\nabla I$ via the gradient of the Farnebäck polynomial:
\begin{equation}
    \nabla I(\vec{x}+\Delta\vec{x})=2\vec{A}(\vec{x})\Delta\vec{x}+\vec{b}(\vec{x})
\end{equation}

At IEKF iteration $n$, and each $\vec{y}\in\mathcal{W}_K(\vec{x})$, we compute approximations of $\nabla I_1^{(\ell)}$ and $\nabla I_2^{(\ell)}$, again centered at $\vec{y}$ and $\widetilde{\vec{y}}$:
\begin{equation}
\begin{aligned}
    \nabla I_1(\vec{y}) &\approx \vec{b}_1(\vec{y}) \\ \nabla I_2(\vec{y}+\vec{f}(\vec{x})) &\approx \nabla I_2(\widetilde{\vec{y}}+\Delta\vec{f}(\vec{x}))\approx2\vec{A}_2(\widetilde{\vec{y}})\Delta\vec{f}(\vec{x})+\vec{b}_2(\widetilde{\vec{y}})
\end{aligned}
\end{equation}

We concatenate these to produce our Kalman models:
\begin{equation}
    \vec{f}(\vec{x})^{(\ell)}=2\vec{f}(\vec{x})^{(\ell-1)}+\vec{w}(\vec{x})^{(\ell)},\quad\vec{w}(\vec{x})^{(\ell)}\sim\mathcal{N}(\vec{0},\omega\vec{I}_2)
\end{equation}
\begin{equation}
    \vec{0}=\underbrace{\begin{bmatrix}2\vec{A}_2(\widetilde{\vec{x}}_1)\Delta\vec{f}(\vec{x})+\vec{b}_2(\widetilde{\vec{x}}_1)-\vec{b}_1(\vec{x}_1)\\\vdots\\2\vec{A}_2(\widetilde{\vec{x}}_{K^2})\Delta\vec{f}(\vec{x})+\vec{b}_2(\widetilde{\vec{x}}_{K^2})-\vec{b}_1(\vec{x}_{K^2})\end{bmatrix}}_{\vec{h}\left(\vec{f}(\vec{x})^{(\ell)}\right)\in\RR^{2K^2}}+\vec{v}(\vec{x})^{(\ell)},\quad\vec{v}(\vec{x})^{(\ell)}\sim\mathcal{N}(\vec{0},\vec{R}(\vec{x})^{(\ell)})
\end{equation}

We linearize. Differentiating the $i$th component of $\vec{h}$ gives:
\begin{equation}
    \left.\frac{\partial h(\vec{f}(\vec{x}))_i}{\partial\vec{f}(\vec{x})}\right|_{\vec{f}(\vec{x})=\vec{f}(\vec{x})^{(\ell)}_{(n-1)}}=\left.2\vec{A}_2(\widetilde{\vec{x}}_i)\right|_{\vec{f}(\vec{x})=\vec{f}(\vec{x})^{(\ell)}_{(n-1)}}=2\vec{A}_2(\widetilde{\vec{x}}_i)
\end{equation}

Evaluating that component gives:
\begin{equation}
    h\left(\vec{f}(\vec{x})^{(\ell)}_{(n-1)}\right)_i=\vec{b}_2(\widetilde{\vec{x}}_i)-\vec{b}_1(\vec{x}_i)
\end{equation}

Thus the $i$th component of our pseudomeasurement is:
\begin{equation}
\label{eq:pfbgmeas}
    \left(z(\vec{x})^{(\ell)}_{(n)}\right)_i=\vec{b}_1(\vec{x}_i)-\vec{b}_2(\widetilde{\vec{x}}_i)+2\vec{A}_2(\widetilde{\vec{x}}_i)\vec{f}(\vec{x})^{(\ell)}_{(n-1)}
\end{equation}

Just as PFB-Intensity mirrors PLK-Intensity, this mirrors the PLK-Gradient pseudomeasurement, Eq.~\eqref{eq:plkgmeas}. Once more we average, as in Section~\ref{sec:3.2.5}, using, instead of $\vec{A}_2$:
\begin{equation}
\label{eq:pfbggrad}
    \vec{A}(\vec{y};\vec{f}(\vec{x}))\triangleq\frac{1}{2}\left(\vec{A}_1(\vec{y})+\vec{A}_2(\vec{y}+\vec{f}(\vec{x}))\right)
\end{equation}

We construct noise covariance $\vec{R}(\vec{x})^{(\ell)}$ according to Section~\ref{sec:3.2.5}, replacing $\nabla I$ with $\vec{b}$:
\begin{equation}
\label{eq:pfbgvar}
\begin{gathered}
    \sigma_{Rx}(\vec{y})^{(\ell)}_{(n)}\triangleq\max\left(\left|b_{1x}(\vec{y})-b_{2x}(\widetilde{\vec{y}})\right|^{\gamma},10^{-6}\right)\\\sigma_{Ry}(\vec{y})^{(\ell)}_{(n)}\triangleq\max\left(\left|b_{1y}(\vec{y})-b_{2y}(\widetilde{\vec{y}})\right|^{\gamma},10^{-6}\right)
\end{gathered}
\end{equation}
\begin{equation}
    \vec{R}(\vec{x})^{(\ell)}_{(n)}\triangleq\operatorname{diag}\left(\sigma_{Rx}(\vec{x}_1)^{(\ell)}_{(n)},\sigma_{Ry}(\vec{x}_1)^{(\ell)}_{(n)},\cdots,\sigma_{Rx}(\vec{x}_{K^2})^{(\ell)}_{(n)},\sigma_{Ry}(\vec{x}_{K^2})^{(\ell)}_{(n)}\right)\odot\vec{\mathcal{W}}(\vec{x})
\end{equation}

Our posterior information matrix becomes:
\begin{equation}
\begin{gathered}
    \vec{Y}(\vec{x})^{(\ell)}_{(n)}\gets\vec{Y}(\vec{x})^{(\ell)}_{(0)}+\sum_{\vec{y}\in\mathcal{W}_K(\vec{x})}w_{\sigma}(\vec{y}-\vec{x})\operatorname{diag}\left(\sigma_{Rx}(\vec{y})^{(\ell)}_{(n)},\sigma_{Ry}(\vec{y})^{(\ell)}_{(n)}\right)^{-1}\aleph\\\aleph=4\vec{A}\left(\vec{y};\vec{f}(\vec{x})^{(\ell)}_{(n-1)}\right)\vec{A}\left(\vec{y};\vec{f}(\vec{x})^{(\ell)}_{(n-1)}\right)
\end{gathered}
\end{equation}

Our posterior information vector becomes:
\begin{equation}
\begin{gathered}
    \vec{y}(\vec{x})^{(\ell)}_{(n)}\gets\vec{y}(\vec{x})^{(\ell)}_{(0)}+\sum_{\vec{y}\in\mathcal{W}_K(\vec{x})}w_{\sigma}(\vec{y}-\vec{x})\operatorname{diag}\left(\sigma_{Rx}(\vec{y})^{(\ell)}_{(n)},\sigma_{Ry}(\vec{y})^{(\ell)}_{(n)}\right)^{-1}\aleph\\\aleph=2\vec{A}\left(\vec{y};\vec{f}(\vec{x})^{(\ell)}_{(n-1)}\right)\left(\vec{b}_1(\vec{y})-\vec{b}_2\left(\vec{y}+\vec{f}(\vec{x})^{(\ell)}_{(n-1)}\right)+\beth\right)\\\beth=2\vec{A}\left(\vec{y};\vec{f}(\vec{x})^{(\ell)}_{(n-1)}\right)\vec{f}(\vec{x})^{(\ell)}_{(n-1)}
\end{gathered}
\end{equation}

\subsubsection{Combined Model}
\label{sec:3.3.3}
With the same motivation and method of concatenation as in Section~\ref{sec:3.2.6}, we may combine the measurement models of PFB-Intensity and PFB-Gradient to form PFB-Combined.

\subsubsection{Summary}
\label{sec:3.3.4}
We have adapted our three PLK variants to incorporate Farnebäck polynomial expansions, replacing image intensity derivatives with weighted least-squares coefficients, to produce three variants of a second novel optical flow framework: PFB-Intensity employs a constancy constraint on the zeroth-order coefficients and linearizes using first-order coefficients as its Jacobian, PFB-Gradient does analogously with first- and second-order coefficients, and PFB-Combined does both simultaneously. Each varies in its measurement model construction, and all adhere to the IEKF framework of Section~\ref{sec:2.4.3}. Each should outperform its predecessor, given additional computation time and memory. We will soon see results for PFB-Combined. \\\\ The complete method, with all three variants, is given by Algorithm~\ref{alg:pfb} in Appendix~\ref{sec:a}.

\section{Evaluation}
\label{sec:4}
We will soon evaluate our PLK-Combined and PFB-Combined algorithms by comparing their empirical performance against their classical counterparts. We first describe our experimental paradigm: our datasets, flow and uncertainty evaluation metrics, uncertainty estimation baselines, algorithm parameter settings, and code implementation details.

\subsection{Optical Flow Datasets}
\label{sec:4.1}
We evaluate our methods using two standard and widely used optical flow benchmark datasets. \\\\ The MPI-Sintel dataset \cite{Butler2012} is adapted from the open-source computer-animated short film Sintel. It provides a training set, consisting of 1,041 image pairs with dense ground-truth flow fields, and a test set, of 552 image pairs and no ground-truth. All images are 8-bit RGB, which we convert to grayscale, and of resolution $436\times1024$. Test-set evaluation is conducted, i.e. final metrics are computed, via uploading test-set flow field estimates to the dataset's website. The benchmark is known for featuring large displacements, non-rigid motions, atmospheric effects (e.g. fog, dust), frequent occlusions, fine structures, and independent object and camera motions. The dataset is provided in three render passes, of which we provide results for the latter two:
\begin{itemize}
    \item \textbf{Albedo}: Simplest, obeys brightness constancy, except at occlusions.
    \item \textbf{Clean}: Adds shading, shadows, specular reflections, and other lighting effects.
    \item \textbf{Final}: Adds motion blur, focus blur, and atmospheric effects.
\end{itemize}

The KITTI-2015 dataset \cite{Geiger2013}, \cite{Menze2015} is captured by a real-world camera on a moving vehicle in urban driving scenarios in Germany. It provides a training set, consisting of 200 image pairs with sparse ground-truth vectors computed using LiDAR, and a test set, of 200 image pairs and no ground-truth. All images are likewise 8-bit RGB, which we again convert to grayscale, and of resolution $375\times1242$. The benchmark features highly rigid motion, but also notably includes ego-motion (camera moves forward, environment shifts outward) and extreme lighting effects (e.g. sun glare, deep shadows, overexposed skies).

\subsection{Evaluation Metrics}
\label{sec:4.2}

\subsubsection{Flows}
\label{sec:4.2.1}
We first define the \textit{endpoint error} (EPE), for a single flow estimate $\widehat{\vec{f}(\vec{x})}$ and its corresponding ground-truth flow $\vec{f}(\vec{x})$, as the $L_2$ norm of their disparity:
\begin{equation}
    \operatorname{EPE}\left(\widehat{\vec{f}(\vec{x})};\vec{f}(\vec{x})\right)\triangleq\left\|\widehat{\vec{f}(\vec{x})}-\vec{f}(\vec{x})\right\|_2
\end{equation}

MPI-Sintel uses \textit{average endpoint error} (AEPE), defined as the average EPE for all pixels in the dataset with valid ground-truth flow. For $N$ such pixels $\Omega$, this is given by:
\begin{equation}
    \operatorname{AEPE}\triangleq\frac{1}{N}\sum_{\vec{x}\in\Omega}\left\|\widehat{\vec{f}(\vec{x})}-\vec{f}(\vec{x})\right\|_2
\end{equation}

KITTI-2015 reports \textit{F1-all}, defined as the percentage of pixels classified as outliers, where a pixel is such if its EPE exceeds both 3 pixels and 5\% of its ground-truth flow magnitude. For $N$ total pixels $\Omega$, this is given by:
\begin{equation}
    \operatorname{F1-all}\triangleq\frac{1}{N}\sum_{\vec{x}\in\Omega}\vec{1}\left(\left\|\widehat{\vec{f}(\vec{x})}-\vec{f}(\vec{x})\right\|_2>\max\left(3,0.05\left\|\vec{f}(\vec{x})\right\|_2\right)\right)
\end{equation}

\subsubsection{Uncertainties}
\label{sec:4.2.2}
In addition to flows $\widehat{\vec{f}(\vec{x})}$, we must describe precisely the quality of our uncertainty estimates $\vec{\Sigma}(\vec{x})$, a procedure comparatively much less common: our benchmarks do not provide ``ground-truth uncertainties" in any sense. Thus we use the canonical quality measure for uncertainty estimates, \textit{area under sparsification error} (AUSE) \cite{Ilg2018}. It assesses how well estimated uncertainties rank prediction errors, and requires a scalar uncertainty estimate $\sigma(\vec{x})$ for each pixel: \underline{we use the trace of $\vec{\Sigma}(\vec{x})$}. \\\\ We first aggregate EPEs $e(\vec{x})=\operatorname{EPE}\left(\widehat{\vec{f}(\vec{x})};\vec{f}(\vec{x})\right)$ and uncertainties $\sigma(\vec{x})$ for all $N$ pixels $\Omega$ in the dataset. For \textit{sparsification level} $\rho\in[0,1)$, we define \textit{predicted retained set} $\mathcal{R}_{\text{pred}}(\rho)$ and \textit{oracle retained set} $\mathcal{R}_{\text{oracle}}(\rho)$ as the sets of $\lfloor(1-\rho)N\rfloor$ pixels with respectively the lowest uncertainties and the lowest errors:
\begin{equation}
\begin{aligned}
    \mathcal{R}_{\text{pred}}(\rho) &= \underset{\substack{S\subset\Omega\\|S|=\lfloor(1-\rho)N\rfloor}}{\arg\min}\sum_{\vec{x}\in S}\sigma(\vec{x}) \\ \mathcal{R}_{\text{oracle}}(\rho) &= \underset{\substack{S\subset\Omega\\|S|=\lfloor(1-\rho)N\rfloor}}{\arg\min}\sum_{\vec{x}\in S}e(\vec{x})
\end{aligned}
\end{equation}

The \textit{sparsification error} at level $\rho$ is then the difference between the average errors of the predicted and oracle retained sets:
\begin{equation}
    \operatorname{SE}(\rho)=\frac{1}{\lfloor(1-\rho)N\rfloor}\sum_{\vec{x}\in\mathcal{R}_{\text{pred}}(\rho)}e(\vec{x})-\frac{1}{\lfloor(1-\rho)N\rfloor}\sum_{\vec{x}\in\mathcal{R}_{\text{oracle}}(\rho)}e(\vec{x})
\end{equation}

Accurate uncertainty estimates are expected to correlate with flow errors, i.e. a large $e(\vec{x})$ should imply a large $\sigma(\vec{x})$. The optimal estimate is the error itself, i.e. the oracle $\sigma(\vec{x})=e(\vec{x})$, which motivates our subtracting by it in $\operatorname{SE}(\rho)$. As $\rho\to1$, $\operatorname{SE}(\rho)\to0$: we aim to minimize initially large $\operatorname{SE}(\rho)$ values for $\rho\approx0$, and define our AUSE metric:
\begin{equation}
    \operatorname{AUSE}\triangleq\int_0^1\operatorname{SE}(\rho)d\rho
\end{equation}

We additionally employ \textit{Spearman's rank correlation coefficient} (CC) \cite{Spearman1904}, a non-parametric measure of correlation between rankings of two variables, which we aim to maximize. For each $\vec{x}\in\Omega$, denote by $r_{\sigma}(\vec{x})$ the rank of $\sigma(\vec{x})$, i.e. $r_{\sigma}(\vec{x})\in\{1,\cdots,N\}$, and by $r_e(\vec{x})$ the rank of $e(\vec{x})$, from lowest to highest. The means of these ranks are simply $\bar{r}_{\sigma}=\bar{r}_e=(N+1)/2$. Then Spearman's CC is simply the Pearson correlation coefficient between these ranks:
\begin{equation}
    \operatorname{CC}\triangleq\frac{\sum_{\vec{x}\in\Omega}(r_{\sigma}(\vec{x})-\bar{r}_{\sigma})(r_e(\vec{x})-\bar{r}_e)}{\sqrt{\sum_{\vec{x}\in\Omega}(r_{\sigma}(\vec{x})-\bar{r}_{\sigma})^2\sum_{\vec{x}\in\Omega}(r_e(\vec{x})-\bar{r}_e)^2}}
\end{equation}

\subsection{Uncertainty Baseline}
\label{sec:4.3}
We compare our confidence estimates against the post-hoc uncertainty method of \textit{forward-backward consistency} (FBC) \cite{Meister2018}, which assumes that reliable flow estimates $\vec{f}(\vec{x})$ from $I_1$ to $I_2$, i.e. forward flows, should resemble their additive inverses $-\widetilde{\vec{f}}(\vec{x})$ from $I_2$ to $I_1$, i.e. backward flows. Thus it first computes $\vec{f}(\vec{x})$ as usual, then swaps the order, $(I_1,I_2)\mapsto(I_2,I_1)$ to compute $\widetilde{\vec{f}}(\vec{x})$. It then defines an uncertainty metric using the residual between the two estimates:
\begin{equation}
    \sigma_{\text{FBC}}(\vec{x})\triangleq\left\|\vec{f}(\vec{x})+\widehat{\vec{f}}(\vec{x}+\vec{f}(\vec{x}))\right\|
\end{equation}

\subsection{Parameter Optimization}
\label{sec:4.4}
The complete PLK parameter set consists of: pyramid height $L$, window size $K$, process noise variance $\omega$, measurement noise variance exponent $\gamma$, weighting standard deviation $\sigma$, per-level iterations $N$. We add convergence tolerance $\epsilon$, so that if the difference between flows at consecutive iterations falls under $\epsilon$, i.e. $\left\|\vec{f}(\vec{x})^{(\ell)}_{(n)}-\vec{f}(\vec{x})^{(\ell)}_{(n-1)}\right\|_2<\epsilon$, then we stop iteration at level $\ell$, before all $N$ iterations conclude, and proceed directly to level $\ell+1$. The PFB parameter set consists of these, plus polynomial fit window size $K_p$ and polynomial fit weighting standard deviation $\sigma_p$. \\\\ For each of our four tests (PLK and PFB each on Sintel and KITTI), we search for an optimal combination of choices $(K,L,\omega,\gamma,N,\epsilon)$, resp. $(K,L,\omega,\gamma,\sigma,K_p,\sigma_p,N,\epsilon)$, over:
\begin{alignat*}{6}
    & L &&\in \{4,5,6,7\}\qquad && K &&\in \{9, 15, 21, 31\}\qquad && \omega &&\in\{10^{-4},10^{-3},10^{-2},10^{-1}\} \\
    & \gamma &&\in \{0,1,2\} && \sigma &&\in \{5,7,10,15\} && K_p &&\in \{5,7,9,11\} \\
    & \sigma_p &&\in \{1,1.2,1.5,2\} && N &&\in \{3,5,7,10\} && \epsilon &&\in \{10^{-4},10^{-3},10^{-2},10^{-1}\}
\end{alignat*}

Because this full parameter space consists of $4^6\cdot3\approx12$K, resp. $4^8\cdot3\approx196$K, combinations, we perform randomized search; we randomly select 100 combinations, and choose the best of those based on AEPE / F1. To further save time, and as test set baseline evaluation is not always possible (as we will discuss in Section~\ref{sec:5.2}), we split our training sets into 20\% ``training" and 80\% validation sets. Namely, for Sintel, we optimize parameters based on $208$ randomly selected training examples, and for KITTI, we do likewise on $40$ examples. We then evaluate our algorithms, with their best settings, alongside our baselines, with applicable parameters $(K,L,N,\epsilon)$, resp. $(K,L,K_p,\sigma_p,N,\epsilon)$, set identically, on remaining validation data, i.e. $833$ Sintel examples, $160$ KITTI examples. For Sintel, we use the Final pass for optimization, and both passes for validation.

\begin{table}[h]
    \centering
    \begin{tabular}{|l|c|c|c|c|}
        \hline
        \textbf{Parameter} & \textbf{PLK, Sintel} & \textbf{PLK, KITTI} & \textbf{PFB, Sintel} & \textbf{PFB, KITTI} \\ \hline
        $L$        &       7 &       7 &       6 &       5 \\
        $K$        &       9 &       9 &      31 &      15 \\
        $\omega$   & $10^{-1}$ & $10^{-2}$ & $10^{-1}$ & $10^{-1}$ \\
        $\gamma$   &       0 &       0 &       1 &       0 \\
        $\sigma$   &      15 &       7 &      10 &       5 \\
        $K_p$      &      -- &      -- &       9 &       5 \\
        $\sigma_p$ &      -- &      -- &       2 &     1.2 \\
        $N$        &       5 &       7 &       5 &      10 \\
        $\epsilon$ & $10^{-4}$ & $10^{-4}$ & $10^{-3}$ & $10^{-4}$ \\
        \hline
    \end{tabular}
    \caption{Optimal Parameters}
    \label{tab:params}
\end{table}

\subsection{Implementation Details}
\label{sec:4.5}
All procedures are implemented in Python. Classical LK and FB are performed via the OpenCV \cite{Bradski2008} API. All of our procedures (image pyramid construction, image padding, gradient computation, IEKF prediction and update steps) are implemented using CuPy \cite{Okuta2017}, which provides vectorized array operations and enables efficient parallel computation on GPUs. The measurement updates are implemented as pixel-wise CuPy kernels, each written in C$++$ and compiled at runtime. All experiments are executed on a Google Colab notebook connected to a single NVIDIA T4 GPU. Code is available at \url{https://github.com/awang124/optical-flow}.

\section{Results}
\label{sec:5}

\begin{figure}[h]
    \centering
    \begin{subfigure}[t]{0.32\textwidth}
        \centering
        \includegraphics[width=\textwidth]{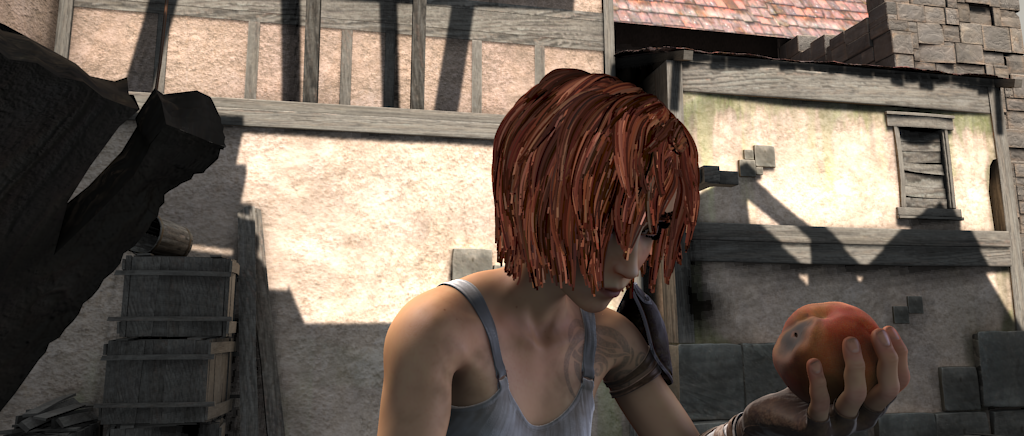}
        \caption{Clean pass first image}
    \end{subfigure}
    \hfill
    \begin{subfigure}[t]{0.32\textwidth}
        \centering
        \includegraphics[width=\textwidth]{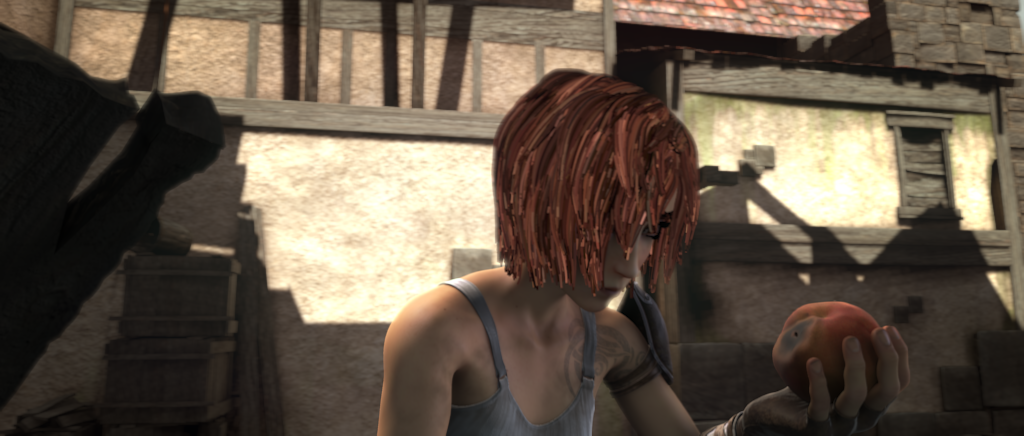}
        \caption{Final pass first image}
    \end{subfigure}
    \hfill
    \begin{subfigure}[t]{0.32\textwidth}
        \centering
        \includegraphics[width=\textwidth]{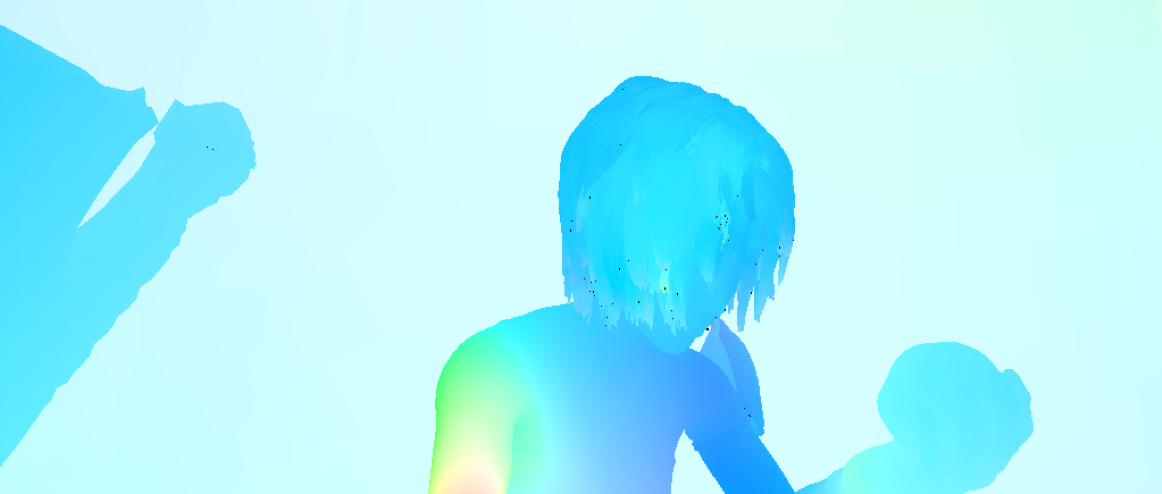}
        \caption{Ground-truth flow}
    \end{subfigure}
    \vspace{0.5em}
    \centering
    \begin{subfigure}[t]{0.32\textwidth}
        \centering
        \includegraphics[width=\textwidth]{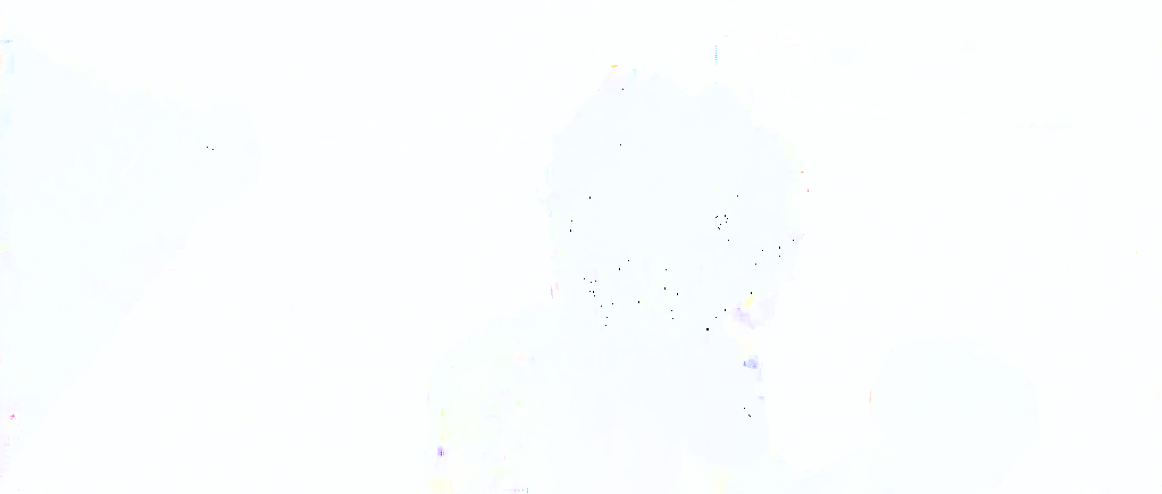}
        \caption{Lucas-Kanade Estimate}
    \end{subfigure}
    \hfill
    \begin{subfigure}[t]{0.32\textwidth}
        \centering
        \includegraphics[width=\textwidth]{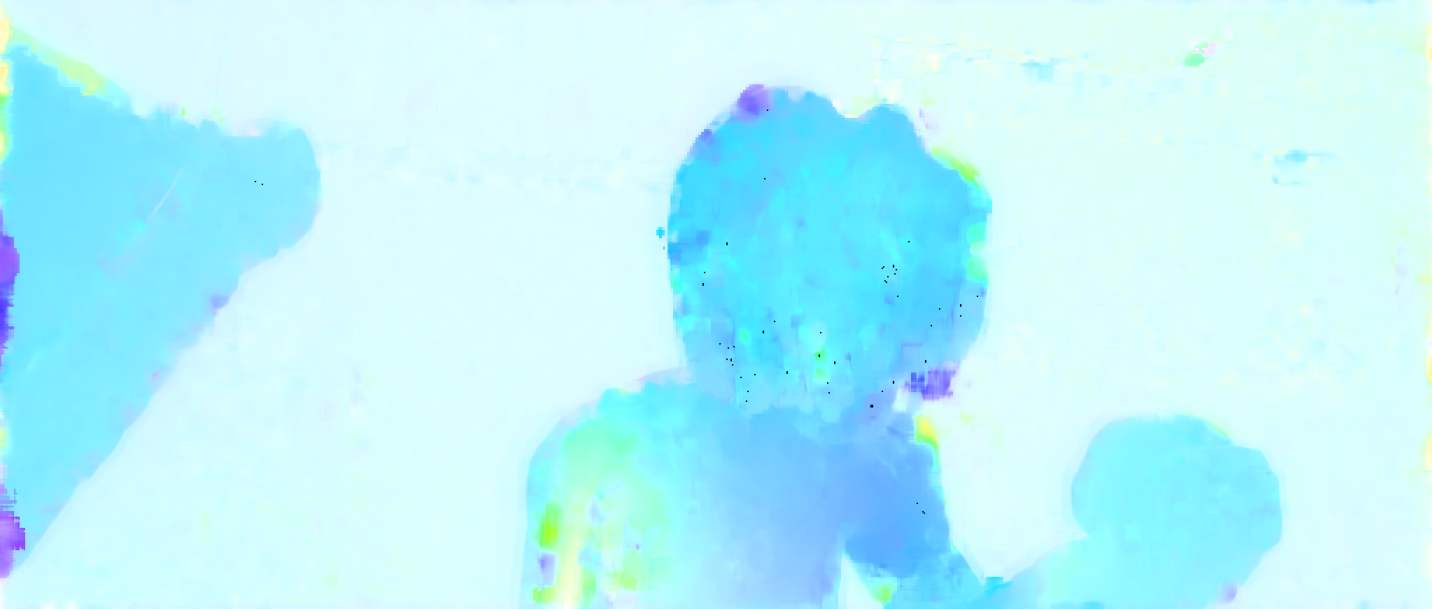}
        \caption{PLK-Combined Estimate}
    \end{subfigure}
    \hfill
    \begin{subfigure}[t]{0.32\textwidth}
        \centering
        \includegraphics[width=\textwidth]{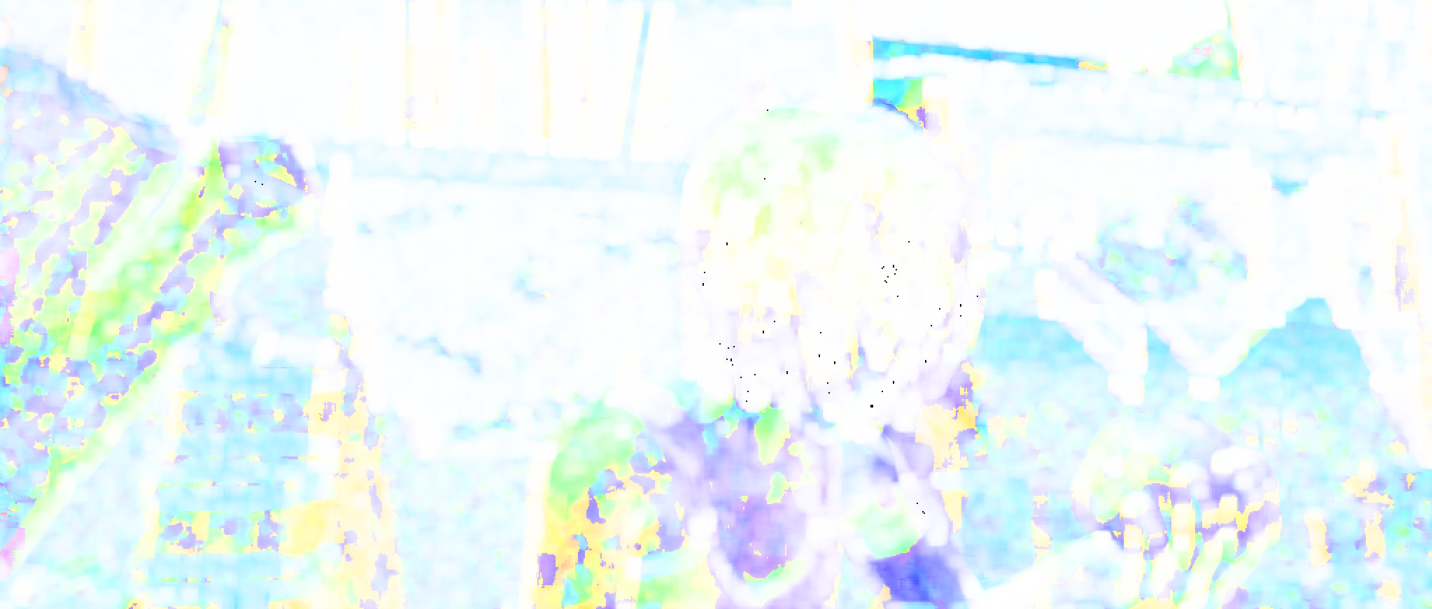}
        \caption{PLK-Combined Uncertainty}
    \end{subfigure}
    \vspace{0.5em}   
    \centering
    \begin{subfigure}[t]{0.32\textwidth}
        \centering
        \includegraphics[width=\textwidth]{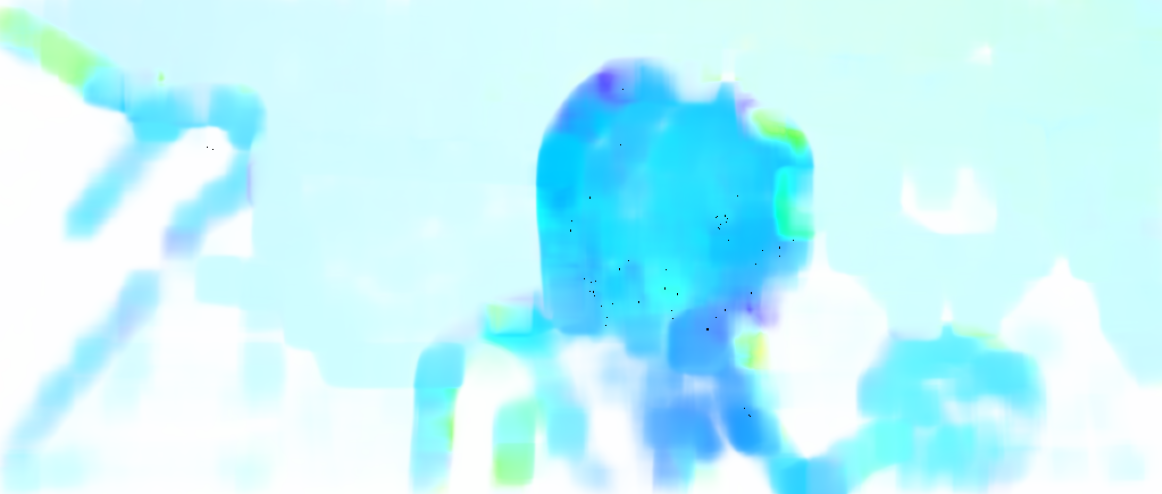}
        \caption{Farnebäck Estimate}
    \end{subfigure}
    \hfill
    \begin{subfigure}[t]{0.32\textwidth}
        \centering
        \includegraphics[width=\textwidth]{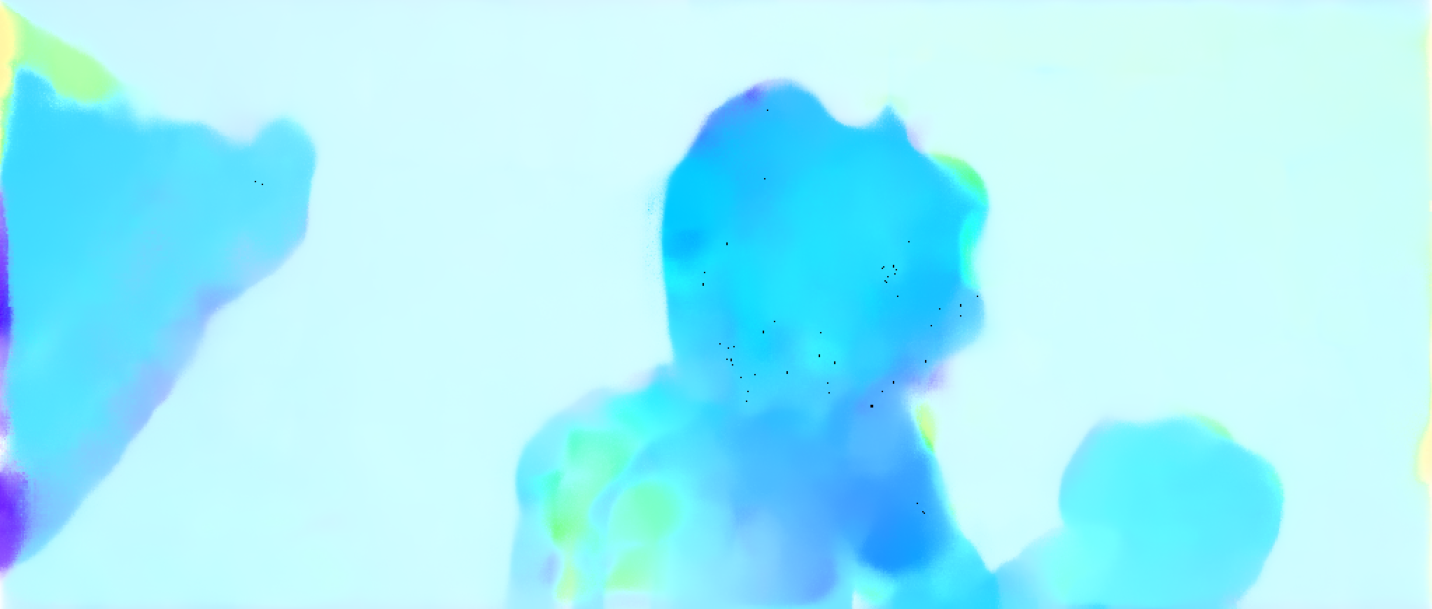}
        \caption{PFB-Combined Estimate}
    \end{subfigure}
    \hfill
    \begin{subfigure}[t]{0.32\textwidth}
        \centering
        \includegraphics[width=\textwidth]{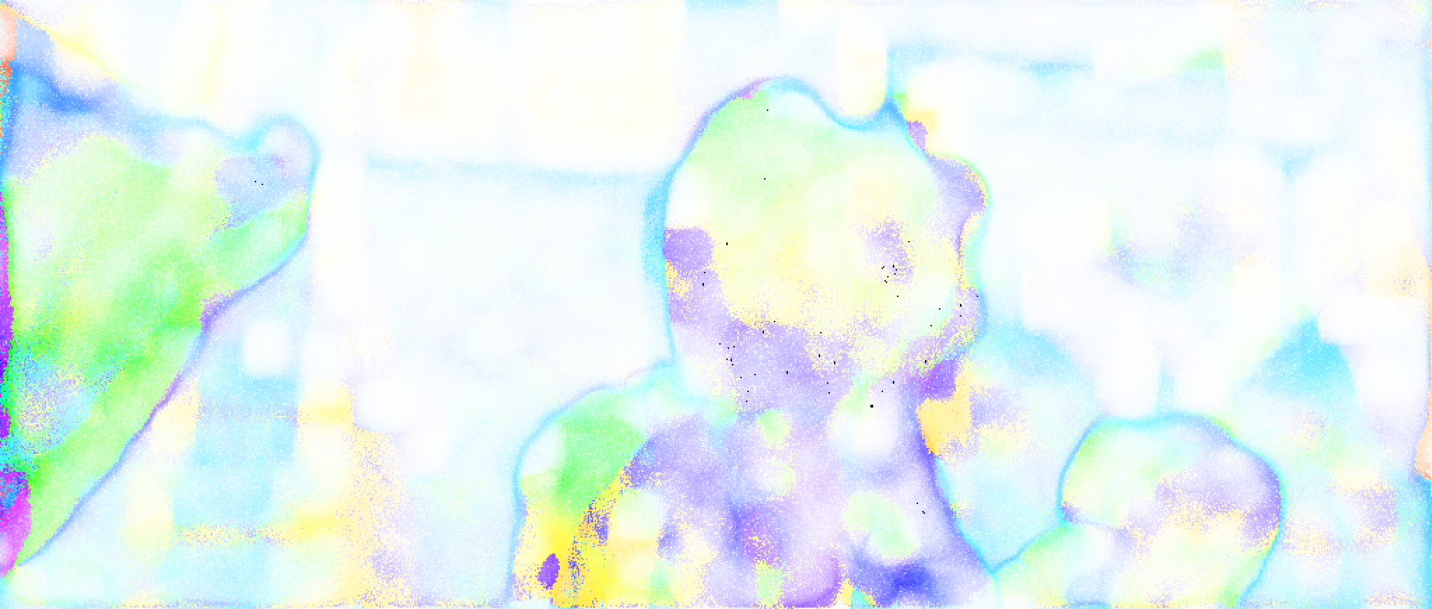}
        \caption{PFB-Combined Uncertainty}
    \end{subfigure}
    \centering
    \begin{minipage}[m]{0.13\textwidth}
        \vspace{0pt}
        \includegraphics[width=\linewidth]{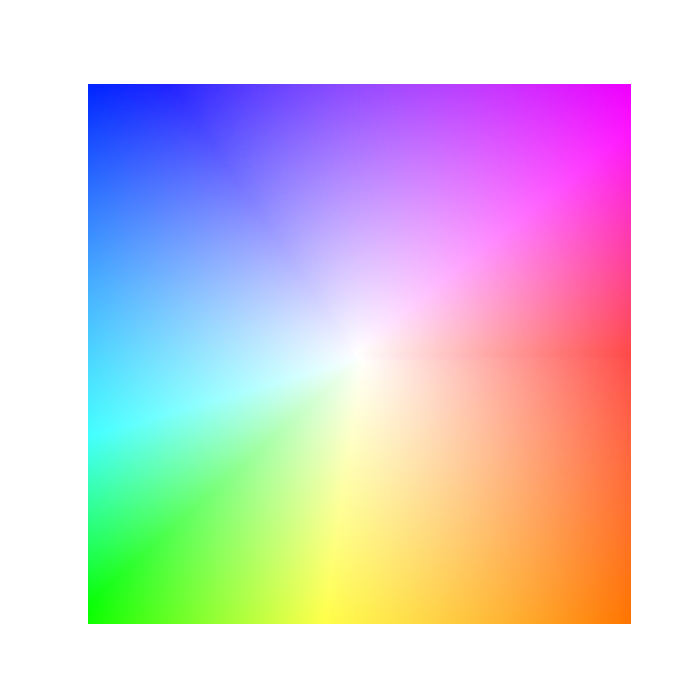}
    \end{minipage}
    \hspace{0.5em}
    \begin{minipage}[m]{0.78\textwidth}
        \vspace{0pt}
        \caption{Results on Sintel (Final) image pair. Vectors encoded using adjacent color wheel. Covariance matrices represented by their dominant eigenvectors, norms proportional to their corresponding (log-)eigenvalues.}
    	\label{fig:sintel}    
    \end{minipage}
\end{figure}

\subsection{Validation Set}
\label{sec:5.1}
We evaluate our methods and baselines on our validation set, as described in Section~\ref{sec:4.4}:
\begin{table}[h]
    \centering
    \renewcommand{\arraystretch}{0.95}
    \begin{tabular}{|l|c|c|c|}
        \hline
        \textbf{Method} & \textbf{Sintel (C) (AEPE)} & \textbf{Sintel (F) (AEPE)} & \textbf{KITTI (F1-all)} \\ \hline
        Lucas-Kanade & 14.798 & 20.677 & 46.7\% \\
        PLK-Combined &  7.125 &  7.855 & 46.1\% \\
        \% Reduction & 51.9\% & 62.0\% &  1.3\% \\ \hline
        Farnebäck    & 10.735 & 11.597 & 53.1\% \\
        PFB-Combined &  7.242 &  8.197 & 43.6\% \\
        \% Reduction & 32.5\% & 29.3\% & 17.9\% \\
        \hline
    \end{tabular}
    \caption{Flow Estimate Metrics (Validation)}
    \label{tab:valid1}
\end{table}

\vspace{-0.5cm}
We evaluate our covariance matrix traces against Forward-Backward consistency (using our algorithms, so that uncertainty metrics are independent of flow quality), as described in Section~\ref{sec:4.3}:
\begin{table}[h]
    \centering
    \renewcommand{\arraystretch}{0.95}
    \begin{tabular}{|l|c|c|c|c|c|c|}
        \hline
        \multirow{2}{*}{\textbf{Method}} & \multicolumn{3}{c|}{\textbf{AUSE}} & \multicolumn{3}{c|}{\textbf{CC}} \\ 
        \cline{2-7}
        & Sintel (C) & Sintel (F) & KITTI & Sintel (C) & Sintel (F) & KITTI \\ 
        \hline
        FBC on PLK     & 0.212 & 0.231 & 0.259 & 0.571 & 0.554 & 0.591 \\
        PLK Covariance & 0.722 & 0.658 & 0.619 & 0.178 & 0.218 & 0.212 \\ \hline
        FBC on PFB     & 0.227 & 0.234 & 0.218 & 0.581 & 0.569 & 0.611 \\
        PFB Covariance & 0.251 & 0.246 & 0.585 & 0.573 & 0.589 & 0.240 \\
        \hline
    \end{tabular}
    \caption{Uncertainty Estimate Metrics (Validation)}
    \label{tab:valid2}
\end{table}

\newpage
\begin{figure}[h]
    \centering
    \begin{subfigure}[t]{0.48\textwidth}
        \centering
        \includegraphics[width=\textwidth]{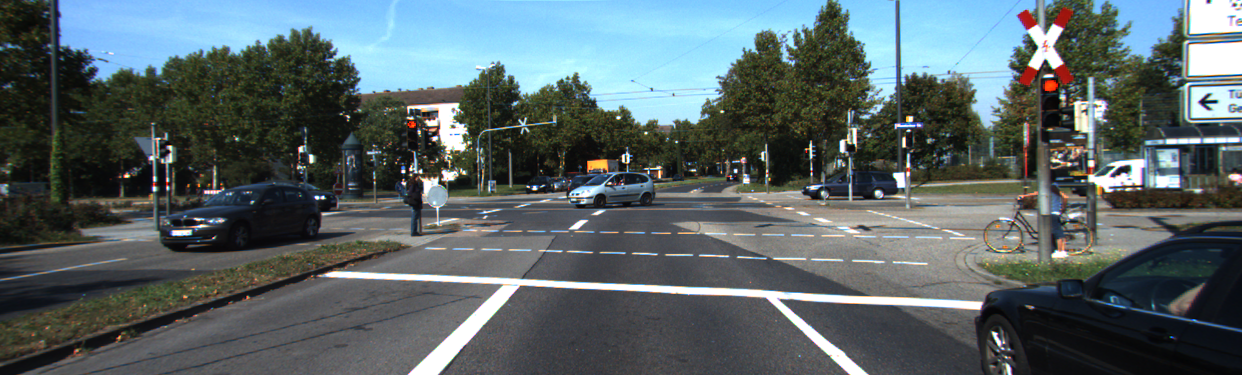}
        \caption{First image}
    \end{subfigure}
    \hfill
    \begin{subfigure}[t]{0.48\textwidth}
        \centering
        \includegraphics[width=\textwidth]{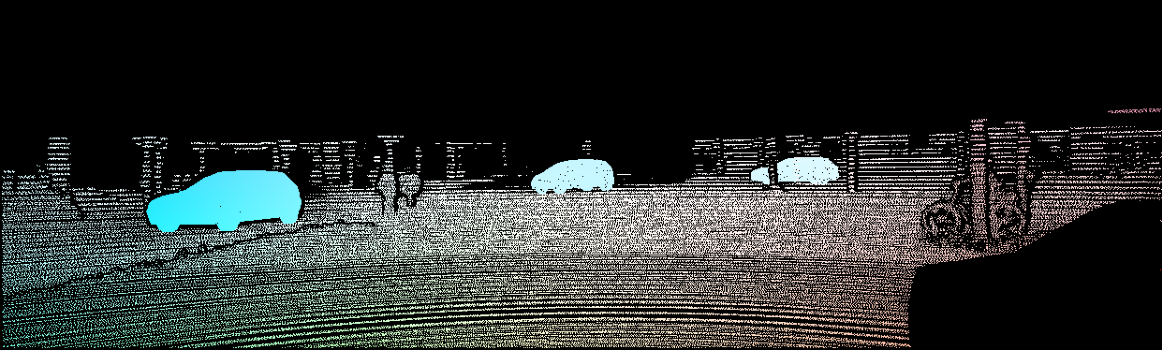}
        \caption{Ground-truth flow}
    \end{subfigure}
    \vspace{0.5em}
    \centering
    \begin{subfigure}[t]{0.48\textwidth}
        \centering
        \includegraphics[width=\textwidth]{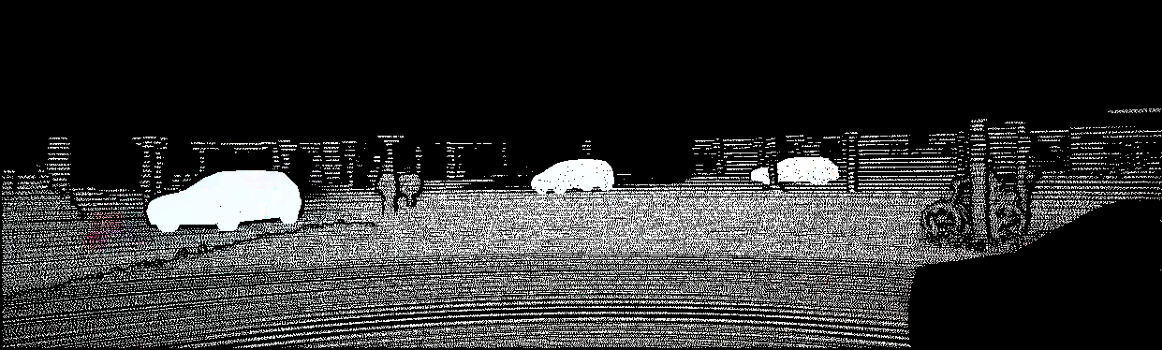}
        \caption{Lucas-Kanade Estimate}
    \end{subfigure}
    \hfill
    \begin{subfigure}[t]{0.48\textwidth}
        \centering
        \includegraphics[width=\textwidth]{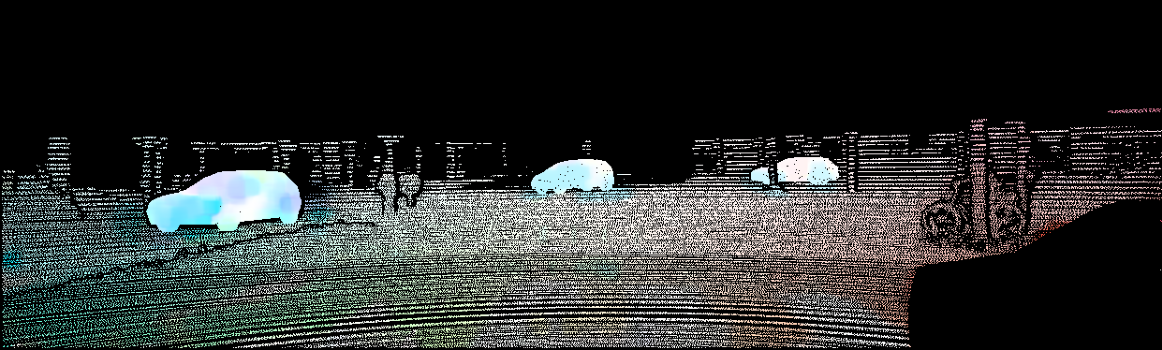}
        \caption{Farnebäck Estimate}
    \end{subfigure}
    \vspace{0.5em}
    \centering
    \begin{subfigure}[t]{0.48\textwidth}
        \centering
        \includegraphics[width=\textwidth]{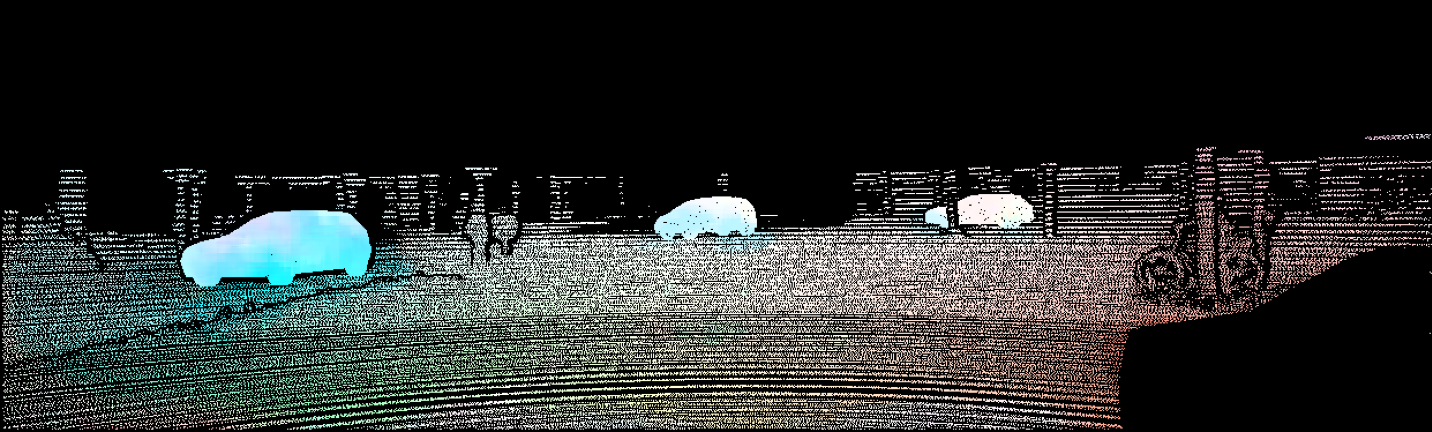}
        \caption{PLK-Combined Estimate}
    \end{subfigure}
    \hfill
    \begin{subfigure}[t]{0.48\textwidth}
        \centering
        \includegraphics[width=\textwidth]{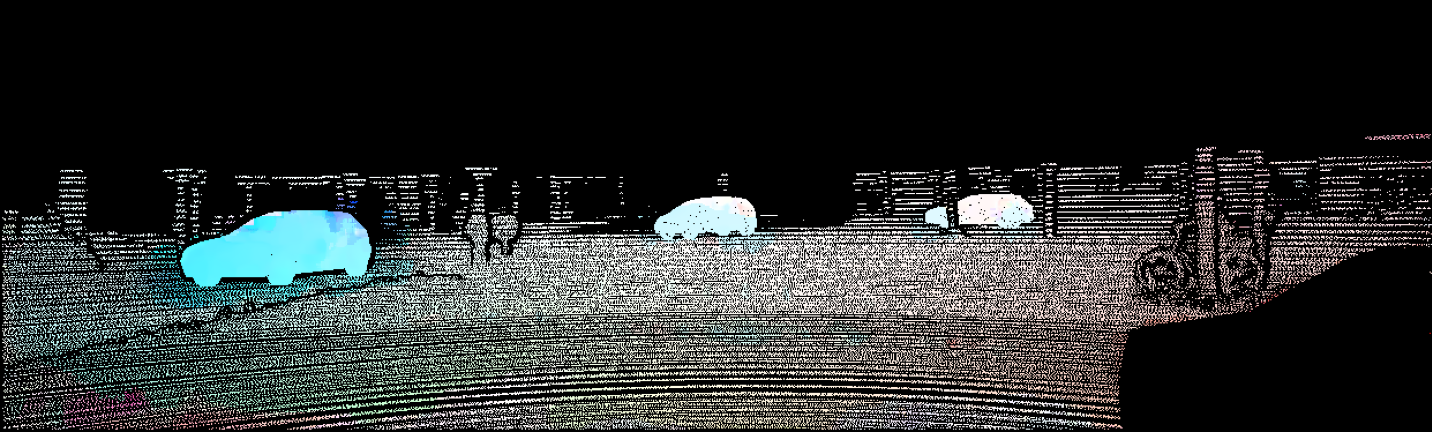}
        \caption{PFB-Combined Estimate}
    \end{subfigure}
    \vspace{0.5em}
    \centering
    \begin{subfigure}[t]{0.48\textwidth}
        \centering
        \includegraphics[width=\textwidth]{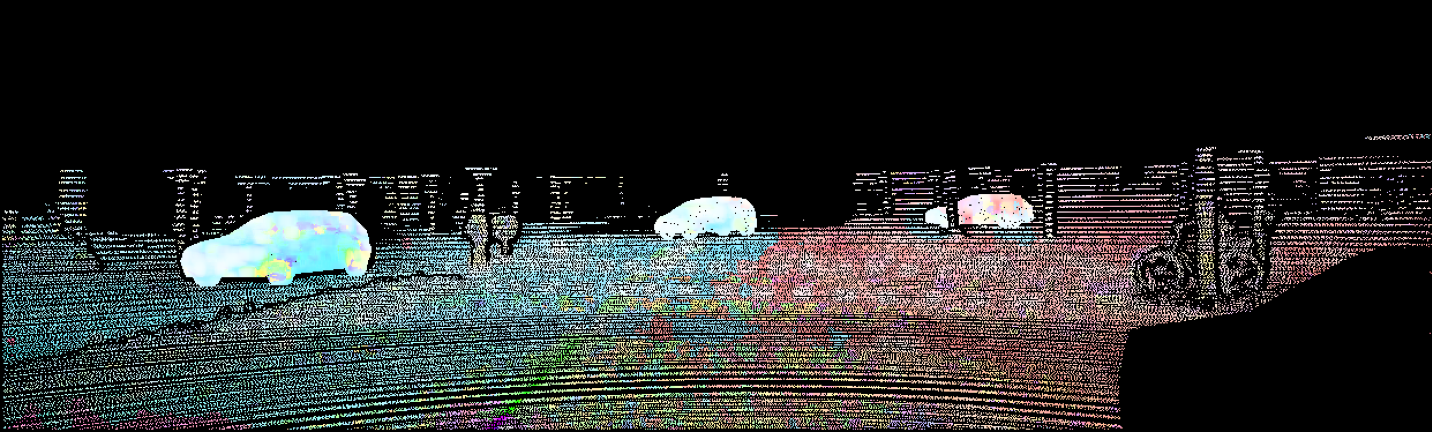}
        \caption{PLK-Combined Uncertainty}
    \end{subfigure}
    \hfill
    \begin{subfigure}[t]{0.48\textwidth}
        \centering
        \includegraphics[width=\textwidth]{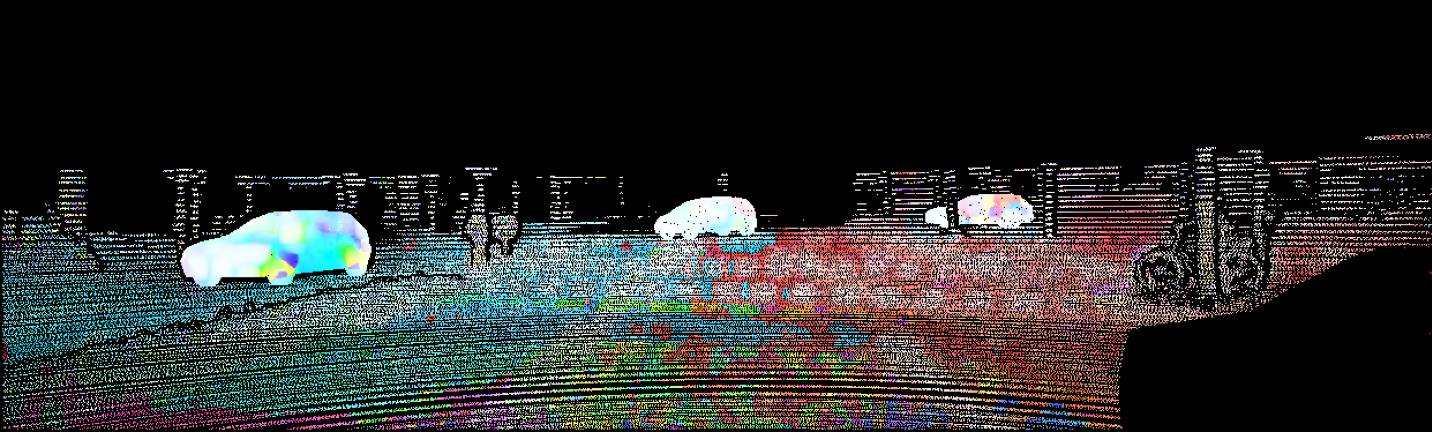}
        \caption{PFB-Combined Uncertainty}
    \end{subfigure}
    \centering
    \begin{minipage}[m]{0.13\textwidth}
        \vspace{0pt}
        \includegraphics[width=\linewidth]{img/colorwheel.png}
    \end{minipage}
    \hspace{0.5em}
    \begin{minipage}[m]{0.78\textwidth}
        \vspace{0pt}
        \caption{Results on KITTI image pair. Vectors and matrices encoded as in Figure~\ref{fig:sintel}. Pixels with no ground-truth set to black.}
    	\label{fig:kitti}
    \end{minipage}
\end{figure}

\subsection{Test Set}
\label{sec:5.2}
For either dataset, we compress and upload test set estimates to their webpage via a user account; test set uncertainty evaluation is thus impossible, motivating our validation set. Baseline results are obtained from datasets' published results tables; Lucas-Kanade and Farnebäck have published submissions only on KITTI, not on Sintel, thus a validation set was further required to compare our methods' performance. Our submitted estimates produced: 

\begin{table}[h]
    \centering
    \renewcommand{\arraystretch}{0.95}
    \begin{tabular}{|l|c|c|c|}
        \hline
        \textbf{Method} & \textbf{Sintel (C) (AEPE)} & \textbf{Sintel (F) (AEPE)} & \textbf{KITTI (F1-all)} \\ \hline
        Lucas-Kanade &   --   &   --   & 72.7\% \\
        PLK-Combined & 10.896 & 11.630 & 46.4\% \\
        \% Reduction &   --   &   --   & 36.2\% \\ \hline
        Farnebäck    &   --   &   --   & 53.1\% \\
        PFB-Combined & 11.118 & 11.809 & 44.6\% \\
        \% Reduction &   --   &   --   & 16.0\% \\
        \hline
    \end{tabular}
    \caption{Flow Estimate Metrics (Test)}
    \label{tab:test}
\end{table}

\section{Discussion}
\label{sec:6}

\subsection{Performance Analysis}
\label{sec:6.1}
PLK-Combined and PFB-Combined consistently outperform their classical counterparts in flow estimation. On the MPI-Sintel validation set, PLK-Combined achieved 51.9\% and 62.0\% AEPE reductions on the Clean and Final passes, respectively, compared to Lucas-Kanade. Similarly, PFB-Combined achieved 32.5\% and 29.3\% reductions compared to Farnebäck. On the KITTI test set, PLK-Combined and PFB-Combined respectively achieved 36.2\% and 16.0\% F1-all reductions. \\\\ These improvements stem primarily from three key architectural enhancements. First, while Lucas-Kanade and Farnebäck rely solely on zeroth-order brightness constancy, our approach fuses intensity and gradient constancy residuals. Incorporating second-order curvature information helps enable invariance to global illumination changes. Second, by modeling pyramid level changes as state evolutions with process noise, our framework prevents error accumulations across resolution scales. Coarse-level estimates serve as informed probabilistic priors rather than hard initializations, allowing fine levels to refine motion fields robustly. Third, while Lucas-Kanade computes gradients solely on $I_1$, which holds them static across intra-level iterations, our algorithms average gradients (and Hessians) across both $I_1$ and warped $I_2$ at every iteration, reducing linearization bias and improving stability. A related note, unlike single-pass linearizations prone to overshooting or divergence under large displacements or curvatures, our IEKF relinearizes its measurement function around successive posterior mean estimates, taking incremental steps more likely to remain within local basins of attraction, thus converging even when there exist large motions and sharp boundaries. \\\\ A large number of pyramid levels $L$, 5 through 7, was required to capture the large displacements of Sintel and KITTI. The window size $K$ varied widely, small (9) on PLK, as its derivatives $\nabla I(\vec{x}),\nabla^2I(\vec{x})$ are local with high spatial frequencies, giving well-conditioned small windows, and large (up to 31) on PFB, as its coefficients $\vec{A}(\vec{x}),\vec{b}(\vec{x})$ computed for nearby pixels being heavily correlated require larger windows to accumulate sufficient distinct structural constraints. Window weighting standard deviation $\sigma$ varied also, large on Sintel (10-15) for broader smoothing to combat noise across its complex, non-rigid, and fine-textured animations, and small on KITTI (5-7) for sharper spatial decay to accommodate large car motions with sharp boundaries against static backgrounds. Full search results showed that algorithms were fairly robust to process noise $\omega$, indicating posterior estimates were dominated and stabilized by measurement data rather than adhering closely to their upsampling priors. Similar robustness existed to per-level iterations $N$ and convergence tolerance $\epsilon$, demonstrating that measurement updates generally converged quickly. \\\\ By far the most sensitive parameter is measurement noise variance exponent $\gamma$. For PLK on Sintel, $\gamma=0,1,2$ respectively give AEPEs $\sim$8, $\sim$10, $\sim$12. For PLK on KITTI, they give F1-alls $\sim$50\%, $\sim$75\%, $\sim$85\%. For PFB on KITTI, they give F1-alls $\sim$50\%, $\sim$65\%, $\sim$85\%. Sintel and KITTI are characterized by widespread brightness and gradient constancy violations. When $\gamma>0$, large residuals cause noise variance $\sigma_R(\vec{y})$ to explode, and the IEKF to aggressively reject valid measurement data and default to its priors, giving stagnant estimates and compounded errors. However, for PFB on Sintel, they give AEPEs $\sim$11, $\sim$8, $\sim$11. PFB's relatively wide polynomial fit acts as a low-pass filter, smoothing out Sintel's high-frequency noise, thus allowing $\gamma=1$ to focus on sharp constancy violations, rejecting these measurements without sacrificing benign data. Only here are our uncertainties well-calibrated, comparable with Forward-Backward consistency. We thus see constancy residuals are a suitable proxy for flow uncertainties, but not for update weights.

\subsection{Limitations and Future Work}
\label{sec:6.2}
There exist several promising directions for future exploration. First, our entire methodology employs grayscale images, i.e. color images are first converted to grayscale. Discarding chromaticity channels (hue and saturation) discards valuable photometric invariants, and boundaries between objects with distinct colors but identical grayscale intensities become invisible, potentially creating large uniform regions and unnecessarily triggering the aperture problem. A straightforward solution for RGB images might untangle their three color channels, enforce our constancy constraints on each separately, and concatenate their residuals. Second, as alluded to, one might design a more complex measurement noise covariance estimation mechanism. Our current approach assumes independence between pixels in each window, and merely uses powers of constancy constraint residuals. A better model could instead use a correlated measurement noise, defined using a spatial covariance kernel, at the sacrifice of independent accumulation, thus requiring additional time and memory. Third, one might consider a more nuanced state transition model. While our measurement models integrate image intensities, gradients, and Hessians in a sophisticated manner, our current state transition model is a simple upsampling with an isotropic process noise covariance. A future model could replace this with an anisotropic or even nonstationary covariance matrix, defined using a similar spatial kernel or estimated using flow differences at previous pyramid levels. Finally, one might employ an adaptive windowing technique. Our current algorithms use constant-size windows across pixel locations and image resolutions, and require a user to manually set this size. A more dynamic model could estimate window size as the smallest size not falling victim to the aperture problem, i.e. with a structure tensor condition number above a given threshold, or allow window sizes to vary spatially, using edge detection or segmentation to separate distinct objects. \\\\ More broadly, first, our methodology assumes displacements to be purely translational, when in reality, objects regularly also undergo rotation, scaling, and shearing. Thus one might augment our state vector from its current 2D translation to a 6D local affine motion model, consisting of a $2\times2$ deformation matrix, modeling these additional motion types, alongside a 2D translation vector. Second, we also assume flows to have Gaussian posterior distributions. However, at sharp motion boundaries or semi-transparent occlusions, where a window straddles motions of different directions, these posteriors are inherently multimodal. In these cases, our algorithms will either collapse onto one mode, ignoring the other(s), or average them into an invalid intermediate flow with an overinflated uncertainty. Thus, one might replace the IEKF with e.g. a Gaussian sum filter (GSF) \cite{Alspach1972}, which approximates a density as a Gaussian mixture. Third, our algorithms enforce spatial coherence only locally. Under the previously mentioned issue of static window sizes, we risk encountering vast textureless regions, where all window pixels lack strong gradient or curvature signal. To combat this, one could introduce a global adjustment between pyramid levels, using e.g. a Gaussian Markov Random Field (GMRF) \cite{Rue2005}, allowing low-uncertainty estimates at edges and corners to diffuse into high-uncertainty regions. Finally, our central technique of Kalman filtering could be used across timesteps for image sequences with more than two frames, in addition to across resolution levels. Here the level $L$ posterior flow and uncertainty estimates from pair $(I_t,I_{t+1})$ would initialize the level $1$ estimates for pair $(I_{t+1},I_{t+2})$.

\subsection{Conclusion}
\label{sec:6.3}
Our integration of classical pyramidal optical flow methods with iterated extended Kalman filtering over image resolution levels provides a principled paradigm and two novel algorithms marking a concrete contribution toward robust motion estimation with transparent uncertainty quantification.
\printbibliography

\newpage
\begin{appendices}
\section{Appendix: Algorithm Pseudocode}
\label{sec:a}

\begin{algorithm}
    \caption{Pyramidal Lucas-Kanade}
    \label{alg:lk}
    \linespread{1.5}\selectfont
    \begin{algorithmic}[1]
        \Require images $I_1,I_2$, tracked pixels $\mathcal{P}=(\vec{x}_m)_{m=1}^M\in\mathcal{D}^M$, pyramid height $L$, window size $K$, weighting standard deviation $\sigma$, per-level iterations $N$, convergence tolerance $\epsilon$
        \Ensure tracked pixels' new positions $\mathcal{P}'=(\vec{x}_m')_{m=1}^M$, flow field $\mathcal{F}=(\vec{f}_m=\vec{x}_m'-\vec{x}_m)_{m=1}^M$
        \State Build and pad image pyramids $(I_1^{(\ell)})_{\ell=1}^L,(I_2^{(\ell)})_{\ell=1}^L$ by Eqs.~\eqref{eq:pyr},~\eqref{eq:pad}
        \For{$\ell=1$ \textbf{to} $L$} \Comment{Coarse-to-fine pyramid loop}
            \For{each tracked pixel $\vec{x}=\vec{x}_m\in\mathcal{P}$} \Comment{Process each tracked point}
                \State Downscale starting position: $\vec{x}^{(\ell)}\gets\vec{x}/2^{L-\ell}$
                \If{$\ell=1$}
                    \State Initialize new position: $\vec{x}'^{(\ell)}_{(0)}\gets\vec{x}^{(\ell)}$
                \Else
                    \State Upscale new position from previous level: $\vec{x}'^{(\ell)}_{(0)}\gets2\vec{x}'^{(\ell-1)}$
                \EndIf
                \For{$n=1$ \textbf{to} $N$} \Comment{Sub-pixel optimization loop}
                    \For{each neighbor $\vec{y}\in\mathcal{W}_K(\vec{x}^{(\ell)})$}
                        \State Compute $I_x(\vec{y}),I_y(\vec{y})$ on $I_1^{(\ell)}$ by Eqs.~\eqref{eq:gradx},~\eqref{eq:grady}
                        \State Compute $I_t(\vec{y})\gets\operatorname{INTERP}\left(\vec{y}+\left(\vec{x}'^{(\ell)}_{(n-1)}-\vec{x}^{(\ell)}\right);I_2^{(\ell)}\right)-I_1^{(\ell)}(\vec{y})$
                    \EndFor
                    \State Construct $\vec{\nabla I}(\vec{x}),\vec{I_t}(\vec{x})$ by Eq.~\eqref{eq:lkmat}
                    \State Compute flow increment $\vec{f}(\vec{x})$ by Eq.~\eqref{eq:lklsq}
                    \State Update new position: $\vec{x}'^{(\ell)}_{(n)}\gets\vec{x}'^{(\ell)}_{(n-1)}+\vec{f}(\vec{x})$
                    \If{$\|\vec{f}(\vec{x})\|_2<\epsilon$} \Comment{Optional early convergence}
                        \State $\vec{x}'^{(\ell)}_{(N)}\gets\vec{x}'^{(\ell)}_{(n)}$
                        \State \textbf{break}
                    \EndIf
                \EndFor
                \State Save this level's final position: $\vec{x}'^{(\ell)}\gets\vec{x}'^{(\ell)}_{(N)}$
                \If{$\ell=L$}
                    \State Save final position: $\vec{x}_m'\gets\vec{x}'^{(\ell)}$
                    \State Save flow vector: $\vec{f}_m\gets\vec{x}_m'-\vec{x}_m$
                \EndIf
            \EndFor
        \EndFor
    \end{algorithmic}
\end{algorithm}

\begin{algorithm}
    \caption{Pyramidal Farnebäck}
    \label{alg:fb}
    \linespread{1.5}\selectfont
    \begin{algorithmic}[1]
        \Require images $I_1, I_2$, tracked pixels $\mathcal{P}=(\vec{x}_m)_{m=1}^M\in\mathcal{D}^M$, pyramid height $L$, window size $K$, weighting standard deviation $\sigma$, polynomial fit window size $K_p$, polynomial fit weighting standard deviation $\sigma_p$, per-level iterations $N$, convergence tolerance $\epsilon$
        \Ensure tracked pixels' new positions $\mathcal{P}'=(\vec{x}_m')_{m=1}^M$, flow field $\mathcal{F}=(\vec{f}_m=\vec{x}_m'-\vec{x}_m)_{m=1}^M$
        \State Build and pad image pyramids $(I_1^{(\ell)})_{\ell=1}^L,(I_2^{(\ell)})_{\ell=1}^L$ by Eqs.~\eqref{eq:pyr},~\eqref{eq:pad}
        \For{$\ell=1$ \textbf{to} $L$} \Comment{Coarse-to-fine pyramid loop}
            \For{each pixel $\vec{y}\in\mathcal{D}^{(\ell)}$} \Comment{Pre-compute Farnebäck coefficients}
                \State Compute $\vec{A}_1^{(\ell)}(\vec{y}),\vec{b}_1^{(\ell)}(\vec{y}),\vec{A}_2^{(\ell)}(\vec{y}),\vec{b}_2^{(\ell)}(\vec{y})$ by Eq.~\eqref{eq:fbplsq}
            \EndFor
            \For{each tracked pixel $\vec{x}=\vec{x}_m\in\mathcal{P}$} \Comment{Process each tracked point}
                \State Downscale coordinate to current level: $\vec{x}^{(\ell)}\gets\vec{x}/2^{L-\ell}$
                \If{$\ell=1$}
                    \State Initialize new position: $\vec{x}'^{(\ell)}_{(0)}\gets\vec{x}^{(\ell)}$
                \Else
                    \State Upscale new position from previous coarser level: $\vec{x}'^{(\ell)}_{(0)}\gets2\vec{x}'^{(\ell-1)}$
                \EndIf
                \For{$n=1$ \textbf{to} $N$} \Comment{Sub-pixel optimization loop}
                    \For{each neighbor $\vec{y}\in\mathcal{W}_K(\vec{x}^{(\ell)})$}
                        \State Compute $\vec{A}(\vec{y})\gets\frac{1}{2}\left(\vec{A}_1^{(\ell)}(\vec{y})+\operatorname{INTERP}\left(\vec{y}+\left(\vec{x}'^{(\ell)}_{(n-1)}-\vec{x}^{(\ell)}\right);\vec{A}_2^{(\ell)}\right)\right)$
                        \State Compute $\Delta\vec{b}(\vec{y})\gets\frac{1}{2}\left(\vec{b}_1^{(\ell)}(\vec{y})-\operatorname{INTERP}\left(\vec{y}+\left(\vec{x}'^{(\ell)}_{(n-1)}-\vec{x}^{(\ell)}\right);\vec{b}_2^{(\ell)}\right)\right)$
                    \EndFor
                    \State Compute flow increment $\vec{f}(\vec{x})$ by Eq.~\eqref{eq:fblsq}
                    \State Update new position: $\vec{x}'^{(\ell)}_{(n)}\gets\vec{x}'^{(\ell)}_{(n-1)}+\vec{f}(\vec{x})$
                    \If{$\|\vec{f}(\vec{x})\|_2<\epsilon$} \Comment{Optional early convergence}
                        \State $\vec{x}'^{(\ell)}_{(N)}\gets\vec{x}'^{(\ell)}_{(n)}$
                        \State \textbf{break}
                    \EndIf
                \EndFor
                \State Save this level's final position: $\vec{x}'^{(\ell)}\gets\vec{x}'^{(\ell)}_{(N)}$
                \If{$\ell=L$}
                    \State Save final position: $\vec{x}_m'\gets\vec{x}'^{(\ell)}$
                    \State Save flow vector: $\vec{f}_m\gets\vec{x}_m'-\vec{x}_m$
                \EndIf
            \EndFor
        \EndFor
    \end{algorithmic}
\end{algorithm}

\begin{algorithm}
    \caption{Iterated Extended Kalman Filter}
    \label{alg:iekf}
    \linespread{1.5}\selectfont
    \begin{algorithmic}[1]
        \Require measurement function $\vec{h}$, measurements $(\vec{z}_t)_{t=1}^T$, model parameters $((\vec{F}_t,\vec{Q}_t,\vec{R}_t))_{t=1}^T$, initial state $(\vec{x}_0,\vec{\Sigma}_0)$, timesteps $T$, per-timestep iterations $N$, convergence tolerance $\epsilon$
        \Ensure state sequence $((\vec{x}_t,\vec{\Sigma}_t))_{t=1}^T$
        \For{$t=1$ \textbf{to} $T$}
            \State Initial mean $\vec{x}^{(t)}_{(0)}\gets\vec{F}_t\vec{x}_{t-1}$
            \State Initial covariance $\vec{\Sigma}^{(t)}_{(0)}\gets\vec{F}_t\vec{\Sigma}_{t-1}\vec{F}_t^T+\vec{Q}_t$
            \State Prior information matrix $\vec{Y}^{(t)}_{(0)}\gets\left(\vec{\Sigma}^{(t)}_{(0)}\right)^{-1}$
            \State Prior information vector $\vec{y}^{(t)}_{(0)}\gets\vec{Y}^{(t)}_{(0)}\vec{x}^{(t)}_{(0)}$
            \For{$n=1$ \textbf{to} $N+1$} \Comment{Iterative approximations}
                \State Measurement Jacobian $\vec{H}^{(t)}_{(n)}\gets\left.\frac{\partial\vec{h}(\vec{x})}{\partial\vec{x}}\right|_{\vec{x}=\vec{x}^{(t)}_{(n-1)}}$
                \State Pseudomeasurement $\vec{z}^{(t)}_{(n)}\gets\vec{z}_t-\vec{h}\left(\vec{x}^{(t)}_{(n-1)}\right)+\vec{H}^{(t)}_{(n)}\vec{x}^{(t)}_{(n-1)}$
                \State Posterior information matrix $\vec{Y}^{(t)}_{(n)}\gets\vec{Y}^{(t)}_{(0)}+\left(\vec{H}^{(t)}_{(n)}\right)^T\vec{R}_t^{-1}\vec{H}^{(t)}_{(n)}$
                \State Posterior information vector $\vec{y}^{(t)}_{(n)}\gets\vec{y}^{(t)}_{(0)}+\left(\vec{H}^{(t)}_{(n)}\right)^T\vec{R}_t^{-1}\vec{z}^{(t)}_{(n)}$
                \State Updated covariance $\vec{\Sigma}^{(t)}_{(n)}\gets\left(\vec{Y}^{(t)}_{(n)}\right)^{-1}$
                \If{$n=N+1$} \Comment{Final covariance update}
                    \State \textbf{break}
                \EndIf
                \State Updated mean $\vec{x}^{(t)}_{(n)}\gets\vec{\Sigma}^{(t)}_{(n)}\vec{y}^{(t)}_{(n)}$
                \If{$\left\|\vec{x}^{(t)}_{(n)}-\vec{x}^{(t)}_{(n-1)}\right\|_2<\epsilon$} \Comment{Optional early convergence}
                    \State $\vec{x}^{(t)}_{(N)}\gets\vec{x}^{(t)}_{(n)}$
                    \State \textbf{goto} final iteration: $n\gets N+1$
                \EndIf
            \EndFor
            \State Save this timestep's final mean: $\vec{x}_t\gets\vec{x}^{(t)}_{(N)}$
            \State Save this timestep's final covariance: $\vec{\Sigma}_t\gets\vec{\Sigma}^{(t)}_{(N+1)}$
        \EndFor
    \end{algorithmic}
\end{algorithm}

\begin{algorithm}
    \caption{Probabilistic Lucas-Kanade}
    \label{alg:plk}
    \linespread{1.5}\selectfont
    \begin{algorithmic}[1]
        \Require images $I_1, I_2$, tracked pixels $\mathcal{P}=(\vec{x}_m)_{m=1}^M\in\mathcal{D}^M$, pyramid height $L$, window size $K$, process noise variance $\omega$, measurement noise variance exponent $\gamma$, weighting standard deviation $\sigma$, per-level iterations $N$, convergence tolerance $\epsilon$
        \Ensure flow field estimate $\mathcal{F}=(\vec{f}(\vec{x}_m))_{m=1}^M$, flow field covariance $\mathcal{S}=(\vec{\Sigma}(\vec{x}_m))_{m=1}^M$, tracked pixels' estimated new positions $\mathcal{P}'=(\vec{x}_m+\vec{f}(\vec{x}_m))_{m=1}^M$
        \State Build and pad image pyramids $(I_1^{(\ell)})_{\ell=1}^L,(I_2^{(\ell)})_{\ell=1}^L$ by Eqs.~\eqref{eq:pyr},~\eqref{eq:pad}
        \For{$\ell=1$ \textbf{to} $L$} \Comment{Coarse-to-fine pyramid loop}
            \For{each tracked pixel $\vec{x}=\vec{x}_m\in\mathcal{P}$} \Comment{Process each tracked point}
                \State Downscale coordinate to current level: $\vec{x}^{(\ell)}\gets\vec{x}/2^{L-\ell}$
                \If{$\ell=1$} \Comment{Uninformative prior}
                    \State Initialize state estimate: $\vec{f}(\vec{x})^{(\ell)}_{(0)}\gets\vec{0}$
                    \State Initialize state covariance: $\vec{\Sigma}(\vec{x})^{(\ell)}_{(0)}\gets10^6\vec{I}_2$
                \Else \Comment{State transition}
                    \State Upscale state estimate from previous level: $\vec{f}(\vec{x})^{(\ell)}_{(0)}\gets2\vec{f}(\vec{x})^{(\ell-1)}$
                    \State Upscale state covariance, add process noise: $\vec{\Sigma}(\vec{x})^{(\ell)}_{(0)}\gets4\vec{\Sigma}(\vec{x})^{(\ell-1)}+\omega\vec{I}_2$
                \EndIf
                \State Compute prior information matrix: $\vec{Y}(\vec{x})^{(\ell)}_{(0)}\gets\left(\vec{\Sigma}(\vec{x})^{(\ell)}_{(0)}\right)^{-1}$
                \State Compute prior information vector: $\vec{y}(\vec{x})^{(\ell)}_{(0)}\gets\vec{Y}(\vec{x})^{(\ell)}_{(0)}\vec{f}(\vec{x})^{(\ell)}_{(0)}$
                \For{$n=1$ \textbf{to} $N+1$} \Comment{Iterated measurement update}
                    \State Initialize $\vec{Y}(\vec{x})\gets\vec{Y}(\vec{x})^{(\ell)}_{(0)}$
                    \State Initialize $\vec{y}(\vec{x})\gets\vec{y}(\vec{x})^{(\ell)}_{(0)}$
                    \For{each neighbor $\vec{y}\in\mathcal{W}_K(\vec{x}^{(\ell)})$}
                        \Statex\hspace{6em} Call one of three method subroutines (see following page)
    \setcounter{ALG@line}{22}
                    \EndFor
                    \State Update state covariance: $\vec{\Sigma}(\vec{x})^{(\ell)}_{(n)}\gets\vec{Y}(\vec{x})^{-1}$
                    \If{$n=N+1$} \Comment{Final covariance update}
                        \State \textbf{break}
                    \EndIf
                    \State Update state estimate: $\vec{f}(\vec{x})^{(\ell)}_{(n)}\gets\vec{\Sigma}(\vec{x})^{(\ell)}_{(n)}\vec{y}(\vec{x})$
                    \If{$\left\|\vec{f}(\vec{x})^{(\ell)}_{(n)}-\vec{f}(\vec{x})^{(\ell)}_{(n-1)}\right\|_2<\epsilon$} \Comment{Optional early convergence}
                        \State $\vec{f}(\vec{x})^{(\ell)}_{(N)}\gets\vec{f}(\vec{x})^{(\ell)}_{(n)}$
                        \State \textbf{goto} final iteration: $n\gets N+1$
                    \EndIf
                \EndFor
        \algstore{plk}
    \end{algorithmic}
\end{algorithm}

\addtocounter{algorithm}{-1}
\begin{algorithm}
    \caption{Probabilistic Lucas-Kanade (continued)}
    \linespread{1.5}\selectfont
    \begin{algorithmic}
        \algrestore{plk}
                \State Save this level's posterior estimate: $\vec{f}(\vec{x})^{(\ell)}\gets\vec{f}(\vec{x})^{(\ell)}_{(N)}$
                \State Save this level's posterior covariance: $\vec{\Sigma}(\vec{x})^{(\ell)}\gets\vec{\Sigma}(\vec{x})^{(\ell)}_{(N+1)}$
                \If{$\ell=L$}
                    \State Save final posterior estimate: $\vec{f}(\vec{x})\gets\vec{f}(\vec{x})^{(\ell)}$
                    \State Save final posterior covariance: $\vec{\Sigma}(\vec{x})\gets\vec{\Sigma}(\vec{x})^{(\ell)}$
                    \State Save estimated new position: $\vec{x}'_m\gets\vec{x}+\vec{f}(\vec{x})$
                \EndIf
            \EndFor
        \EndFor
    \end{algorithmic}
\end{algorithm}

\setcounter{algorithm}{0}
\renewcommand{\thealgorithm}{4.\arabic{algorithm}}
\begin{algorithm}
    \caption{PLK-Intensity Subroutine}
    \label{alg:plki}
    \linespread{1.5}\selectfont
    \begin{algorithmic}[1]
    \setcounter{ALG@line}{17}
        \State Compute $\nabla I(\vec{y})\gets\nabla I^{(\ell)}\left(\vec{y};\vec{f}(\vec{x})^{(\ell)}_{(n-1)}\right)$, as in Eq.~\eqref{eq:plkigrad} by Eq.~\eqref{eq:plkispat}
        \State Compute $I_t(\vec{y})\gets I_1^{(\ell)}(\vec{y})-I_2^{(\ell)}\left(\vec{y}+\vec{f}(\vec{x})^{(\ell)}_{(n-1)}\right)$, as in Eq.~\eqref{eq:plkimeas} by Eq.~\eqref{eq:plkitemp}
        \State Compute $\sigma_R(\vec{y})^{(\ell)}_{(n)}$ using $\vec{f}(\vec{x})^{(\ell)}_{(n-1)}$ by Eq.~\eqref{eq:plkivar}
        \State Accumulate $\vec{Y}(\vec{x})\gets\vec{Y}(\vec{x})+\left(\sigma_R(\vec{y})^{(\ell)}_{(n)}\right)^{-1}w_{\sigma}(\vec{y}-\vec{x})\nabla I(\vec{y})\nabla I(\vec{y})^T$
        \State Accumulate $\vec{y}(\vec{x})\gets\vec{y}(\vec{x})+\left(\sigma_R(\vec{y})^{(\ell)}_{(n)}\right)^{-1}w_{\sigma}(\vec{y}-\vec{x})\nabla I(\vec{y})\left(I_t(\vec{y})+\nabla I(\vec{y})^T\vec{f}(\vec{x})^{(\ell)}_{(n-1)}\right)$
    \end{algorithmic}
\end{algorithm}

\begin{algorithm}
    \caption{PLK-Gradient Subroutine}
    \label{alg:plkg}
    \linespread{1.5}\selectfont
    \begin{algorithmic}[1]
    \setcounter{ALG@line}{17}
        \State Compute $\nabla^2I(\vec{y})\gets\nabla^2I^{(\ell)}\left(\vec{y};\vec{f}(\vec{x})^{(\ell)}_{(n-1)}\right)$, as in Eq.~\eqref{eq:plkggrad} by Eq.~\eqref{eq:plkgspat}
        \State Compute $\nabla I_t(\vec{y})\gets\nabla I_1^{(\ell)}(\vec{y})-\nabla I_2^{(\ell)}\left(\vec{y}+\vec{f}(\vec{x})^{(\ell)}_{(n-1)}\right)$, as in Eq.~\eqref{eq:plkgmeas} by Eq.~\eqref{eq:plkispat}
        \State Compute $\vec{R}(\vec{y})\gets\operatorname{diag}\left(\sigma_{Rx}(\vec{y})^{(\ell)}_{(n)},\sigma_{Ry}(\vec{y})^{(\ell)}_{(n)}\right)$ using $\vec{f}(\vec{x})^{(\ell)}_{(n-1)}$ by Eq.~\eqref{eq:plkgvar}
        \State Accumulate $\vec{Y}(\vec{x})\gets\vec{Y}(\vec{x})+w_{\sigma}(\vec{y}-\vec{x})\nabla^2I(\vec{y})\vec{R}(\vec{y})^{-1}\nabla^2I(\vec{y})$
        \State Accumulate $\vec{y}(\vec{x})\gets\vec{y}(\vec{x})+w_{\sigma}(\vec{y}-\vec{x})\nabla^2I(\vec{y})\vec{R}(\vec{y})^{-1}\left(\nabla I_t(\vec{y})+\nabla^2I(\vec{y})\vec{f}(\vec{x})^{(\ell)}_{(n-1)}\right)$
    \end{algorithmic}
\end{algorithm}

\begin{algorithm}
    \caption{PLK-Combined Subroutine}
    \label{alg:plkc}
    \linespread{1.5}\selectfont
    \begin{algorithmic}[1]
    \setcounter{ALG@line}{17}
        \State Call PLK-Intensity subroutine above
        \State Call PLK-Gradient subroutine above
    \end{algorithmic}
\end{algorithm}

\setcounter{algorithm}{4}
\renewcommand{\thealgorithm}{\arabic{algorithm}}
\begin{algorithm}
    \caption{Probabilistic Farnebäck}
    \label{alg:pfb}
    \linespread{1.5}\selectfont
    \begin{algorithmic}[1]
        \Require images $I_1, I_2$, tracked pixels $\mathcal{P}=(\vec{x}_m)_{m=1}^M\in\mathcal{D}^M$, pyramid height $L$, window size $K$, process noise variance $\omega$,  measurement noise variance exponent $\gamma$, weighting standard deviation $\sigma$, polynomial fit window size $K_p$, polynomial fit weighting standard deviation $\sigma_p$, per-level iterations $N$, convergence tolerance $\epsilon$
        \Ensure flow field estimate $\mathcal{F}=(\vec{f}(\vec{x}_m))_{m=1}^M$, flow field covariance $\mathcal{S}=(\vec{\Sigma}(\vec{x}_m))_{m=1}^M$, tracked pixels' estimated new positions $\mathcal{P}'=(\vec{x}_m+\vec{f}(\vec{x}_m))_{m=1}^M$
        \State Build and pad image pyramids $(I_1^{(\ell)})_{\ell=1}^L,(I_2^{(\ell)})_{\ell=1}^L$ by Eqs.~\eqref{eq:pyr},~\eqref{eq:pad}
        \For{$\ell=1$ \textbf{to} $L$} \Comment{Coarse-to-fine pyramid loop}
            \For{each pixel $\vec{y}\in\mathcal{D}^{(\ell)}$} \Comment{Pre-compute Farnebäck coefficients}
                \State Compute and save $\vec{A}_1^{(\ell)}(\vec{y}),\vec{b}_1^{(\ell)}(\vec{y}),c_1^{(\ell)}(\vec{y}),\vec{A}_2^{(\ell)}(\vec{y}),\vec{b}_2^{(\ell)}(\vec{y}),c_2^{(\ell)}(\vec{y})$ by Eq.~\eqref{eq:fbplsq}
            \EndFor
            \For{each tracked pixel $\vec{x}=\vec{x}_m\in\mathcal{P}$} \Comment{Process each tracked point}
                \State Downscale coordinate to current level: $\vec{x}^{(\ell)}\gets\vec{x}/2^{L-\ell}$
                \If{$\ell=1$} \Comment{Uninformative prior}
                    \State Initialize state estimate: $\vec{f}(\vec{x})^{(\ell)}_{(0)}\gets\vec{0}$
                    \State Initialize state covariance: $\vec{\Sigma}(\vec{x})^{(\ell)}_{(0)}\gets10^6\vec{I}_2$
                \Else \Comment{State transition}
                    \State Upscale state estimate from previous level: $\vec{f}(\vec{x})^{(\ell)}_{(0)}\gets2\vec{f}(\vec{x})^{(\ell-1)}$
                    \State Upscale state covariance, add process noise: $\vec{\Sigma}(\vec{x})^{(\ell)}_{(0)}\gets4\vec{\Sigma}(\vec{x})^{(\ell-1)}+\omega\vec{I}_2$
                \EndIf
                \State Compute prior information matrix: $\vec{Y}(\vec{x})^{(\ell)}_{(0)}\gets\left(\vec{\Sigma}(\vec{x})^{(\ell)}_{(0)}\right)^{-1}$
                \State Compute prior information vector: $\vec{y}(\vec{x})^{(\ell)}_{(0)}\gets\vec{Y}(\vec{x})^{(\ell)}_{(0)}\vec{f}(\vec{x})^{(\ell)}_{(0)}$
                \For{$n=1$ \textbf{to} $N+1$} \Comment{Iterated measurement update}
                    \State Initialize $\vec{Y}(\vec{x})\gets\vec{Y}(\vec{x})^{(\ell)}_{(0)}$
                    \State Initialize $\vec{y}(\vec{x})\gets\vec{y}(\vec{x})^{(\ell)}_{(0)}$
                    \For{each neighbor $\vec{y}\in\mathcal{W}_K(\vec{x}^{(\ell)})$}
                        \Statex\hspace{6em} Call one of three method subroutines (see following page)
    \setcounter{ALG@line}{25}
                    \EndFor
                    \State Update state covariance: $\vec{\Sigma}(\vec{x})^{(\ell)}_{(n)}\gets\vec{Y}(\vec{x})^{-1}$
                    \If{$n=N+1$} \Comment{Final covariance update}
                        \State \textbf{break}
                    \EndIf
                    \State Update state estimate: $\vec{f}(\vec{x})^{(\ell)}_{(n)}\gets\vec{\Sigma}(\vec{x})^{(\ell)}_{(n)}\vec{y}(\vec{x})$
                    \If{$\left\|\vec{f}(\vec{x})^{(\ell)}_{(n)}-\vec{f}(\vec{x})^{(\ell)}_{(n-1)}\right\|_2<\epsilon$} \Comment{Optional early convergence}
                        \State $\vec{f}(\vec{x})^{(\ell)}_{(N)}\gets\vec{f}(\vec{x})^{(\ell)}_{(n)}$
                        \State \textbf{goto} final iteration: $n\gets N+1$
                    \EndIf
                \EndFor
        \algstore{pfb}
    \end{algorithmic}
\end{algorithm}

\addtocounter{algorithm}{-1}
\begin{algorithm}
    \caption{Probabilistic Farnebäck (continued)}
    \linespread{1.5}\selectfont
    \begin{algorithmic}
        \algrestore{pfb}
                \State Save this level's posterior estimate: $\vec{f}(\vec{x})^{(\ell)}\gets\vec{f}(\vec{x})^{(\ell)}_{(N)}$
                \State Save this level's posterior covariance: $\vec{\Sigma}(\vec{x})^{(\ell)}\gets\vec{\Sigma}(\vec{x})^{(\ell)}_{(N+1)}$
                \If{$\ell=L$}
                    \State Save final posterior estimate: $\vec{f}(\vec{x})\gets\vec{f}(\vec{x})^{(\ell)}$
                    \State Save final posterior covariance: $\vec{\Sigma}(\vec{x})\gets\vec{\Sigma}(\vec{x})^{(\ell)}$
                    \State Save estimated new position: $\vec{x}'_m\gets\vec{x}+\vec{f}(\vec{x})$
                \EndIf
            \EndFor
        \EndFor
    \end{algorithmic}
\end{algorithm}

\setcounter{algorithm}{0}
\renewcommand{\thealgorithm}{5.\arabic{algorithm}}
\begin{algorithm}
    \caption{PFB-Intensity Subroutine}
    \label{alg:pfbi}
    \linespread{1.5}\selectfont
    \begin{algorithmic}[1]
    \setcounter{ALG@line}{20}
        \State Compute $\vec{b}(\vec{y})\gets\vec{b}^{(\ell)}\left(\vec{y};\vec{f}(\vec{x})^{(\ell)}_{(n-1)}\right)$, as in Eq.~\eqref{eq:pfbigrad} by Eq.~\eqref{eq:pfbispat}
        \State Compute $\Delta c(\vec{y})\gets c_1^{(\ell)}(\vec{y})-c_2^{(\ell)}\left(\vec{y}+\vec{f}(\vec{x})^{(\ell)}_{(n-1)}\right)$, as in Eq.~\eqref{eq:pfbimeas} by Eq.~\eqref{eq:pfbispat}
        \State Compute $\sigma_R(\vec{y})^{(\ell)}_{(n)}$ using $\vec{f}(\vec{x})^{(\ell)}_{(n-1)}$ by Eq.~\eqref{eq:pfbivar}
        \State Accumulate $\vec{Y}(\vec{x})\gets\vec{Y}(\vec{x})+\left(\sigma_R(\vec{y})^{(\ell)}_{(n)}\right)^{-1}w_{\sigma}(\vec{y}-\vec{x})\vec{b}(\vec{y})\vec{b}(\vec{y})^T$
        \State Accumulate $\vec{y}(\vec{x})\gets\vec{y}(\vec{x})+\left(\sigma_R(\vec{y})^{(\ell)}_{(n)}\right)^{-1}w_{\sigma}(\vec{y}-\vec{x})\vec{b}(\vec{y})\left(\Delta c(\vec{y})+\vec{b}(\vec{y})^T\vec{f}(\vec{x})^{(\ell)}_{(n-1)}\right)$
    \end{algorithmic}
\end{algorithm}

\begin{algorithm}
    \caption{PFB-Gradient Subroutine}
    \label{alg:pfbg}
    \linespread{1.5}\selectfont
    \begin{algorithmic}[1]
    \setcounter{ALG@line}{20}
        \State Compute $\vec{A}(\vec{y})\gets\vec{A}^{(\ell)}\left(\vec{y};\vec{f}(\vec{x})^{(\ell)}_{(n-1)}\right)$, as in Eq.~\eqref{eq:pfbggrad} by Eq.~\eqref{eq:pfbispat}
        \State Compute $\Delta\vec{b}(\vec{y})\gets\vec{b}_1^{(\ell)}(\vec{y})-\vec{b}_2^{(\ell)}\left(\vec{y}+\vec{f}(\vec{x})^{(\ell)}_{(n-1)}\right)$, as in Eq.~\eqref{eq:pfbgmeas} by Eq.~\eqref{eq:pfbispat}
        \State Compute $\vec{R}(\vec{y})\gets\operatorname{diag}\left(\sigma_{Rx}(\vec{y})^{(\ell)}_{(n)},\sigma_{Ry}(\vec{y})^{(\ell)}_{(n)}\right)$ using $\vec{f}(\vec{x})^{(\ell)}_{(n-1)}$ by Eq.~\eqref{eq:pfbgvar}
        \State Accumulate $\vec{Y}(\vec{x})\gets\vec{Y}(\vec{x})+4w_{\sigma}(\vec{y}-\vec{x})\vec{A}(\vec{y})\vec{R}(\vec{y})^{-1}\vec{A}(\vec{y})$
        \State Accumulate $\vec{y}(\vec{x})\gets\vec{y}(\vec{x})+2w_{\sigma}(\vec{y}-\vec{x})\vec{A}(\vec{y})\vec{R}(\vec{y})^{-1}\left(\Delta\vec{b}(\vec{y})+2\vec{A}(\vec{y})\vec{f}(\vec{x})^{(\ell)}_{(n-1)}\right)$
    \end{algorithmic}
\end{algorithm}

\begin{algorithm}
    \caption{PFB-Combined Subroutine}
    \label{alg:pfbc}
    \linespread{1.5}\selectfont
    \begin{algorithmic}[1]
    \setcounter{ALG@line}{20}
        \State Perform the PFB-Intensity subroutine above
        \State Perform the PFB-Gradient subroutine above
    \end{algorithmic}
\end{algorithm}
\end{appendices}
\end{document}